\documentclass{article}

\RequirePackage[pagebackref,breaklinks]{hyperref}

\usepackage{eccvabbrv}

\usepackage{graphicx}
\usepackage{booktabs}

\usepackage[accsupp]{axessibility}  %

\usepackage{orcidlink}

\usepackage[utf8]{inputenc} %
\usepackage[T1]{fontenc}    %

\usepackage{url} %

\usepackage[dvipsnames,table]{xcolor} %

\usepackage{graphicx}       %
\usepackage{booktabs}       %
\usepackage{multicol}       %
\usepackage{multirow}       %
\usepackage{makecell}       %
\usepackage{colortbl}       %
\usepackage{caption}        %
\usepackage{subcaption}     %
\usepackage{wrapfig}        %
\usepackage{adjustbox}      %
\usepackage{titletoc}       %

\usepackage{mathtools}
\usepackage{amsmath} %
\usepackage{amsfonts}       %
\usepackage{amssymb}        %
\usepackage{bm}             %
\usepackage{nicefrac}       %
\usepackage{microtype}      %
\usepackage[scaled=1,varqu]{zi4} %
\usepackage{enumitem}       %
\usepackage{listings}       %
\usepackage{textcomp}       %

\usepackage[capitalize,nameinlink]{cleveref}  %

\usepackage{comment}        %
\usepackage{array}
\usepackage{relsize}         %

\usepackage{doi}            %
\usepackage[normalem]{ulem} %
\usepackage{placeins}       %
\usepackage[detect-all]{siunitx} %

\definecolor{backcolour}{HTML}{F5F5F5}
\definecolor{rulecolour}{HTML}{E0E0E0}
\definecolor{codegreen}{HTML}{008000}
\definecolor{codemagenta}{HTML}{AA22FF}
\definecolor{codered}{HTML}{BA2121}
\definecolor{codeblue}{HTML}{408080}
\usepackage{bbding}         %
\definecolor{cold}{HTML}{008acd}
\def\frozen{\raisebox{-.15ex}{\color{cold}\SnowflakeChevron}}
\DeclareUnicodeCharacter{2744}{\frozen}%

\usepackage{pifont}         %
\def\cmark{\ding{51}} %
\def\xmark{\ding{55}}
\definecolor{vlightgray}{gray}{0.87}

\definecolor{lightgray}{gray}{0.7}
\def\na{{\smaller\color{lightgray}N/A}}

\newcommand{\gray}[1]{{\color{gray}#1}}

\definecolor{tblue}{HTML}{1F77B4}
\definecolor{torange}{HTML}{FF7F0E}
\definecolor{tgreen}{HTML}{2CA02C}
\definecolor{tred}{HTML}{FF0000}

\definecolor{linkcolor}{HTML}{991408}  %
\definecolor{citecolor}{HTML}{2E7E2A}  %
\definecolor{filecolor}{HTML}{131877}  %
\definecolor{menucolor}{HTML}{727500}  %
\definecolor{runcolor} {HTML}{137776}  %
\definecolor{urlcolor} {HTML}{0a2bbf}  %
\definecolor{linkblue}{HTML}{0057B8}%

\makeatletter
\DeclareRobustCommand*{\corrmark@}{%
  \begingroup
  \setbox\z@\hbox{$\m@th\ast$}%
  \setbox\tw@\hbox{1}%
  \resizebox{!}{\ht\tw@}{\raisebox{\dp\z@}{\usebox\z@}}%
  \endgroup}
\newcommand*{\corrmark}{{}\corrmark@}
\makeatother

\newcommand{\appref}[1]{\hyperref[#1]{Appendix~\ref*{#1}}}%

\crefname{section}{Sec.}{Sections}
\crefname{appendix}{Appx.}{Appendices}
\renewcommand*{\appref}[1]{\hyperref[#1]{Appx.~\ref*{#1}}}

\def\Snospace~{\S{}}%

\newcommand{\mypara}[1]{\noindent\textbf{#1}}

\newcommand{\best}[1]{\ifmmode\mathbf{#1}\else\textbf{#1}\fi}
\newcommand{\secbest}[1]{\underline{#1}}
\newcommand{\sbest}[1]{\secbest{#1}}

\newcommand{\tpm}[1]{{\footnotesize$\pm #1$}}
\newcommand{\missing}[0]{--}

\newcommand{\incr}[1]{{\color{ForestGreen}$#1$}}
\newcommand{\decr}[1]{{\color{red}$#1$}}

\newcommand{\aincr}[1]{{\smaller\smaller\color{ForestGreen}$#1$}}
\newcommand{\adecr}[1]{{\smaller\smaller\color{red}$#1$}}

\newcommand{\hider}[1]{}

\newcommand{\todo}[1]{}

\newif\ifshowchanges
\showchangesfalse

\usepackage{etoolbox}       %
\newtoggle{arxiv}%
\toggletrue{arxiv}%
\newtoggle{useappendix}
\toggletrue{useappendix}
\iftoggle{arxiv}{
    \hypersetup{colorlinks=true, allcolors=linkblue}

    \usepackage[parfill]{parskip}

    \usepackage[%
        a4paper,
        inner=25mm,
        outer=25mm,%
        top=25mm,
        bottom=25mm,
        marginparsep=5mm,
        marginparwidth=40mm,
    ]{geometry}

    \newlength{\defbaselineskip}
    \setlength{\defbaselineskip}{\baselineskip}
    \setlength{\parskip}{6pt}%
    \setlength{\parindent}{0pt}%

    \RequirePackage[T1]{fontenc}
    \RequirePackage[tt=false, type1=true]{libertine}
    \RequirePackage[libertine]{newtxmath}

    \usepackage{fancyhdr}
    \pagestyle{fancy}
    \fancyhead[L]{Bootleg: Self-distillation of hidden layers for masked SSL}
    \fancyhead[R]{Scott C. Lowe, et al. (2026)}

    \usepackage{natbib}

    \captionsetup{font=small,labelfont=it}
}{
}

\providecommand{\citep}{\cite}
\providecommand{\citet}{\cite}
\providecommand{\citealp}{\cite}

\title{Self-Distillation of Hidden Layers for Masked Self-Supervised Representation~Learning}

\makeatletter
\def\@fnsymbol#1{%
   \ifcase#1\or
   \TextOrMath{\larger\larger\larger\textasteriskcentered}*\or
   \TextOrMath \textdagger \dagger\or
   \TextOrMath \textdaggerdbl \ddagger \or
   \TextOrMath \textsection  \mathsection\or
   \TextOrMath \textparagraph \mathparagraph\or
   \TextOrMath \textbardbl \|\or
   \TextOrMath {\textdagger\textdagger}{\dagger\dagger}\or
   \TextOrMath {\textdaggerdbl\textdaggerdbl}{\ddagger\ddagger}\else
   \@ctrerr \fi
}
\makeatother
\author{
\normalfont
\textbf{Scott~C.~Lowe}\textsuperscript{1,}\thanks{Correspondence: \texttt{scott.lowe@vectorinstitute.ai}}\quad
\textbf{Anthony~Fuller}\textsuperscript{1,2}\quad
\textbf{Sageev~Oore}\textsuperscript{1,3}\\
\textbf{Evan~Shelhamer}\textsuperscript{1,4}\quad
\textbf{Graham~W.~Taylor}\textsuperscript{1,5}
\vspace{0.4em} \\
\textsuperscript{1}Vector Institute\quad
\textsuperscript{2}Carleton University\quad
\textsuperscript{3}Dalhousie University\\
\textsuperscript{4}University of British Columbia\quad
\textsuperscript{5}University of Guelph
}

\date{}%

\begin{document}

\maketitle

\begin{abstract}

The landscape of self-supervised learning (SSL) is currently dominated by generative approaches (\eg{} MAE) that reconstruct raw low-level data, and predictive approaches (\eg{} I-JEPA) that predict high-level abstract embeddings.
While generative methods are stable due to their reliable training targets based on ground-truth data, they are computationally inefficient for high-redundancy modalities like imagery, and their training objective does not prioritize learning high-level, conceptual features.
Conversely, predictive methods often suffer from training instability due to their reliance on the non-stationary targets of final-layer self-distillation.
We introduce \textbf{Bootleg}, a method that bridges this divide by tasking the model with predicting latent representations from \emph{multiple hidden layers} of a teacher.
This hierarchical objective forces the model to capture features at varying levels of abstraction simultaneously.
We demonstrate Bootleg significantly outperforms comparable single-view baselines (+10\% vs. I-JEPA) on frozen probe classification of ImageNet-1K, iNaturalist-21, and VTAB, and semantic segmentation of ADE20K, Cityscapes, and COCO-Stuff.
Code is available at \url{https://github.com/scottclowe/bootleg}.

\end{abstract}

\begin{figure*}
    \centering
    \includegraphics[width=1.0\linewidth]{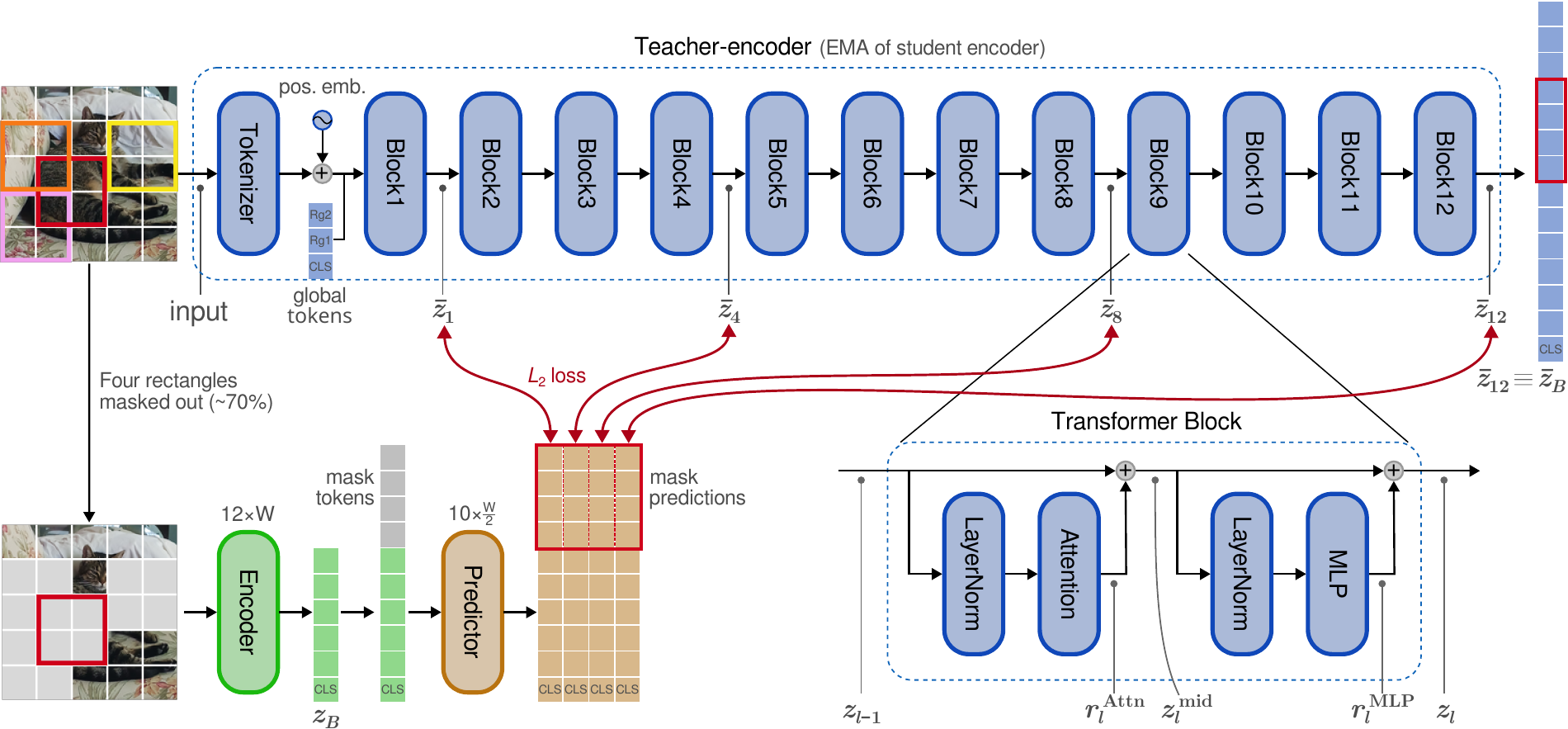}
    \caption{%
    \textbf{Multi-layer self-distillation with Bootleg.}
    The teacher-encoder (blue), student-encoder (green), and predictor (orange) are ViTs, made of repeated transformer blocks.
    A schematic of a single transformer block is overlaid (bottom right).
    The teacher-encoder is an EMA of the student-encoder, and processes the full image.
    The student-encoder sees a subset of the image and must create embeddings of them to facilitate the predictor.
    The predictor processes the embeddings to predict representations at multiple layers within the teacher-encoder.}
    \label{fig:method}
\end{figure*}

\section{Introduction}

Large-scale pretraining using self-supervised learning (SSL) is an incredibly powerful representation learning technique, able to efficiently use large volumes of unlabelled data, and is thus used in many domains and applications.
The most powerful image SSL methods at present, such as DINOv3 \citep{simeoni2025dinov3} and Franca \citep{venkataramanan2025franca}, are joint-embedding methods that enforce agreement between multiple augmented views of each image (multi-view self-distillation and clustering methods in the taxonomy of \citealp{balestriero2023cookbook}), requiring domain-specific data augmentation and several forward passes per image, resulting in tremendous computational resources being needed (\eg{} the smallest Franca training is distributed across 32 H100 GPUs).
Meanwhile, single-view masked methods are more computationally accessible (in processing a single view of each image) and general (in not requiring specific augmentations).

In particular, single-view masked techniques introduced by MAE \citep{he2022masked} and I-JEPA \citep{assran2023ijepa} have been widely adopted across domains and modalities due to their flexibility \eg{} in video \citep{wang2023videomaev2, bardes2024vjepa, assran2025vjepa2}, audio \citep{huang2022audiomae, fei2024ajepa, tuncay2025audiojepa, yuksel2025wavjepa}, remote sensing \citep{cong2022satmae, fuller2022satvit}, medical imaging \citep{zhou2023self, xiao2023delving}, 3D point clouds \citep{zhang2022point, chen2023pimae}, time series \citep{nie2023a, li2023ti}, DNA modelling \citep{safari2025enhancingdnafoundationmodels}, multimodal data \citep{bachmann2022multimae, chen2025vljepa, lei2025m3jepa}, and more \citep{dong2024brainjepa, Riou2024_StemJEPA, thimonier2025tjepa, hachana2025using}.
However, both have their own disadvantages.
MAE builds a representation learning model out of a generative modelling task, resulting in embeddings which are useful in pixel-space, but are less capable of representing high-level features.
Consequently, MAE pretrained models require fine-tuning to be useful for downstream tasks, and cannot reasonably be used without fine-tuning.
On the other hand, I-JEPA relies on \emph{last layer} self-distillation, which is less stable to train because the targets are derived from the network we are training.
This circularity means the training targets are less stimulus-driven/grounded, which can result in the model being detached from reality and collapsing during training.
We thus introduce Bootleg, which learns superior representations by reconstructing \emph{multiple hidden layers} in its own teacher-encoder.
Our contributions:
\begin{enumerate}[nosep, leftmargin=*]
\item Propose predicting multiple hidden layers of the teacher-encoder; apply this to several existing methods, and show it increases training stability and robustness.
\item Evaluate extensively on classification and segmentation, with frozen probes and fine-tuning.
\item Identify and fix five bugs in I-JEPA's masking implementation.
\item Find every 4th block to be the best set of targets across three model sizes.
\item Find that much slower learning rate warmup improves JEPA-style training.
\item Introduce a no-forward MSE loss that more than halves the cost of computing many loss terms.
\end{enumerate}

\section{Background: ViTs and masked modelling}

We review further related work, including methods that exploit intermediate layers or frozen teachers, in \cref{s:related-work}.

\mypara{Vision transformers.}
ViTs \citep{dosovitskiy2021vit} split images into patches: $\mathbb{R}^{R \times R \times C} \to \mathbb{R}^{N \times P \cdot P \cdot C}$, where $N$ is the number of patches, $P$ is patch size, $R$ is image height and width, and $C$ is number of channels.
Patches are then projected to the model embedding dimension $D$ with positional embeddings added to make ``patch tokens'': $x \in \mathbb{R}^{N \times P \cdot P \cdot C} \to z^{\text{patch}} \in \mathbb{R}^{N \times D}$.
ViTs then optionally prepend $G$ global tokens (class and register tokens, \citealp{darcet2024registers}) and process the token embeddings with $B$ blocks: $z_0:=(z^{\text{glob}},z^{\text{patch}})$ and $z_{l+1}=\operatorname{Block}_l(z_l)$, where $z_l \in  \mathbb{R}^{(G+N) \times D}$ is the feature map after the $l$-th block.
Each block is comprised of a self-attention sub-block and a two-layer perceptron (MLP) sub-block, each of which adds to the representation through a residual connection:
$r_l^\text{Attn} := \operatorname{Attn}_l(\operatorname{LN}(z_{l-1}))$,
$z_l^\text{mid} := z_{l-1} + r_l^\text{Attn}$,
$r_l^\text{MLP}:=\operatorname{MLP}_l(\operatorname{LN}(z_l^\text{mid}))$, and
$z_l := z_l^\text{mid} + r_l^\text{MLP}$.
ViTs can efficiently process a sparse subset of patches via ``masking'', in which a subset of the $N$ patch tokens are dropped according to a masking strategy.
Masking significantly decreases computational cost since masked-out tokens (and their self-attentive interactions with other tokens) do not need to be processed;
this pairs well with ``fill-in-the-blank'' style pretraining algorithms such as those we consider here.

\mypara{MAE and I-JEPA.}
The {masked autoencoder} (MAE) and {image-based joint-embedding predictive architecture} (I-JEPA) are popular methods for pretraining that rely on masking for self-supervised learning.
Both methods train a model comprised of a ViT encoder and predictor, trained end to end.
Most image patches (around 75\%) are masked-out, and the remaining unmasked/visible patches are given to the encoder to produce visible-patch embeddings, $z_B^\text{vis}$.
The visible-patch embeddings are concatenated with mask embeddings, $m^\text{mask}$, that represent the hidden-patch locations---these act as placeholders to be updated with the decoder's prediction.
In MAE, the model's training objective is to predict the \emph{pixels} of the hidden patches ($\text{target}=x^\text{mask}$); whereas in I-JEPA, the models target the \emph{final embeddings} of the hidden patches ($\text{target}=\bar{z}_B^\text{mask}$).
I-JEPA's target embeddings are computed by processing the full image with a teacher-encoder model, which follows the (student) encoder by using an exponential moving average (EMA) of its weights.
Conceptually, this choice of \emph{target} is the key difference between MAE and I-JEPA.
Yet, as we will demonstrate, there are several implementation details that are also crucial.
For example, MAE samples masks uniformly at random, whereas I-JEPA uses a more involved masking strategy where four contiguous rectangular regions are masked out.

\mypara{Data2vec.}
Data2vec \citep{baevski2022data2vec, baevski2023data2vec2} uses a similar training methodology to MAE and I-JEPA with self-distillation.
Its target is the mean of instance-norm standardized vectors from the final $K$ layers of the teacher, with the implicit assumption that representations across a large section of the network can be linearly integrated together.
With a focus on efficient training, its predictor is a small convolutional module %
and each pass through the teacher module is accompanied with 16 passes through the student encoder-predictor with different randomly sampled masks.\looseness-1

\mypara{V-JEPA~2.1.}
Concurrently with our work, V-JEPA~2.1 \citep{murlabadia2026vjepa21} also predicts intermediate layers of the teacher (``deep self-supervision'') for video SSL, together with a dense loss on the context tokens and multi-level inputs to the predictor.
We evaluate these additional components with Bootleg in \cref{s:vjepa21}.

\begin{figure*}
    \centering%
    \includegraphics[scale=\iftoggle{arxiv}{0.41}{0.36}]{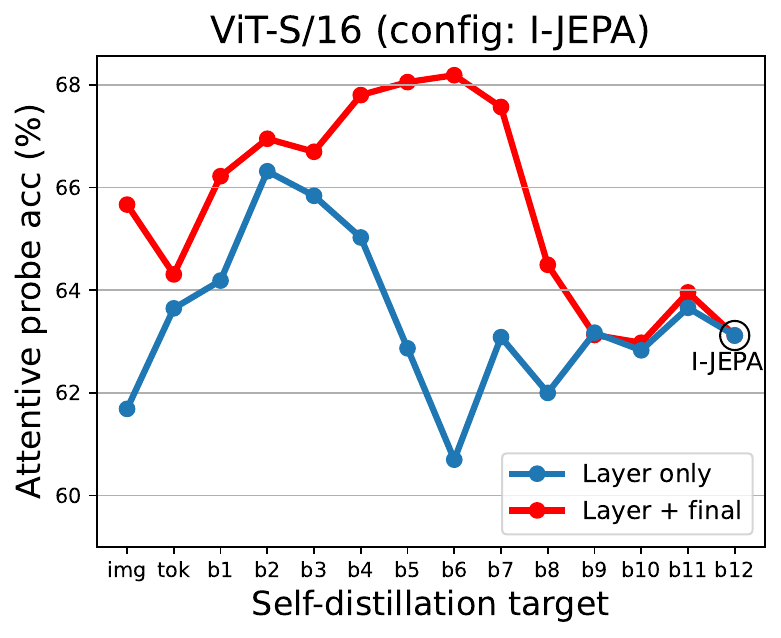}
    \includegraphics[scale=\iftoggle{arxiv}{0.41}{0.36}]{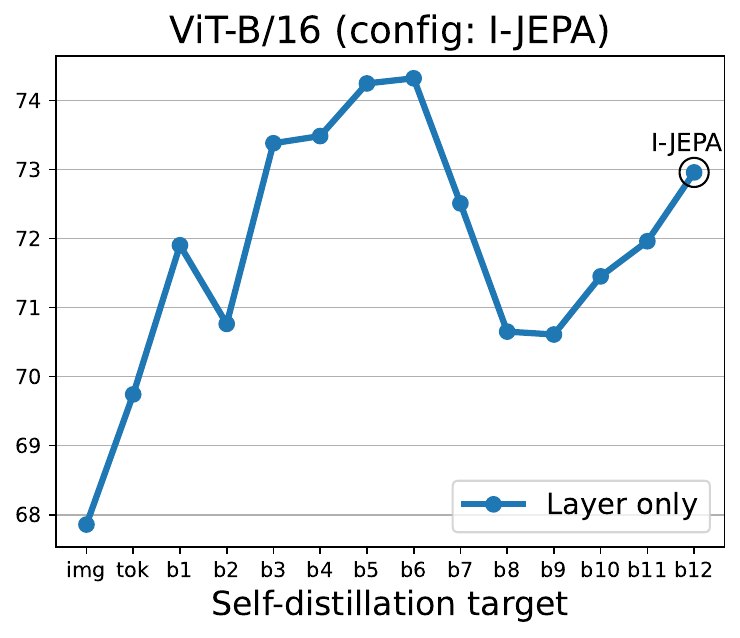}
    \includegraphics[scale=\iftoggle{arxiv}{0.41}{0.36}]{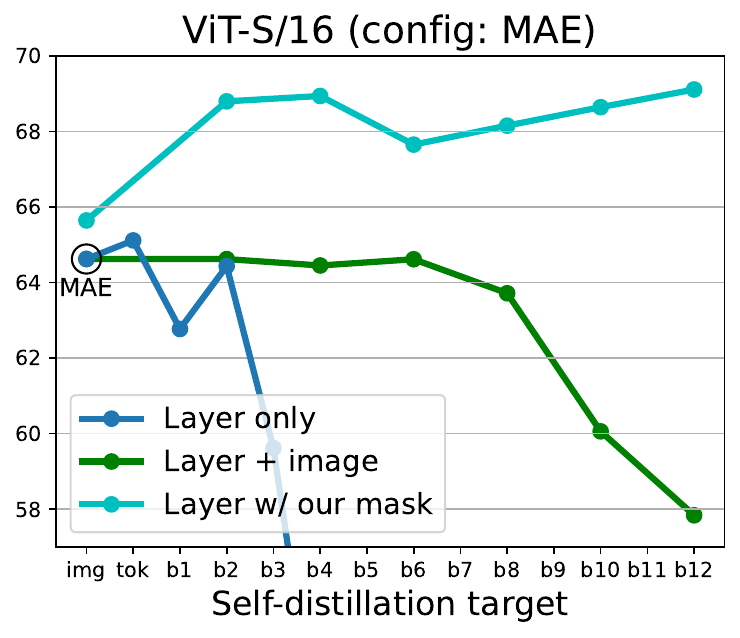}%
    \caption{
    \textbf{Bridging I-JEPA and MAE with %
    targets across hidden layers.}
    \textit{Left:} We train ViT-S with I-JEPA, except for the target which we change to be a hidden layer ($x$-axis) of the teacher-encoder instead of its final output (blue curve) or in addition to the final output (red). We plot frozen attentive probe top-1 accuracy on IN-1k ($y$-axis) for the respective encoders.
    \textit{Middle:} Similar, but for ViT-B to verify at larger scale.
    \textit{Right:} We train ViT-S using MAE, except we add EMA for self-distillation and change the target to be a hidden layer of the EMA model instead of input pixels (blue) or in addition to  image pixels (green). We compare to a single target with our masking strategy (cyan). In all cases, self-distillation of a hidden target is able to improve on predicting only input or output.
}
    \label{fig:stabilizing}
\end{figure*}

\section{Motivation for hidden targets}

\subsection{Intuition: abstraction, tasks, and stability}

\mypara{Deeper embeddings are more abstract.} The abstraction level of representations generally increases with increasing network depth.
Image models receive pixels, which are the lowest level of abstraction, as their input and transform them layer-by-layer into features that are more useful for the task.
Models tend to represent ``low-level'' features in early layers (\eg{} colours, edges, and textures) and ``high-level'' features in deep layers (\eg{} objects and semantic concepts).
This has been observed for CNNs \citep{zeiler2014visualizing, bau2017network, zhou2014object} and ViTs \citep{Raghu2021doViTsSeeCNNs, vilas2023analyzing} trained with supervised learning or various SSL algorithms \citep{park2023what, lepori2024beyond}.
Similarly, other work has found that hidden representations in CNNs and ViTs are correlated with the visual hierarchy in human brains, with early layers of the network best correlated with activity in the primary visual cortex that represents low-level features (oriented edges) and subsequent layers correlated with activity in progressively higher-level brain regions \citep{Yamins2014, Yamins2016, Horikawa2017, Tang2025, raugel2025disentanglingfactorsconvergencebrains}.
Our training approach can be seen as analogous to a predictive coding model of the brain \citep{Rao1999, friston2005, Friston2009, Clark2013, Keller2018, Nejad2025}---if we view the embeddings at masked locations as being desynchronous forward propagations of the same stimulus, the neurons corresponding to seen locations aim to predict the afferent representations which are about to be received from all layers to which they are connected.

\mypara{Deepest may not be best.}
The layer with the most useful features depends on the pretraining and downstream tasks.
For example, the abstraction level can increase, peak, then decrease across layers for ViTs pretrained to predict masked \emph{pixels} \citep{park2023what}, in which case we may want to target a specific hidden layer to maximize the abstraction level of our target.
And even if the final layer is the most abstract, mid-level embeddings are more useful for many tasks \citep{bolya2025perception, evci2022headtoe}.
All of these studies consider \emph{trained} models, and their findings can directly inform distillation from a frozen teacher \citep{liu2024dbot, li2025salt}.
Conversely, we pretrain from scratch via self-distillation, which makes choosing the target layer more complex: the optimal choice may change throughout pretraining and create feedback effects that influence which layers become most useful.
To avoid committing to a specific hidden layer, our Bootleg predictor simultaneously targets early, mid, and deep embeddings, ``using the whole buffalo''.

\mypara{Stability.}
Self-distillation methods are known to be unstable, whereas MAE is reliable because it targets pixels, which prevent destructive feedback.
For example, slight changes to I-JEPA's masking strategy result in ${>}10$\% decreases in accuracy \citep{assran2023ijepa}.
By targeting the teacher's embeddings at various depths (\eg{} early, middle, and late), we include targets that are minimally processed (\ie{}, by fewer layers) and hence are more stimulus-driven, providing grounding to the training objective, which helps prevent degenerate solutions to the task.

\mypara{Information bottleneck.}
Our training objective can be viewed through an information compression lens, with information passed through a bottleneck between the encoder and the predictor.
We task the model with predicting latent representations of the stimulus at multiple levels (\eg{} $|L|=4$) of abstraction, with a total of $|L| \! \times\! |M| \!\times\! D$ target elements, but the encoder must pass this information to the predictor through a bottleneck sized $1 \!\times\! |N| \!\times\! D_\text{pred}$, with $|N| \!<\! |M|$ and $D_\text{pred} < D$ since the majority of tokens are masked out and the predictor network is narrower than the encoder.
By increasing $\!|L|$, we increase the bottleneck ratio and hence increase the compression it must perform, which is only possible if it understands the stimulus sufficiently well \citep{tishby2000informationbottleneckmethod}.
Moreover, there can be redundancy across embedding dimension $D$, sequence dimension $M$ and target layers $L$.
Intuitively, such correlation can cause collapse during training if entropy of the targets is smaller than the size of the bottleneck, but (1) there is less correlation across sequence dimension for early layers \citep{park2023what, Raghu2021doViTsSeeCNNs}
and (2)
by training the model to self-distil multiple levels of abstraction into its output layer, we ask it to compress information from all these levels into the space of a single layer, repeatedly.

\subsection{Illustrative experiments}
\label{sec:bridging-exps}

As a proof-of-concept, we explore bridging MAE (which predicts pixels) and I-JEPA (which predicts the final layer) by sweeping the target from the input layer to the output layer.
\cref{fig:stabilizing} shows the results of training ViT-S with a basic, \ie{}, ablated form of Bootleg: I-JEPA where the target is a layer of intermediate abstraction (a hidden layer within the teacher-encoder). We observe that it is superior to using the image or final embedding as the target, with the best self-distillation target being the output of the second layer of the teacher (left, blue line).
A similar trend is seen for ViT-B (middle), where the best single target is the 6th layer. %
In both cases, the performance falls for deeper layers before rising again when the target is the output layer (12th layer).
Furthermore, multiple targets (a hidden layer of the teacher-encoder concatenated with its final layer), even with the same predictor capacity, yields superior performance to using a single hidden target (left, red line).
Notably, for ViT-S the 6th layer is the worst single target yet the best partner for the final layer: we hypothesise that alone it is neither grounded (like early layers) nor abstract (like the final layer), but once the final layer enforces abstraction, a mid-depth target---the most decorrelated from the output (\cref{sec:inter-layer-sim})---best keeps information flowing through the full depth of the encoder (per-block analysis in \cref{fig:knn-fig2-targets}).

When we train with MAE but change the prediction target to a hidden layer, we find that performance decreases, with the training instabilities causing a complete collapse in the model if the target is in the 4th layer or deeper (right, blue line).
This issue is alleviated by predicting both the image pixels and a hidden layer (green line), which prevents the instabilities and model collapse. However, this is no better at the downstream task than predicting only pixels.
Furthermore, we find that using the image pixels as one type of target amongst multiple levels of representation (\ie{} when the weighting on the pixel reconstruction loss term is less than a quarter of the total loss) is insufficient to prevent collapse during training (not shown).
When we change the masking strategy from uniform random to either I-JEPA's masks (not shown) or Bootleg's masks (cyan line), this sole change to the MAE is sufficient to stabilize training with hidden-self-distillation, and we observe a trend that is similar to the hidden-self-distillation version of I-JEPA (left).

\section{Method}
\label{s:method}

We now describe %
our hidden-self-distillation training method, Bootleg, illustrated in \cref{fig:method}.

\mypara{Objective.} Given ``seen'' patches of an image, create high-level representations of these patches to predict the missing representations of ``masked'' patches, with targets selected from multiple levels of the layer hierarchy.
Our methodology is most similar to I-JEPA \citep{assran2023ijepa}, with the main conceptual difference being the expansion of the targets to multiple hidden layers, instead of only the final layer.
In full, Bootleg differs from I-JEPA in:
(i)~its targets, the standardized outputs of every 4th block of the teacher-encoder rather than only its final output;
(ii)~its masking, which keeps I-JEPA's multi-block strategy but fixes five implementation bugs (\cref{s:masking-implementation-improvements});
(iii)~its architecture, with CLS and register tokens and a modified predictor (\cref{tab:hparams}); and
(iv)~its hyperparameters, including a longer warmup and a different EMA schedule (\cref{s:pretraining-hparams}).

\mypara{Masking.} Our masking method is based on that of I-JEPA, with technical improvements in implementation as described in \iftoggle{useappendix}{\cref{s:masking-implementation-improvements}}{the supplementary materials}, but without significant conceptual changes.
For a given image, we split it into patches (of $P\times P$ pixels), as per the ViT tokenizer.
We randomly select a size and aspect ratio of rectangles to mask out, then randomly select the placement of four rectangular masks.
These four masks, which may overlap, determine the missing representations that the student-encoder and predictor must learn to predict.
Any unmasked patches (in the complement of the union of the masks) are the input for the student-encoder, $z_0^\text{vis}$.

\mypara{Targets.}
The whole, unmasked, image is passed through a teacher-encoder.
The teacher-encoder is a ViT with its weights an EMA of the student-encoder.
For each of the masked locations, we collect the embeddings of the corresponding patch tokens outputted by certain ViT blocks within the teacher-encoder, $\bar{z}_l$ for $l \in L$.
The $|L|$ target blocks are spread evenly across the depth of the encoder from the first transformer block to the final block, so as to maximize the diversity of embeddings used as targets.
Specifically, we use every 4th block within the transformer, starting with the 1st and 4th, then 8th, \etc{} %
The embeddings for each mask location, $j$, and each embedding depth, $l$, are independently z-scored by subtracting the mean and dividing by the standard deviation over the embedding dimension, $D$; $t_{l,j}:=\operatorname{zscore}(\bar{z}_{l,j})$.
The targets for a given location $j$ are concatenated to make a single target vector of length $|L| \times D$ for each of the $M$ mask patches.

\mypara{Student-Encoder.}
The student-encoder starts by embedding all unmasked patches into tokens.
Our encoder has a further five global tokens: one class (CLS) token and four register tokens (Reg) \citep{darcet2024registers}; these are concatenated to the sequence after patch tokenization.
From the output of the encoder, $z_B$, the patch tokens and CLS token are passed to the predictor while the register tokens are discarded.

\mypara{Predictor.}
For each of the four masked regions, we take the CLS and patch embeddings from the encoder and append $|M|/4$ mask tokens.
A frozen sin-cos position embedding vector is added to all tokens.
The four mask regions are processed in parallel by the predictor, with self-attention allowing interaction between the tokens within a mask, but with no interaction between tokens in different mask regions.
Additionally, we prepend 4 predictor-register tokens---these serve a similar ``global processing'' role as the register tokens in the encoder, but have their own learnable embeddings.
We use a single predictor module to predict the concatenated target vector.

\section{Experiments}

To verify the utility of the Bootleg training procedures, we perform experiments with self-supervised training on ImageNet-1k (IN-1k) \citep{ImageNet}.
We pretrain all ViT models on 224\texttimes 224 images with a patch size of 16\texttimes 16.
Full implementation details and hyperparameters are described in \iftoggle{useappendix}{\cref{s:pretraining-hparams}}{the supplementary materials}.

\begin{table*}[tb]
\centering
\caption{
\textbf{Results for masked self-supervised learning with Bootleg and baselines.}
Top-1 accuracy (\%) on IN-1k and iNat21 with frozen probes (Lin: linear; X-Blk: attentive) and on VTAB categories (linear), and mean-IoU (\%) on ADE20K, Cityscapes, and COCO-Stuff with a bilinear decoder on the frozen encoder.
Per architecture, the first block is single-view masked SSL and the second is multi-view joint-embedding SSL (DINO~\citep{caron2021emerging}, iBOT~\citep{zhou2022image}, MoCo-v3~\citep{chen2021empirical}, SimDINOv2~\citep{wu2025simdino}), all trained on IN-1k.
{\color{gray}Grey}:~smaller patch size (14\texttimes14; 30\% more tokens).
(dst):~distilled from ViT-g.
MV:~multi-view (\cmark) or single-view (\xmark).
$^{*}$:~trained by us; other baselines use official checkpoints.
We highlight the \best{best} and \sbest{second best} within each block.
}
\label{tab:main-results}
\adjustbox{max width=\textwidth}{
{\setlength{\tabcolsep}{4pt}
\footnotesize
\begin{tabular}{llclr@{\hspace{2\tabcolsep}}cc@{\hspace{2\tabcolsep}}cc@{\hspace{2\tabcolsep}}ccc@{\hspace{2\tabcolsep}}ccc}
\toprule
 &  & &  &  & \multicolumn{2}{c}{IN-1k} & \multicolumn{2}{c}{iNat21} & \multicolumn{3}{c}{VTAB-1k linear} & \multicolumn{3}{c}{Sem.\ seg.\ lin.\ (mIoU)} \\
\cmidrule(r){6-7} \cmidrule(r){8-9} \cmidrule(r){10-12} \cmidrule{13-15}
Arch & Method & MV & Data & Ep. & Lin & X-Blk & Lin & X-Blk & Nat. & Spec. & Struc. & ADE & City & \!\!COCO\!\! \\
\midrule
ViT-S/16 & MAE$^{*}$ & \xmark & IN-1k & 800 & 49.8 & 66.4 & 26.2 & 57.5 & 46.2 & 77.3 & 39.1 & 14.1 & 24.7 & 15.6 \\
 & CrossMAE & \xmark & IN-1k & 800 & 51.8 & \sbest{68.8} & 26.7 & \sbest{59.8} & \sbest{47.9} & \sbest{77.4} & \best{41.2} & \sbest{15.2} & \sbest{25.7} & \sbest{15.8} \\
 & data2vec 2.0$^{*}$ & \xmark & IN-1k & 200 & 41.7 & 62.2 & 15.4 & 47.4 & 38.8 & 75.6 & 40.1 & \phantom{0}9.6 & 22.2 & \phantom{0}9.5 \\
 & I-JEPA$^{*}$ & \xmark & IN-1k & 600 & \sbest{52.4} & 61.9 & \sbest{26.8} & 48.4 & 40.2 & 74.3 & 37.7 & 11.8 & 19.8 & 13.8 \\
 & Bootleg (ours) & \xmark & IN-1k & 600 & \best{70.4} & \best{75.3} & \best{47.8} & \best{67.4} & \best{59.8} & \best{79.0} & \sbest{40.4} & \best{26.6} & \best{32.1} & \best{28.4} \\
\cmidrule(r){2-5}
 & DINO & \cmark & IN-1k & 800 & \sbest{76.1} & \sbest{76.3} & \best{60.9} & \sbest{68.8} & \sbest{66.8} & \sbest{81.4} & \sbest{38.1} & 20.4 & 25.4 & 20.8 \\
 & iBOT & \cmark & IN-1k & 800 & \best{77.0} & \best{77.5} & \sbest{60.8} & \best{69.7} & \best{67.6} & \best{81.7} & \best{38.9} & \best{26.9} & \best{28.8} & \best{28.8} \\
 & MoCo-v3 & \cmark & IN-1k & 300 & 73.1 & 74.6 & 52.4 & 64.5 & 64.8 & 79.7 & 37.8 & \sbest{23.4} & \sbest{25.6} & \sbest{24.5} \\
\cmidrule(r){2-5}
{\color{gray}ViT-S/14} & {\color{gray}DINOv2 (dst)} & {\color{gray}\cmark} & {\color{gray}LVD} & {\color{gray}Unk.} & {\color{gray}\best{80.6}} & {\color{gray}\best{81.5}} & {\color{gray}\best{74.0}} & {\color{gray}\best{78.9}} & {\color{gray}\best{72.1}} & {\color{gray}\best{81.6}} & {\color{gray}\best{39.3}} & {\color{gray}\best{39.7}} & {\color{gray}\best{46.4}} & {\color{gray}\best{37.5}} \\
\midrule
ViT-B/16 & MAE & \xmark & IN-1k & 1600 & \sbest{67.2} & \sbest{76.0} & \sbest{43.3} & \sbest{70.7} & 54.3 & 76.8 & 42.2 & \sbest{24.7} & 30.5 & \sbest{26.1} \\
 & CrossMAE & \xmark & IN-1k & 800 & 65.5 & 75.6 & 41.7 & 70.2 & \sbest{56.4} & \sbest{77.6} & \best{45.7} & 24.1 & \sbest{31.2} & 24.6 \\
 & data2vec 2.0 & \xmark & IN-1k & 200 & 62.2 & 73.7 & 31.3 & 61.3 & 52.6 & 76.6 & \sbest{44.9} & 22.5 & 27.8 & 22.6 \\
 & I-JEPA$^{*}$ & \xmark & IN-1k & 600 & 67.0 & 72.4 & 41.4 & 63.0 & 50.4 & 77.2 & 39.8 & 19.3 & 24.3 & 20.4 \\
 & Bootleg (ours) & \xmark & IN-1k & 600 & \best{76.7} & \best{79.2} & \best{58.3} & \best{74.2} & \best{61.5} & \best{82.1} & 39.8 & \best{30.9} & \best{35.9} & \best{32.2} \\
\cmidrule(r){2-5}
 & DINO & \cmark & IN-1k & 400 & 77.5 & 77.7 & \sbest{63.7} & 70.1 & \sbest{69.4} & \sbest{82.3} & 39.0 & 27.1 & 30.5 & 27.8 \\
 & iBOT & \cmark & IN-1k & 400 & \sbest{78.8} & \best{79.2} & 63.4 & \sbest{72.5} & 69.3 & 82.0 & \sbest{40.2} & \best{32.6} & \best{33.8} & \best{33.2} \\
 & MoCo-v3 & \cmark & IN-1k & 300 & 76.2 & 77.3 & 57.3 & 68.4 & 67.1 & \best{82.8} & 37.7 & 29.1 & 27.7 & 29.6 \\
 & SimDINOv2 & \cmark & IN-1k & 100 & \best{79.3} & \best{79.2} & \best{66.3} & \best{73.0} & \best{69.7} & 81.7 & \best{41.6} & \sbest{31.4} & \sbest{32.7} & \sbest{32.0} \\
\cmidrule(r){2-5}
{\color{gray}ViT-B/14} & {\color{gray}Franca} & {\color{gray}\cmark} & {\color{gray}IN-22k} & {\color{gray}(90)} & {\color{gray}\sbest{81.7}} & {\color{gray}\sbest{83.0}} & {\color{gray}\sbest{73.6}} & {\color{gray}\sbest{81.0}} & {\color{gray}\sbest{71.3}} & {\color{gray}\sbest{82.1}} & {\color{gray}\sbest{38.6}} & {\color{gray}\sbest{38.0}} & {\color{gray}\sbest{46.1}} & {\color{gray}\sbest{38.4}} \\
{\color{gray}} & {\color{gray}DINOv2 (dst)} & {\color{gray}\cmark} & {\color{gray}LVD} & {\color{gray}Unk.} & {\color{gray}\best{83.7}} & {\color{gray}\best{84.2}} & {\color{gray}\best{80.4}} & {\color{gray}\best{84.3}} & {\color{gray}\best{74.6}} & {\color{gray}\best{83.1}} & {\color{gray}\best{41.3}} & {\color{gray}\best{42.8}} & {\color{gray}\best{49.6}} & {\color{gray}\best{40.1}} \\
\midrule
ViT-L/16 & MAE & \xmark & IN-1k & 1600 & \sbest{75.5} & 79.5 & \sbest{51.8} & \sbest{76.3} & \sbest{60.1} & 80.2 & 45.4 & 28.8 & 34.7 & \sbest{28.8} \\
 & CrossMAE & \xmark & IN-1k & 800 & 71.8 & 78.7 & 50.1 & 75.1 & 59.1 & \sbest{80.5} & \best{46.5} & 28.8 & \sbest{36.7} & \sbest{28.8} \\
 & data2vec 2.0 & \xmark & IN-1k & 200 & 72.7 & \sbest{80.0} & 41.5 & 72.7 & 51.9 & 77.8 & \best{46.5} & \sbest{30.0} & 35.5 & 27.7 \\
 & I-JEPA$^{*}$ & \xmark & IN-1k & 600 & 68.4 & 72.3 & 41.1 & 61.3 & 54.0 & 78.4 & 38.0 & 21.8 & 25.5 & 23.6 \\
 & Bootleg (ours) & \xmark & IN-1k & 600 & \best{79.1} & \best{80.6} & \best{61.2} & \best{77.1} & \best{64.9} & \best{81.9} & 41.5 & \best{34.7} & \best{39.1} & \best{34.4} \\
\cmidrule(r){2-5}
 & iBOT & \cmark & IN-1k & 250 & \sbest{80.3} & \sbest{79.2} & \sbest{66.5} & \sbest{74.3} & \sbest{68.8} & \sbest{81.3} & \sbest{40.0} & \sbest{33.2} & \sbest{35.0} & \sbest{34.0} \\
 & SimDINOv2 & \cmark & IN-1k & 100 & \best{81.9} & \best{81.6} & \best{69.0} & \best{77.1} & \best{70.1} & \best{83.1} & \best{42.0} & \best{36.6} & \best{37.5} & \best{36.4} \\
\cmidrule(r){2-5}
{\color{gray}ViT-L/14} & {\color{gray}CAPI} & {\color{gray}\xmark} & {\color{gray}IN-1k} & {\color{gray}6394} & {\color{gray}77.6} & {\color{gray}83.2} & {\color{gray}47.6} & {\color{gray}80.7} & {\color{gray}62.4} & {\color{gray}80.8} & {\color{gray}\best{44.9}} & {\color{gray}38.9} & {\color{gray}43.7} & {\color{gray}37.4} \\
{\color{gray}} & {\color{gray}Franca} & {\color{gray}\cmark} & {\color{gray}LAION} & {\color{gray}(3.2)} & {\color{gray}\sbest{83.4}} & {\color{gray}\sbest{84.4}} & {\color{gray}\sbest{77.0}} & {\color{gray}\sbest{86.2}} & {\color{gray}\sbest{71.2}} & {\color{gray}\sbest{82.2}} & {\color{gray}38.9} & {\color{gray}\sbest{41.7}} & {\color{gray}\sbest{48.1}} & {\color{gray}\best{40.3}} \\
{\color{gray}} & {\color{gray}DINOv2 (dst)} & {\color{gray}\cmark} & {\color{gray}LVD} & {\color{gray}Unk.} & {\color{gray}\best{85.4}} & {\color{gray}\best{85.9}} & {\color{gray}\best{83.2}} & {\color{gray}\best{87.3}} & {\color{gray}\best{75.5}} & {\color{gray}\best{84.4}} & {\color{gray}\sbest{42.5}} & {\color{gray}\best{43.4}} & {\color{gray}\best{50.9}} & {\color{gray}\sbest{40.2}} \\
\bottomrule
\end{tabular}
}
}
\end{table*}

\subsection{Image classification}
\label{s:exp:in1k}

After SSL pretraining of the ViT model with Bootleg, we evaluate the model's performance at supervised classification of IN-1k and iNaturalist-2021 (iNat21) images.
iNaturalist \citep{Horn2018_iNaturalist} is a long-tailed, fine-grained classification task with domain shift from IN-1k. %
We report the linear probe performance as the maximum of patch average or CLS; for the methodology, see \cref{a:eval-method-classification-probe}.
We also report linear probe performance on VTAB \citep{Zhai2019_VTAB}, a benchmark comprised of 19 diverse visual classification tasks in three categories: Natural (set in the natural world), Specialized (domain-specific), and Structured (tasks requiring geometric or spatial reasoning, using synthetic images).

\mypara{Results.}
Our results, shown in \cref{tab:main-results}, demonstrate the effectiveness of Bootleg for image classification.
Across all model sizes, ViTs pretrained with Bootleg outperform the within-category competitors. %
This demonstrates the effectiveness of the pretraining methodology at learning to represent high-level image features.
The largest margins ($>\!10\%$) were seen at the smallest model sizes.
Bootleg's advantage over I-JEPA also holds (+7.4--12.5\,pp) under I-JEPA's own linear protocol (\cref{s:ijepa-linprobe}), which reproduces published MAE and I-JEPA ViT-H/14 results; our I-JEPA ViT-B/L reproductions are 3.4--7.8\,pp below published values\footnote{Lacking official model checkpoints and training configs, the reason for the unreproducibility of ViT-B/L results is unclear.} (72.9\%, 77.5\%), which Bootleg still exceeds.
The best IN-1k multi-view methods lead on linear probes (by 2.6--6.6\,pp), but Bootleg is within 2.2\,pp with attentive probes and competitive on segmentation (see below).
On VTAB, our results show Bootleg models excel at representing naturalistic images (corresponding to the IN-1k pretraining domain) and specialized imagery (\eg{} satellite and medical images), but are less performant on synthetic images and counting tasks.

\subsection{Semantic segmentation}

To check the performance of ViTs pretrained with Bootleg on standard dense-prediction tasks, we run experiments on the ADE20K \citep{ade20k}, Cityscapes \citep{Cordts2016_Cityscapes}, and COCO-Stuff \citep{caesar2018_COCOStuff} semantic segmentation benchmarks.
For full implementation details, see \iftoggle{useappendix}{\cref{s:method-semseg}}{the supplementary materials}.

\mypara{Results.}
Bootleg performs competitively at semantic segmentation, as shown in \cref{tab:main-results}.
The frozen probes show Bootleg is much better able to perform pixel-level segmentation than its self-distillation competitor, I-JEPA, demonstrating that distillation of features along the visual pathway facilitates the model's ability to perform at a range of levels of granularity.
PCA visualizations of the patch features (\cref{s:feature-pca}) show that Bootleg separates objects from the background at all depths.

\subsection{Fine-tuning}

We evaluated the pretrained models with fine-tuning on IN-1k and ADE20K, as described in \cref{s:results-ft}.
Bootleg performed best for full fine-tuning on IN-1k with 100\% or 1\% (low-shot) of the training data, although the margins are smaller than for probing because the models are trained further using a strong training methodology optimized for MAE.
For LoRA fine-tuning on IN-1k and full finetuning on ADE20K, Bootleg performs best for ViT-S and B, and data2vec best for ViT-L.

\section{Ablations}
\label{s:ablations}

\subsection{Choice of target layers}
\label{s:ablation:target}

The core mechanism of Bootleg is self-distillation of hidden layers---which poses the question of which hidden layers to distil.
We explore the performance of ViT-S, -B, -L with different hidden targets for self-distillation during pretraining. %
In addition to predicting the output of various hidden blocks, $z_l$, we explore performance when predicting  mid-block representations, $z_l^\text{mid}$, and  residual terms $r_l^\text{Attn}$ and $r_l^\text{MLP}$ (tabulated in \iftoggle{useappendix}{\cref{s:target-choice}}{supplementary materials}).
For ViT-S, we find two peaks in performance: one when using 3--4 equispaced block outputs as targets, and another when targeting multiple layers within every block.
However, targeting residual layers was only sometimes beneficial, and only seen for ViT-S.
Overall, targeting the output of every fourth block in the teacher is a rule-of-thumb which gives consistently strong performance, and often the best performance of the configurations tried.

In addition to investigating how many hidden layers to target and their types,
we consider where the targets should be taken from.
As shown in \cref{tab:change-target-group}, we found using four spaced-out targets outperforms using consecutive targets from the start, middle, or final blocks of the teacher.
Data2vec uses an average of the final 10 layers as its target, implicitly assuming the training signal across multiple hidden layers can be linearly integrated together.
We, on the other hand, use a concatenation of the latent embeddings from multiple blocks as the distillation target, treating latents as distinct representations of intermediate abstraction.
Using multiple layers of abstraction as distillation training targets improves performance over distilling a single target (\cref{tab:change-target-group}).

\begin{table*}[tbh]
\centering
\caption{
\textbf{Target construction strategy.}
We compare four spaced-out targets (ours) to consecutive targets from early, middle, or final network layers.
We also consider two options of how to merge targets: either standardize and concatenate (ours) or standardize, average, and re-standardize (data2vec-style).
Spaced-out, concatenated, targets yield the best training targets.
We also show the final layer alone, and pixels with the final layer (concatenated, or with a separate decoder per target).
Experiments trained for 300 ep. on IN-1k (based on our final training recipe; see \iftoggle{useappendix}{\cref{s:pretraining-hparams}}{supplement}) and evaluated with frozen probe.
}
\label{tab:change-target-group}
\footnotesize
\adjustbox{max width=\textwidth}{
\begin{tabular}{llcccccc}
\toprule
 &  & \multicolumn{3}{c}{IN-1k acc.} & \multicolumn{3}{c}{ADE20K mIoU} \\
\cmidrule(r){3-5} \cmidrule(r){6-8}
Targets & Merge op. & Patch & CLS & X-Blk & kNN & Lin & Blk \\
\midrule
Pixel targets (MAE-style)                & \na{}   & 52.9 & 51.0 & 66.7 & 13.6 & 17.7 & 27.5 \\
Block 12 (final only)                    & \na{}   & 56.7 & 54.6 & 68.6 & 13.1 & 18.8 & 26.8 \\
Pixels + block 12                        & 2 decoders & 58.5 & 58.3 & 70.1 & 16.4 & 21.1 & 31.0 \\
Pixels + block 12                        & Concat  & 59.0 & 59.6 & 70.0 & 17.2 & 21.7 & 31.1 \\
Blocks 1--4 (early only)                 & Concat  & 58.0 & 57.1 & 69.5 & 16.5 & 21.0 & 30.1 \\
Blocks 5--8 (mid only)                   & Concat  & 61.6 & 61.7 & 72.2 & 18.6 & 23.0 & 33.0 \\
Blocks 3--12 (last 10)                   & Concat  & \sbest{64.6} & \sbest{66.0} & \sbest{73.4} & \sbest{20.9} & \sbest{24.5} & \sbest{33.4} \\
Blocks 9--12 (late only)                 & Concat  & 60.2 & 60.8 & 69.7 & 16.0 & 19.7 & 29.3 \\
\rowcolor{vlightgray}
Blocks $\{1, 4, 8, 12\}$ (spaced)        & Concat  & \best{67.8} & \best{68.9} & \best{74.4} & \best{22.1} & \best{26.6} & \best{34.2} \\
Blocks 3--12 (last 10; d2v-style)        & Average & 44.1 & 37.9 & 59.9 & \phantom{0}8.8 & 12.8 & 24.2 \\
Blocks $\{1, 4, 8, 12\}$ (spaced) & Average & 57.2 & 54.6 & 69.8 & 15.6 & 19.8 & 30.5 \\
\bottomrule
\end{tabular}
}
\end{table*}

\subsection{Bootleg editions of other masking methods}
\label{s:ablation:bootleg-others}

We explore the effect of adding our main conceptual contribution---self-distillation of hidden layers---to other masked image modelling methods with minimal other changes (\cref{tab:add-targets-to-others}).
We train a ViT-S for 300 epochs on IN-1k using either the MAE, CrossMAE, or I-JEPA framework, then evaluate with frozen probes on IN-1k, as described in \cref{a:eval-method}.

\mypara{Adding EMA.} To facilitate self-distillation, we need to add an EMA teacher to MAE and CrossMAE; we first evaluate the performance of the EMA encoder without using it for distillation.
Its performance is negligibly different from that of the encoder without EMA.

\mypara{Masking.} As we show in \cref{sec:bridging-exps}, the masking strategy of MAE is a limiting factor for the adoption of self-distillation of targets deeper than Block 2.
We thus change its masking strategy to our implementation of I-JEPA masking, which improved performance.
For MAE, we also apply the predictor once per mask region, allowing self-attention interactions within the tokens of each region.
For CrossMAE, the use of cross- instead of self-attention means there is no interaction between the mask tokens, and so this is not a consideration; we use the concatenation of the mask tokens from all four regions for efficiency\footnote{Note that the overlap of mask regions means there are redundant mask tokens for the same patch location for the CrossMAE predictor; these serve only to increase the loss weighting on resampled mask locations, and for consistency we did not deduplicate these.}.

\mypara{Hidden-self-distillation.} With these changes in place, we next change the target from the image pixels (MAE and CrossMAE), and final layer (I-JEPA), to our target: a concatenation of hidden latent vectors.
This greatly increases performance for MAE ($+11.2$, $+12.2$, and $+6.1$\,pp on the Patch, CLS, and X-Blk probes), CrossMAE ($+8.3$, $+6.3$, and $+4.0$\,pp), and I-JEPA ($+6.1$ and $+7.0$\,pp on Patch and X-Blk).
These findings indicate masked image modelling can in general benefit from using our hidden-self-distillation, provided the masking strategy supports it.\looseness=-1

\begin{table}[tbh]
\centering
\caption{%
\textbf{Effect of combining key Bootleg components with existing pretraining methods.}
Experiments conducted with ViT-S, 300 ep., and evaluated with frozen probe on IN-1k (top-1 acc, \%) and ADE20K (mIoU, \%).
Modifications are cumulative.
See \cref{s:ablation:bootleg-others} for details.
}
\label{tab:add-targets-to-others}
\footnotesize
\adjustbox{max width=\textwidth}{
\def\tablespec{ll@{\hspace{2\tabcolsep}}r@{\hspace{\tabcolsep}}r@{\hspace{2\tabcolsep}}r@{\hspace{\tabcolsep}}r@{\hspace{2\tabcolsep}}r@{\hspace{\tabcolsep}}r@{\hspace{3\tabcolsep}}r@{\hspace{\tabcolsep}}r@{\hspace{2\tabcolsep}}r@{\hspace{\tabcolsep}}r}
\expandafter\tabular\expandafter{\tablespec}
\toprule
        &                     & \multicolumn{6}{c}{IN-1k acc.} & \multicolumn{4}{c}{ADE20K mIoU} \\
\cmidrule(r){3-8} \cmidrule{9-12}
Base    & Modification        & \multicolumn{2}{l}{Patch} & \multicolumn{2}{l}{CLS} & \multicolumn{2}{l}{X-Blk} & \multicolumn{2}{l}{kNN} & \multicolumn{2}{l}{Lin} \\
\midrule
MAE      & Unmodified           & 44.7 &              & 47.0 &              & 65.1 &              & 9.3 &              & 13.1 &              \\
         & +EMA                 & 44.8 & \aincr{+0.1} & 46.9 & \adecr{-0.1} & 65.1 & \adecr{-0.0} & 9.1 & \adecr{-0.2} & 13.1 & \adecr{-0.0} \\
         & +our masking         & 51.2 & \aincr{+6.5} & 50.7 & \aincr{+3.8} & 66.0 & \aincr{+0.9} & 12.2 & \aincr{+3.1} & 16.6 & \aincr{+3.5} \\
         & +change targets      & 62.4 & \aincr{+11.2} & 62.9 & \aincr{+12.2} & 72.1 & \aincr{+6.1} & 19.5 & \aincr{+7.3} & 24.1 & \aincr{+7.4} \\
\addlinespace
CrossMAE & Unmodified           & 47.6 &              & 48.3 &              & 66.4 &              & 9.4 &              & 14.2 &              \\
         & +EMA                 & 47.6 & \adecr{-0.0} & 48.3 & \adecr{-0.0} & 66.6 & \aincr{+0.1} & 9.4 & \aincr{+0.0} & 14.2 & \adecr{-0.0} \\
         & +our masking         & 51.9 & \aincr{+4.3} & 50.9 & \aincr{+2.7} & 66.6 & \aincr{+0.1} & 11.8 & \aincr{+2.5} & 17.0 & \aincr{+2.8} \\
         & +change targets      & 60.2 & \aincr{+8.3} & 57.3 & \aincr{+6.3} & 70.7 & \aincr{+4.0} & 15.3 & \aincr{+3.5} & 21.3 & \aincr{+4.2} \\
\addlinespace
I-JEPA   & Unmodified           & 53.4 &              & \na{} &              & 62.3 &              & 10.5 &              & 13.5 &              \\
         & +change targets      & 59.5 & \aincr{+6.1} & \na{} &              & 69.3 & \aincr{+7.0} & 15.1 & \aincr{+4.6} & 19.3 & \aincr{+5.7} \\
\bottomrule
\endtabular
}
\end{table}

\subsection{Interpolating between I-JEPA and Bootleg}

We carefully identify and evaluate the differences between Bootleg and the closest existing masked image modelling method: I-JEPA.
As shown in \cref{tab:config-interpolation}, we find changing the targets to be the outputs of multiple blocks is the largest single improvement made to the I-JEPA training configuration, followed by improving the implementation of the masking strategy and the addition of a CLS token.
Averaged over adding each change first or last, masking and targets tie (+3.6\,pp; \cref{s:config-interpolation}).

\begin{table*}[tbh]
\centering
\caption{
\textbf{Ablation of the differences between I-JEPA and Bootleg.}
Each row isolates a single change to the training configuration.
We show the performance of a model trained with a config equal to I-JEPA plus one change from Bootleg (``Only with''), and with a config equivalent to changing everything from the I-JEPA to Bootleg config except for one component (``Only without'').
Deltas show the effect of adding the component first (to I-JEPA) or last (to reach Bootleg).
Experiments performed with ViT-S, 300 ep.
}
\label{tab:config-interpolation}
\small
\adjustbox{max width=\textwidth}{
\begin{tabular}{lrlrlrlrlc}
\toprule
           & \multicolumn{4}{c}{IN-1k acc., X-Blk probe} & \multicolumn{4}{c}{ADE mIoU, Lin probe} &  \\
\cmidrule(r){2-5}\cmidrule(r){6-9}
Ablation  & \multicolumn{2}{c}{Only with} & \multicolumn{2}{c}{Only without} & \multicolumn{2}{c}{Only with} & \multicolumn{2}{c}{Only without} & Avg $\Delta$ \\
\midrule
I-JEPA (= with none) & 62.3 &             &      &             & 13.5 &             &      &             & \\
\midrule
Data transforms: +hflip, $\uparrow$min crop size & 63.2 & \incr{+0.8} & 74.2 & \incr{+0.2} & 13.3 & \decr{-0.2} & 25.8 & \incr{+0.8} & \incr{+0.4} \\
Add CLS token        & 65.9 & \incr{+3.5} & 74.1 & \incr{+0.3} & 17.4 & \incr{+3.9} & 25.6 & \incr{+1.0} & \incr{+2.2} \\
Add 4/4 reg tokens to encoder/predictor & 64.7 & \incr{+2.4} & 73.9 & \incr{+0.5} & 15.5 & \incr{+1.9} & 25.6 & \incr{+1.0} & \incr{+1.4} \\
Predictor size: $\uparrow$ depth, $\uparrow$ heads & 64.1 & \incr{+1.8} & 72.5 & \incr{+1.9} & 13.9 & \incr{+0.3} & 23.6 & \incr{+3.0} & \incr{+1.8} \\
Masking: implementation improvements & 68.2 & \incr{+5.9} & 72.6 & \incr{+1.8} & 18.0 & \incr{+4.5} & 24.5 & \incr{+2.2} & \incr{+3.6} \\
Targets: $T=\{12\} \rightarrow \{1, 4, 8, 12\}$ & 69.3 & \incr{+7.0} & 73.9 & \incr{+0.4} & 19.3 & \incr{+5.7} & 25.2 & \incr{+1.4} & \incr{+3.6} \\
Loss: Smooth L1 $\rightarrow$ L2 & 65.1 & \incr{+2.7} & 74.5 & \decr{-0.1} & 14.9 & \incr{+1.3} & 25.5 & \incr{+1.1} & \incr{+1.3} \\
Hyperparams: $\uparrow$ LR, $\uparrow$ WU, $\downarrow$ WD & 63.7 & \incr{+1.3} & 73.1 & \incr{+1.2} & 13.0 & \decr{-0.5} & 24.1 & \incr{+2.5} & \incr{+1.2} \\
\midrule
Bootleg (= with all) &      &             & 74.4 &             &      &             & 26.6 &             & \\
\bottomrule
\end{tabular}
}
\end{table*}

\begin{table*}[tbh]
\centering
\caption{
\textbf{Changing masking strategy.}
Experiments trained for 300 ep. on IN-1k whilst using different masking strategies. Evaluated with frozen probe on IN-1k and ADE20K.
}
\label{tab:config-mask-strat}
\footnotesize
\adjustbox{max width=\textwidth}{
\begin{tabular}{@{}lllcccccc@{}}
\toprule
 & &  & \multicolumn{3}{c}{IN-1k acc.} & \multicolumn{3}{c}{ADE20K mIoU} \\
\cmidrule(r){4-6} \cmidrule(r){7-9}
Masking strategy & Seen (\%) & Target (\%) & Patch & CLS & X-Blk & kNN & Lin & Blk \\
\midrule
Random (MAE-style)                  & 25.0 & 75.0 & 32.3 & 28.8 & 54.4 & \phantom{0}5.0 & \phantom{0}9.5 & 21.4 \\
Green noise (ColorMAE-style)        & 25.0 & 75.0 & 35.9 & 18.4 & 57.6 & \phantom{0}6.9 & 10.3 & 23.2 \\
Inverse block (data2vec-style; 1 rep.) & 20.0 & 80.0 & 58.0 & 53.6 & 69.5 & 16.3 & 21.5 & 30.2 \\
Inverse block (data2vec-style; 4 rep.) & 20.0 & 80.0 & 59.2 & 53.1 & 70.1 & 16.9 & 21.1 & 31.8 \\
Cyclic block (CAPI/Franca) & 25.0 (20--30) & 75.0 (70--80) & 67.0 & 65.8 & 73.7 & 20.8 & 25.1 & 32.2 \\
Cyclic block (CAPI/Franca) & 30.0 (25--35) & 70.0 (65--75) & \best{68.0} & \sbest{67.4} & \sbest{74.0} & \sbest{21.4} & \sbest{25.8} & \sbest{32.4} \\
Multi-block (I-JEPA)                & 24.3 (\phantom{0}9--37) & 44.5 (24--64) & 65.3 & 66.5 & 72.6 & 20.6 & 24.5 & 31.7 \\
\rowcolor{vlightgray}
Multi-block (ours)                  & 29.2 (13--43) & 47.5 (26--71) & \sbest{67.8} & \best{68.9} & \best{74.4} & \best{22.1} & \best{26.6} & \best{34.2} \\
\bottomrule
\end{tabular}
}
\end{table*}

\subsection{Changing masking strategy}
\label{s:ablation:masking}

We ablated our masking strategy, using methods from recent papers. Results are shown in \cref{tab:config-mask-strat}.
When training with uniform random masks as per MAE, or ``green'' structured noise from ColorMAE \citep{hinojosa2024colormae}, final model performance was significantly impaired, but training did not collapse.
Cyclic block masking, used by CAPI and Franca \citep{darcet2025capi, venkataramanan2025franca}, %
yielded performance competitive with our multi-block implementation, and outperformed the I-JEPA implementation of multi-block.

Larger mask blocks yielded better performance, which we attribute to neighbouring patches having highly correlated activations, especially deep in the network (see \cref{sec:rep-analysis}): when target patches are adjacent to seen patches, the model exploits this as a short-cut.
This is analogous to contrastive learning, where heavy colour augmentations are needed to prevent the model using RGB-histogram correlations as a short-cut \citep{chen2020simple}.

\subsection{Training stability}
\label{s:ablation:epochs}

I-JEPA is known to saturate its performance after only 300 epochs and can become unstable and collapse \citep{assran2023ijepa}, whereas MAE is stable and continues to improve in performance given longer training \citep{he2022masked}.
We hence compared the stability of Bootleg over prolonged training with ablations targeting only the final teacher layer or the image pixels.

\begin{figure*}
    \centering%
    \includegraphics[scale=\iftoggle{arxiv}{0.41}{0.365}]{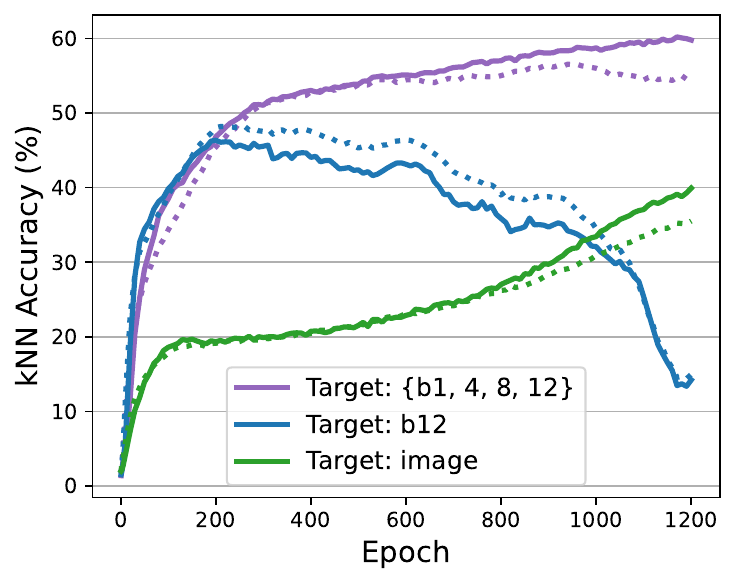}
    \includegraphics[scale=\iftoggle{arxiv}{0.41}{0.365}]{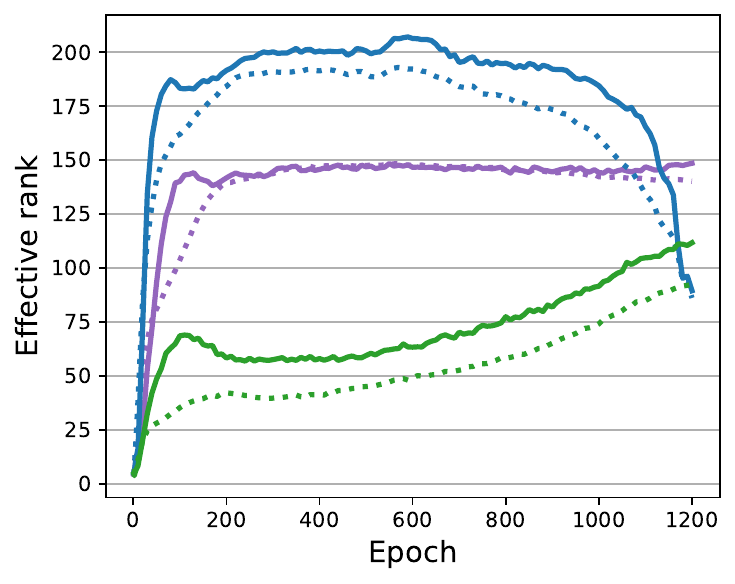}
    \includegraphics[scale=\iftoggle{arxiv}{0.41}{0.365}]{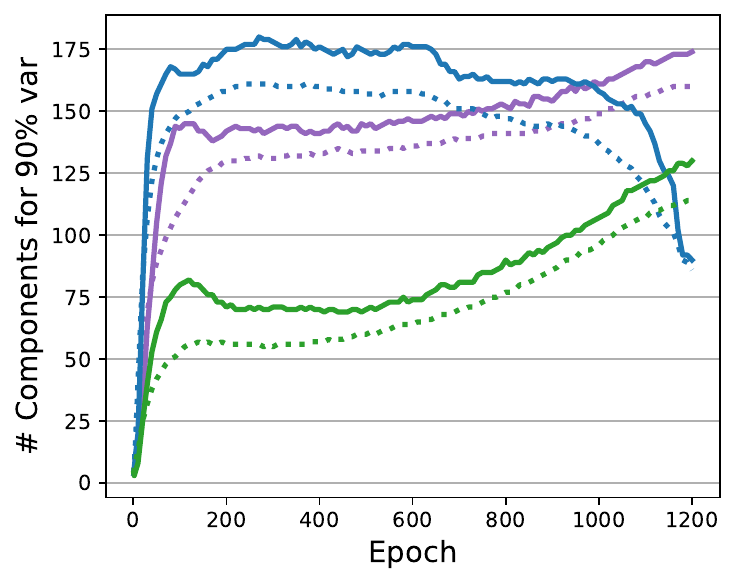}%
    \caption{%
    \textbf{Training stability of Bootleg} with multiple hidden-self-distillation targets (purple), final layer only self-distillation (blue) and image generation (green) over 1200 epochs ViT-S pretraining.
    We show the accuracy of a kNN monitor (10\% of IN-1k's training set, every 10 epochs) and the effective rank of the 384-d data estimated using the Roy--Vetterli method~\citep{roy2007effective} and a 90\% variance explained threshold.
    Solid lines: CLS token; dashed lines: patch average.
}
    \label{fig:train-curve}
\end{figure*}

As shown in \cref{fig:train-curve}, kNN performance rises quickly for self-distillation, but not for pixel generation.
Beyond 200 ep, predicting only the final layer makes kNN performance fall and the effective rank collapse, whereas predicting multiple hidden layers or the pixels gives increasing performance and stable or increasing rank.

\section{Discussion}

We presented Bootleg, a self-supervised method whose student learns to predict latent representations from multiple hidden layers of a teacher.

\mypara{Strengths.}
Hidden-self-distillation aligns embeddings with image classification without supervision, uses a per-sample objective with no cross-sample terms, and uses no domain-specific augmentations, so it should transfer to other domains and integrate into any existing JEPA framework with minimal modification.

\mypara{Limitations.}
Bootleg pays the standard teacher-encoder cost ($\sim$12\% of training time); multiple hidden targets add only $\sim$2\% as all are gathered in one teacher pass (\cref{s:no-forward-mse}).
It relies on a contiguous-block masking strategy (\cref{s:ablation:masking}); tuning masking per dataset is left to future work.
We have not validated Bootleg beyond ViT-L or on other modalities, and lack a theoretical guarantee against collapse: without its EMA teacher Bootleg still collapses (\cref{s:no-ema}), so multiple targets stabilize, but do not replace, the EMA.

\subsection*{Acknowledgements}

Thanks to Patrik Reizinger for insightful discussions.

Resources used in preparing this research were provided, in part, by the Province of Ontario, the Government of Canada through CIFAR, and \href{https://vectorinstitute.ai/partnerships/current-partners/}{companies sponsoring} the Vector Institute.
This research was enabled in part by support provided by the Digital Research Alliance of Canada.
GWT and SCL are supported by the INSPIRE (Integrated Network for the Surveillance of Pathogens: Increasing REsilience and capacity in Canada's pandemic response) project funded through the Canada Biomedical Research Fund (grant CBRF2-2023-00008), the Biomedical Research Infrastructure Fund, and the Ontario Research Fund.
GWT is also supported by the Canada Research Chairs program (grant~CRC-2021-00561).
ES is supported by a Canada CIFAR AI Chair and the Natural Sciences and Engineering Research Council of Canada (NSERC) Discovery Grant (RGPIN-2025-06878).

\subsection*{Author contributions}

Scott~C.~Lowe: conceptualization, methodology, investigation, writing (original draft), visualization.
Anthony~Fuller: methodology, writing (original draft).
Sageev~Oore: writing (review \& editing), supervision.
Evan~Shelhamer: writing (review \& editing), supervision, resources.
Graham~W.~Taylor: writing (review \& editing), supervision, resources, funding acquisition.

{
    \small
    \bibliographystyle{iclr2025_conference}
    \bibliography{main}
}

\iftoggle{arxiv}{
\clearpage
\appendix\crefalias{section}{appendix}\crefalias{subsection}{appendix}\crefalias{subsubsection}{appendix}
\section*{\Large Appendices}

{
    \hypersetup{hidelinks}%
    \startcontents[sections]
    \printcontents[sections]{l}{1}{\setcounter{tocdepth}{2}}
}
\FloatBarrier
\clearpage

\FloatBarrier
\section{Related Work}
\label{s:related-work}
We group SSL algorithms into two categories: single-view and multi-view methods.
Single-view methods, such as MAE, I-JEPA, and our Bootleg, have two key advantages: (1) they do not require hand-crafted data augmentation strategies or rules to generate pairs of ``positive'' views to promote embedding similarity, and (2) they process a single (masked) view of each image, rather than several augmented views, and do not use (explicit or implicit) ``negatives'' to promote embedding dissimilarity and avoid representational collapse.
By processing one view per image, single-view methods are more computationally accessible.
By not requiring data augmentation strategies, single-view methods can be easily adapted to other domains and input shapes (\eg{} time series, 3D point clouds, \etc{}).
We thus focus our experimental comparisons of Bootleg on comparable single-view methods, and additionally report several multi-view methods trained on IN-1k (DINO, iBOT, MoCo~v3, SimDINOv2) in \cref{tab:main-results} for reference.

\mypara{Single-view methods do \emph{not} require hand-crafted augmentations.}
The first masked modelling methods for imagery encode \emph{mask tokens} alongside patch tokens, following BERT in natural language \citep{devlin-etal-2019-bert}.
They are pretrained to predict the masked-out pixels \citep{xie2022simmim, dosovitskiy2021vit}, discrete patch tokens \citep{bao2021beit, dong2023peco}, or features \citep{zhou2022image, wei2022masked} with a small decoder.
MAE \citep{he2022masked} introduced encoding visible tokens \emph{only} (no mask tokens) and a larger decoder, which reduces computation and improves representations; this token-dropping method has been widely leveraged since, \eg{} by data2vec 2.0 \citep{baevski2023data2vec2}, I-JEPA \citep{assran2023ijepa}, and CrossMAE \citep{fu2025crossmae}.
CrossMAE replaces MAE's self-attention-based decoder with a cross-attention-based decoder for reduced pretraining computation with similar downstream accuracy.
Our Bootleg follows MAE/I-JEPA's asymmetric encoder-decoder design for efficient pretraining.

\mypara{Intermediate-layer targets and representations.}
Several works exploit intermediate layers of masked image models.
Data2vec \citep{baevski2022data2vec, baevski2023data2vec2} regresses the \emph{average} of the teacher's final $K$ layers, and found averaging multiple layers beneficial for audio but not for images; in contrast, we use concatenated targets from layers spread across the network, which we find beneficial for images (\cref{tab:change-target-group}).
MIM-Refiner \citep{alkin2025mimrefiner} observes that the strongest representations of large MIM models lie in intermediate blocks, and refines pretrained MIM models with contrastive heads attached to those blocks.
DAMIM \citep{ma2025damim} shows that the choice of reconstruction target matters for masked image modelling under domain shift.
Concurrently, V-JEPA~2.1 \citep{murlabadia2026vjepa21} predicts several intermediate teacher layers for video, alongside a dense context loss and multi-level predictor inputs; we evaluate those components with Bootleg in \cref{s:vjepa21}.
Outside vision, IA2 \citep{mishra2025ia2} aligns a language model's activations at every layer with those of the same model given in-context demonstrations: a form of multi-layer self-distillation from a teacher with access to more of the input, as in Bootleg.

\mypara{Frozen and bootstrapped teachers.}
Instead of an EMA teacher, some methods distil from a teacher that is frozen after an earlier stage of training.
dBOT \citep{liu2024dbot} performs multi-stage masked feature distillation, starting from a randomly initialized teacher and replacing the teacher with the distilled student at each stage; it finds that a careful choice of target representation is unnecessary, as different targets lead to similarly behaved models.
SALT \citep{li2025salt} first trains a video teacher by MAE-style pixel reconstruction, then freezes it and trains a student to predict the teacher's latents at masked positions, JEPA-style; it finds that student quality is robust to the quality of the teacher.
Both avoid the non-stationary targets of EMA self-distillation by holding the teacher fixed within each stage.
Bootleg instead keeps the EMA teacher and stabilizes self-distillation by targeting multiple hidden layers; these approaches are complementary, as multi-layer targets could equally be drawn from a frozen teacher.

\mypara{Multi-view methods require augmentations and multiple views.}
Canonical contrastive learning algorithms \citep{chen2020improved, chen2021empirical, chen2020simple} promote embedding similarity (for positives) and dissimilarity (for negatives) via InfoNCE-style losses \citep{oord2018representation}.
And C-JEPA \citep{mo2024connecting} adds a contrastive-loss term to I-JEPA.
iBOT \citep{zhou2022image} combines masked image modelling with multi-view self-distillation, and CAPI \citep{darcet2025capi}, although single-view, clusters masked patch representations with Sinkhorn-Knopp normalization across the batch.
Clustering methods that use Sinkhorn-Knopp \citep{Caron2020_SwAV} typically require batches that contain several thousand samples, and multi-GPU training to meet their memory requirements.
Methods such as Barlow Twins \citep{zbontar2021barlow}, VICReg \citep{bardes2021vicreg}, and MMCR \citep{yerxa2023learning}, use several hundred samples per batch, at minimum, and promote similarity between positives generated via data augmentation.
These new methods spread-out the embedding space over the batch, for example by encouraging the standard deviation of each embedding dimension---calculated per batch, and is thus batch-size dependent---to be above a threshold.
Similarly, Bootstrap Your Own Latent (BYOL, \citep{grill2020bootstrap}) was an earlier method that also requires implicit negatives to avoid collapse.
It partially inspires our Bootleg name.
State-of-the-art image SSL methods such as DINOv2 \citep{oquab2023dinov2} and Franca \citep{venkataramanan2025franca} have many self-supervised objectives, some of which also require multiple views and large batches (\eg{} 16,384).
Our Bootleg does \emph{not} require data augmentation, thus is more general and less domain-specific, and processes a single view per image, so it is computationally accessible.\looseness=-1

\section{Additional illustrative experiments}

To accompany \iftoggle{useappendix}{\cref{sec:bridging-exps}}{Section~3.2 of the main paper}, we present the zoomed out figure for the MAE config.
This figure shows the performance of ViT-S with unmodified MAE config, except instead of predicting the pixels in the input image, we predict the representation from a hidden layer within an EMA teacher (blue line).
In this zoomed out version of the plot (\cref{fig:stabilizing2}), we can see the severity of the collapse in performance when doing hidden-self-distillation of blocks 5--11 and using the unmodified MAE config.
Other work has indicated that using the L1 loss provides increased stability compared to using L2 or smooth-L1 \citep{bardes2024vjepa, assran2025vjepa2}, however we did not find evidence that this could alleviate our issue (\cref{fig:stabilizing2}, yellow line).

For ViT-B, the drop in performance when using deeper hidden layers as the distillation target was more gradual (\cref{fig:stabilizing2}, right panel).
Since ViT-B is wider than ViT-S, this, albeit weakly, suggests that a higher-dimensional target vector may help with training stability.

\begin{figure*}
    \centering
    \includegraphics[scale=0.4]{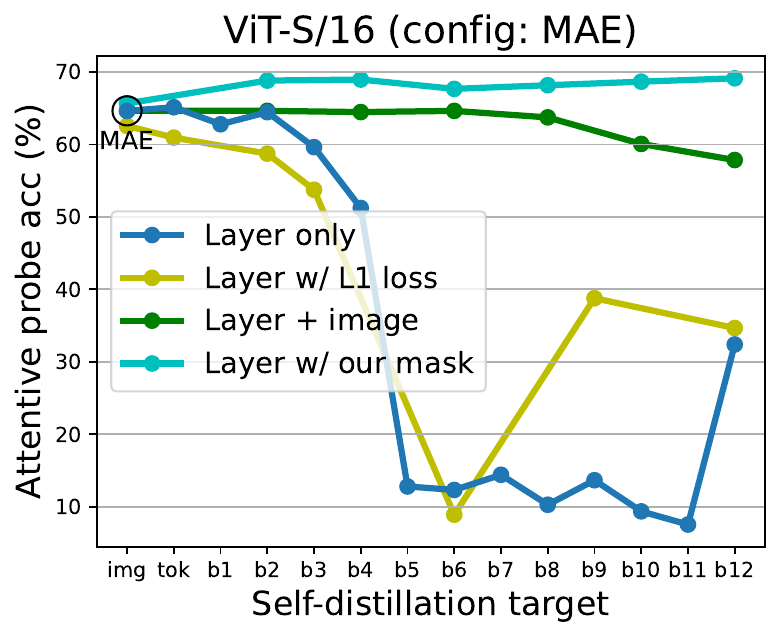} ~~
    \includegraphics[scale=0.4]{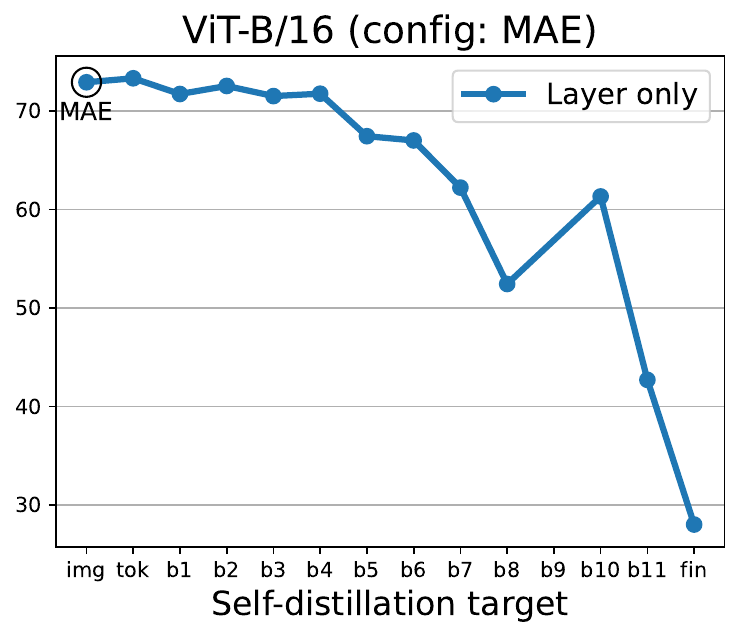}
    \caption{\textit{Accompanies Fig.~2 of the main text.}
    Bridging I-JEPA and MAE with %
    targets across hidden layers.
    \textit{Left:} As in Fig.~2 of the main text, we train ViT-S using with MAE, except we add EMA for self-distillation and change the target to be a hidden layer of the EMA model instead of the input pixels (blue) or in addition to the image pixels (green). We compare to a single target with our masking strategy (cyan), and to a single target but using the L1 loss (yellow).
    Using the L1 loss is insufficient to prevent training instabilities when using deeper layers as targets.
    \textit{Right:} Similar, but for ViT-B with MAE config.
}
    \label{fig:stabilizing2}
\end{figure*}

\section{Pretraining hyperparameters}
\label{s:pretraining-hparams}

We selected the hyperparameters for pretraining Bootleg in a multi-step process of iterative refinement.

Initially, we referred to the hyperparameters used in MAE \citep{he2022masked} and I-JEPA \citep{assran2023ijepa}.
In our preliminary experiments (\cf{} \iftoggle{useappendix}{\cref{fig:stabilizing}}{Fig.~2}, \textit{``Bridging I-JEPA and MAE with targets across hidden layers''}, of the main paper), we found training using the MAE hyperparameter configuration with the I-JEPA masking strategy and a hidden target for self-distillation outperformed similar experiments using the I-JEPA hyperparameter configuration.
We thus initialized our hyperparameter configuration with that of MAE.
Through a series of preliminary experiments conducted with ViT-S, 300 ep., we considered changing hyperparameters to that of I-JEPA or V-JEPA \citep{bardes2024vjepa}, resulting in the hyperparameters shown in \cref{tab:hparams-prelim}.
We additionally considered adding register tokens \citep{darcet2024registers} since this had been shown to be highly beneficial for networks pretrained with DINO \citep{darcet2024registers}.
We found adding (separate) registers to each of the encoder and predictor networks was beneficial. %

We then conducted a series of line searches with ViT-S, 300 ep., on the predictor size, learning rate schedule, weight decay, and number of warm-up steps.
To assess the scaling of the predictor size relative to the encoder size, we conducted a small number of experiments with ViT-B and -L encoders, varying both the width and depth of the encoder.
To assess the learning rate, which may vary depending on model size, we conducted limited experiments on ViT-B and -L using the schedule established with ViT-S.
Our final hyperparameters are shown in \cref{tab:hparams}.
We found that a larger crop size, longer LR warmup duration, and higher maximum and minimum LR values improved performance of the resulting model.
We scale the warmup duration based on the total number of epochs, $E$, as $E_\text{wu} = 33 + 0.12 \, E$.

We changed the model seed used between each round of our hparam search to avoid overfitting to a specific model.
The final evaluations shown in the main table were performed on models trained from scratch using a fresh seed which had not been included in the hyperparameter search.

\begin{table*}[tbh]
\centering
\caption{%
Hyperparameter configuration comparison between Bootleg, I-JEPA, MAE for ViT-B/16.
For all training methods, targets are z-scored in a sample-independent manner across patch and feature dimensions (implemented as a LayerNorm layer without an affine transform).
Learning rate (LR) values are tabulated for a batch size of 2048 samples, and linearly scaled to the batch size used for training.
Mask seen rate and target rate are empirically estimated (mean $\pm$ stdev) for I-JEPA and Bootleg with a batch size per GPU of 256.
}
\label{tab:hparams}
\small
\adjustbox{max width=\textwidth}{
\begin{tabular}{@{}lllll@{}}
\toprule
          & Hyperparameter    & MAE                            & I-JEPA                          & Bootleg (ours)                \\
\midrule
Encoder   & Architecture      & ViT-B/16                       & ViT-B/16                        & ViT-B/16                      \\
          & Depth             & 12                             & 12                              & 12                            \\
          & Width             & 768                            & 768                             & 768                           \\
          & Attention heads   & 12                             & 12                              & 12                            \\
          & Patch size        & 16                             & 16                              & 16                            \\
          & CLS tokens        & {1}                            & {0}                             & {1}                           \\
          & Register tokens   & {0}                            & {0}                             & {4}                           \\
\addlinespace
Predictor & Depth             & {8}                            & {6}                             & {10}                          \\
          & Width             & {512}                          & {384}                           & 384                           \\
          & Attention heads   & {16}                           & {12}                            & {16}                          \\
          & Register tokens   & {0}                            & {0}                             & {4}                           \\
\addlinespace
Data      & Transforms        & RandCrop(0.2, 1.0)\,+\,Hflip   & RandCrop(0.3, 1.0)              & RandCrop(0.35, 1.0)\,+\,Hflip \\
          & Interpolation     & Bicubic                        & Bilinear                        & Bicubic                       \\
          & Input size        & 224\texttimes 224              & 224\texttimes 224               & 224\texttimes 224             \\
\addlinespace
Masking   & Strategy          & {Uniform random}               & {4 rectangular blocks}          & {4 rectangular blocks}        \\
          & Mask seen rate    & 0.25                           & 0.243 $\pm$ 0.037               & 0.292 $\pm$ 0.069             \\
          & Mask target rate  & 0.75                           & 0.445 $\pm$ 0.070               & 0.475 $\pm$ 0.076             \\
\addlinespace
Training  & Target(s)         & Image pixels                   & Final (block 12) output         & Block 1, 4, 8, 12 outputs     \\
     & Target standardization & Z-score (patch \& feat. dims) & Z-score (patch \& feat. dims) & Z-score (patch \& feat. dims) \\
          & Target length     & 768                            & 768                             & 3072                          \\
          & Loss              & {Mean squared error}           & {Smooth L1}                     & {Mean squared error}          \\
          & Optimizer         & AdamW({\footnotesize$\beta\!=\!(0.9, 0.95)$}) & AdamW({\footnotesize$\beta\!=\!(0.9, 0.999)$}) & AdamW({\footnotesize$\beta\!=\!(0.9, 0.95)$})\\
          & {EMA initial}     & \na                            & 0.996                           & 0.9985                        \\
          & {EMA final}       & \na                            & 1.0                             & 0.9985                        \\
          & LR schedule       & Cosine annealing               & Cosine annealing                & Cosine annealing              \\
          & LR schedule warmup& 40 epochs                      & 40 epochs                       & 105 epochs                    \\
          & LR initial        & 0.0                            & 0.0002                          & 0.00003                       \\
          & LR maximum        & {0.0012}                       & {0.001}                         & {0.003}                       \\
          & LR final          & 0.0                            & 0.000001                        & 0.00003                       \\
          & WD initial        & 0.05                           & 0.04                            & 0.05                          \\
          & WD final          & {0.05}                         & {0.40}                          & {0.05}                        \\
          & Schedule stretch  & 1.0                            & 1.0                             & 1.0                           \\
          & Batch size        & 4096                           & 2048                            & 2048                          \\
          & {Num epochs}      & 1600                           & 600                             & 600                           \\
\bottomrule
\end{tabular}
}
\end{table*}

Some hyperparameters vary between model sizes.
The resulting values are shown in \cref{tab:hparam-scaling}.
We use a ViT predictor of depth 10 with width equal to half that of the encoder, with heads equal to same as the encoder with a minimum of 16 heads.
We found ViT-S needed a higher initial and final LR (1/30 of the maximum) than ViT-B and -L (1/100 of the maximum LR).

Our results include ViT-S models, which is outside the original training recipes of MAE and I-JEPA, hence we additionally share the configs we used for these in \cref{tab:hparam-scaling}.
For MAE, we follow the ViT-S recipe of \citep{fu2025crossmae}.
For I-JEPA, we halve the width of the predictor to maintain the same capacity relative to the encoder as used for ViT-B, and let the number of heads equal that of the encoder following the rule from I-JEPA \citep{assran2023ijepa}.

\begin{table*}[tbh]
\centering
\caption{%
Hyperparameter changes across encoder sizes ViT-S, -B, -L for MAE, I-JEPA, and Bootleg models.
Learning rate (LR) values are tabulated for a batch size of 2048 samples, and linearly scaled to the batch size used for training.
}
\label{tab:hparam-scaling}
\small
\adjustbox{max width=\textwidth}{
\begin{tabular}{@{}lllllllllll@{}}
\toprule
          &                   & \multicolumn{3}{c}{MAE}          & \multicolumn{3}{c}{I-JEPA}      & \multicolumn{3}{c}{Bootleg (ours)} \\
\cmidrule(r){3-5} \cmidrule(r){6-8} \cmidrule(r){9-11}
          & Hyperparameter    & ViT-S    & ViT-B    & ViT-L      & ViT-S    & ViT-B    & ViT-L     & ViT-S    & ViT-B    & ViT-L     \\
\midrule
Encoder   & Depth             & 12       & 12       & 24         & 12       & 12       & 24        & 12       & 12       & 24        \\
          & Width             & 384      & 768      & 1024       & 384      & 768      & 1024      & 384      & 768      & 1024      \\
          & Attention heads   & 6        & 12       & 16         & 6        & 12       & 16        & 6        & 12       & 16        \\
          & Patch size        & 16       & 16       & 16         & 16       & 16       & 16        & 16       & 16       & 16        \\
          & CLS tokens        & 1        & 1        & 1          & 0        & 0        & 0         & 1        & 1        & 1         \\
          & Register tokens   & 0        & 0        & 0          & 0        & 0        & 0         & 4        & 4        & 4         \\
\addlinespace
Predictor & Depth             & 8        & 8        & 8          & 6        & 6        & 12        & 10       & 10       & 10        \\
          & Width             & 256      & 512      & 512        & 192      & 384      & 384       & 192      & 384      & 512       \\
          & Attention heads   & 8        & 16       & 16         & 6        & 12       & 16        & 16       & 16       & 16        \\
          & Register tokens   & 0        & 0        & 0          & 0        & 0        & 0         & 4        & 4        & 4         \\
\addlinespace
Training  & Target(s)         & Input    & Input    & Input      & Final    & Final    & Final     & b1,4,8,12& b1,4,8,12& b1,4,8,12,16,20,24 \\
          & Target length     & 384      & 768      & 1024       & 384      & 768      & 1024      & 1536     & 3072     & 7168      \\
          & LR initial        & 0.0      & 0.0      & 0.0        & 2e-4     & 2e-4     & 2e-4      & 1e-4     & 3e-5     & 3e-5      \\
          & LR maximum        & 1.2e-3   & 1.2e-3   & 1.2e-3     & 1e-3     & 1e-3     & 1e-3      & 3e-3     & 3e-3     & 3e-3      \\
          & LR final          & 0.0      & 0.0      & 0.0        & 1e-6     & 1e-6     & 1e-6      & 1e-4     & 3e-5     & 3e-5      \\
\bottomrule
\end{tabular}
}
\end{table*}

Note that in preliminary experiments, we also considered replacing the absolute position embeddings with rotary position embeddings (RoPE) \citep{Su2024rope} as used in V-JEPA-2 \citep{assran2025vjepa2}.
We found RoPE was beneficial when not using register tokens, with a comparable effect size to adding register tokens, but using both at once was anti-synergistic.
We suspect this is because the unrotated global tokens embeddings are handled in the same way as one of the patch tokens (either the top-left corner, or the centre of the image, depending on the RoPE vector initialization strategy).
This effectively co-locates the CLS token and all four register tokens at a position within the image, potentially also colliding with one of the patch tokens, which makes it difficult for the attention heads to chose to address the register tokens.
Furthermore, since we use five global tokens and mask the patch tokens presented to the encoder, the model will see 13 times as many global tokens as patch tokens at the position for which they collide.
Since RoPE had a much higher compute burden (around +30\%) compared to adding registers (negligible), we chose to use register tokens with 2d sin-cos absolute position embeddings.
This also allows us to better compare with our benchmark models, since MAE, CrossMAE, data2vec2, and I-JEPA all use absolute position embeddings.

\FloatBarrier
\section{Masking strategy}
\label{s:masking-implementation-improvements}

As mentioned in \iftoggle{useappendix}{\cref{s:method}}{Sec.~4, \textit{``Method''}, of the main paper}, our masking strategy is conceptually similar to that of I-JEPA, with only improvements in technical details and implementation.
In this section, we describe this masking strategy, and highlight the improvements in our implementation compared to I-JEPA.

\paragraph{Overview}
The total batch size of 2048 is processed by some number of distributed GPU workers.
For example, with 8 GPU workers, each GPU worker will process 256 samples.
For a given GPU worker's batch of samples, we generate masks using the same general process for both Bootleg and I-JEPA:
\begin{enumerate}
    \item Choose a seed to use for the random number generator (RNG) for the masks in this worker's batch.
    \item Sample the height and width to use for the predictor mask rectangles: $(p_h, p_w)$. The same size is used for all four predictor mask rectangle across the worker's batch.
    \item Sample the height and width to use for the background mask: $(b_h, b_w)$. The same size is used for all background masks across the worker's batch.
    \item For each image in the batch, sample its mask positions:
    \begin{enumerate}
        \item For each of the four predictor mask rectangles, sample their placement: $p^{(i)}_{j,x}, p^{(i)}_{j,y}$.
        \item For the background mask, sample its placement: $b^{(i)}_x, b^{(i)}_y$.
        \item Let the visible mask be the intersection between the background mask and the complement of the predictor mask rectangles: $v^{(i)} := b^{(i)} \setminus (p^{(i)}_1 \cup p^{(i)}_2 \cup p^{(i)}_3 \cup p^{(i)}_4)$
    \end{enumerate}
    \item For GPU efficiency, all visible masks need to be the same length. Truncate all the visible masks to the length of the shortest visible mask within the GPU worker's batch.
\end{enumerate}

Note that we need to truncate the visible masks because the four predictor mask rectangles will overlap by a random amount.
For some samples in the worker's batch, there will be very little overlap, resulting in the visible mask having fewer elements.
Whereas for other samples in the worker's batch, there will be a lot of overlap (potentially all four predictor mask rectangles overlap entirely), which would mean more visible tokens for these samples without truncation.
The predictor mask rectangles are always the same size within the worker's batch and so do not need truncation to have the same length.

As presented below, we make technical improvements to \textit{all} steps within this pipeline.
The most significant of these improvements is in the visible token truncation strategy.

\begin{figure*}
    \centering
    \textbf{I-JEPA} \\
    \vspace{1.5mm}
    \includegraphics[scale=\iftoggle{arxiv}{0.23}{0.2}]{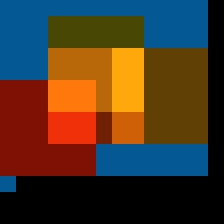}
    \includegraphics[scale=\iftoggle{arxiv}{0.23}{0.2}]{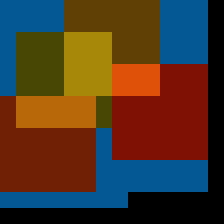}
    \includegraphics[scale=\iftoggle{arxiv}{0.23}{0.2}]{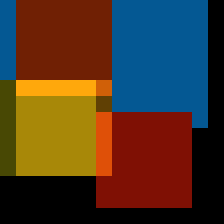}
    \includegraphics[scale=\iftoggle{arxiv}{0.23}{0.2}]{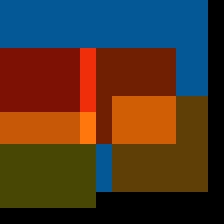}~~~~
    \includegraphics[scale=\iftoggle{arxiv}{0.23}{0.2}]{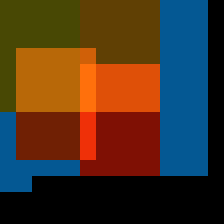}
    \includegraphics[scale=\iftoggle{arxiv}{0.23}{0.2}]{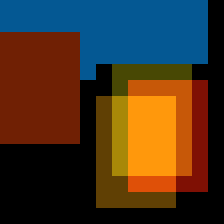}
    \includegraphics[scale=\iftoggle{arxiv}{0.23}{0.2}]{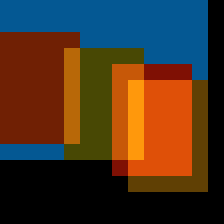}
    \includegraphics[scale=\iftoggle{arxiv}{0.23}{0.2}]{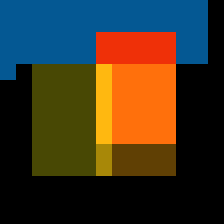}

    \vspace{2mm}

    \includegraphics[scale=\iftoggle{arxiv}{0.23}{0.2}]{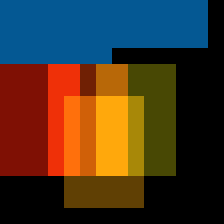}
    \includegraphics[scale=\iftoggle{arxiv}{0.23}{0.2}]{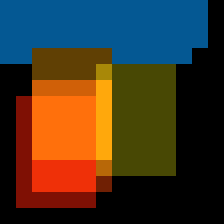}
    \includegraphics[scale=\iftoggle{arxiv}{0.23}{0.2}]{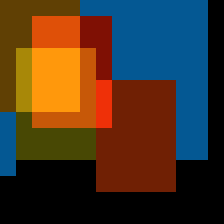}
    \includegraphics[scale=\iftoggle{arxiv}{0.23}{0.2}]{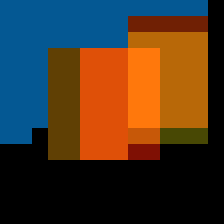}~~~~
    \includegraphics[scale=\iftoggle{arxiv}{0.23}{0.2}]{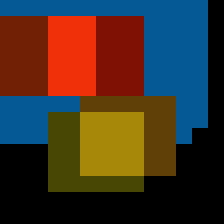}
    \includegraphics[scale=\iftoggle{arxiv}{0.23}{0.2}]{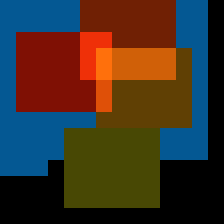}
    \includegraphics[scale=\iftoggle{arxiv}{0.23}{0.2}]{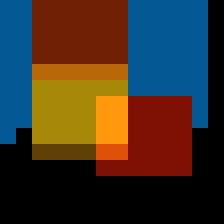}
    \includegraphics[scale=\iftoggle{arxiv}{0.23}{0.2}]{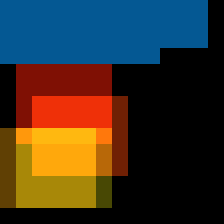}

    \vspace{2mm}
    \textbf{Bootleg} \\
    \vspace{1.5mm}
    \includegraphics[scale=\iftoggle{arxiv}{0.23}{0.2}]{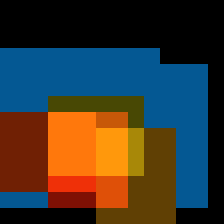}
    \includegraphics[scale=\iftoggle{arxiv}{0.23}{0.2}]{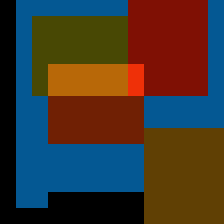}
    \includegraphics[scale=\iftoggle{arxiv}{0.23}{0.2}]{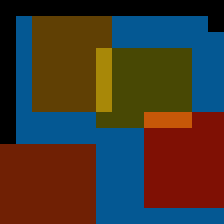}
    \includegraphics[scale=\iftoggle{arxiv}{0.23}{0.2}]{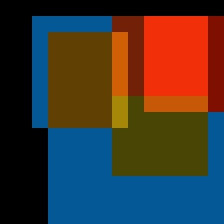}~~~~
    \includegraphics[scale=\iftoggle{arxiv}{0.23}{0.2}]{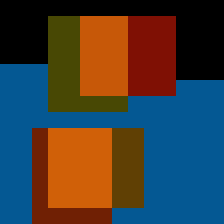}
    \includegraphics[scale=\iftoggle{arxiv}{0.23}{0.2}]{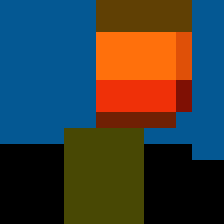}
    \includegraphics[scale=\iftoggle{arxiv}{0.23}{0.2}]{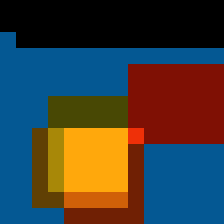}
    \includegraphics[scale=\iftoggle{arxiv}{0.23}{0.2}]{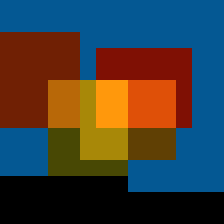}

    \vspace{2mm}

    \includegraphics[scale=\iftoggle{arxiv}{0.23}{0.2}]{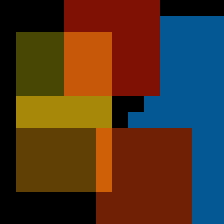}
    \includegraphics[scale=\iftoggle{arxiv}{0.23}{0.2}]{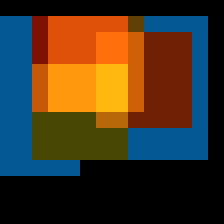}
    \includegraphics[scale=\iftoggle{arxiv}{0.23}{0.2}]{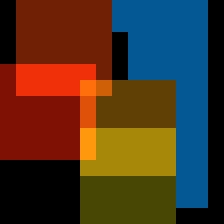}
    \includegraphics[scale=\iftoggle{arxiv}{0.23}{0.2}]{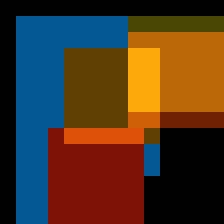}~~~~
    \includegraphics[scale=\iftoggle{arxiv}{0.23}{0.2}]{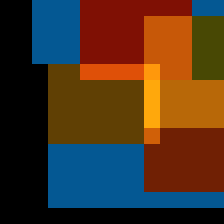}
    \includegraphics[scale=\iftoggle{arxiv}{0.23}{0.2}]{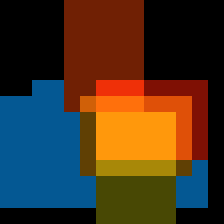}
    \includegraphics[scale=\iftoggle{arxiv}{0.23}{0.2}]{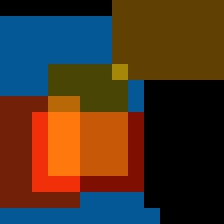}
    \includegraphics[scale=\iftoggle{arxiv}{0.23}{0.2}]{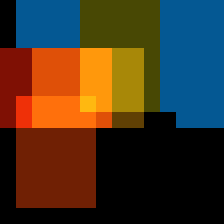}

    \caption{
Sample masks generated by I-JEPA and Bootleg's masking strategies.
Visible tokens are shown in blue.
Prediction masks are shown in green/yellow/orange/red with one colour per mask rectangle.
These prediction masks can overlap, leading to brighter shades of orange/yellow.
Unused tokens (seen by the teacher-encoder, but not by either the student-encoder or predictor) are shown in black.
For both I-JEPA and Bootleg, we show samples from worker-batches of 256 samples, generated with seeds 0,1,2,3.
We show the first 4 examples from each batch (grouped horizontally).
These samples illustrate how the visible tokens (blue) are always at the top of the image for I-JEPA's masking, but are better distributed for Bootleg's masks.
}
    \label{fig:mask-samples}
\end{figure*}

\begin{figure*}
    \centering
    \includegraphics[scale=\iftoggle{arxiv}{0.24}{0.2}]{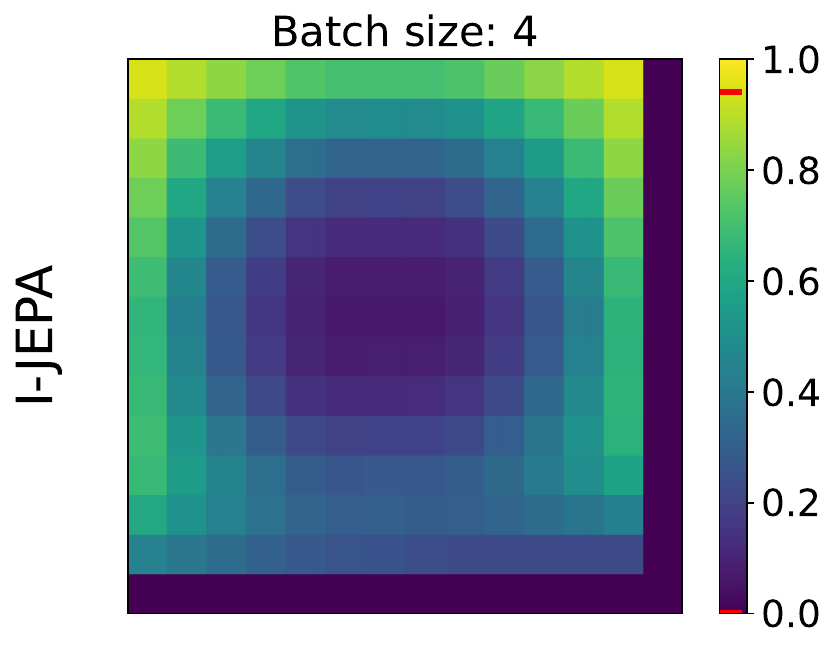}~~
    \includegraphics[scale=\iftoggle{arxiv}{0.24}{0.2}]{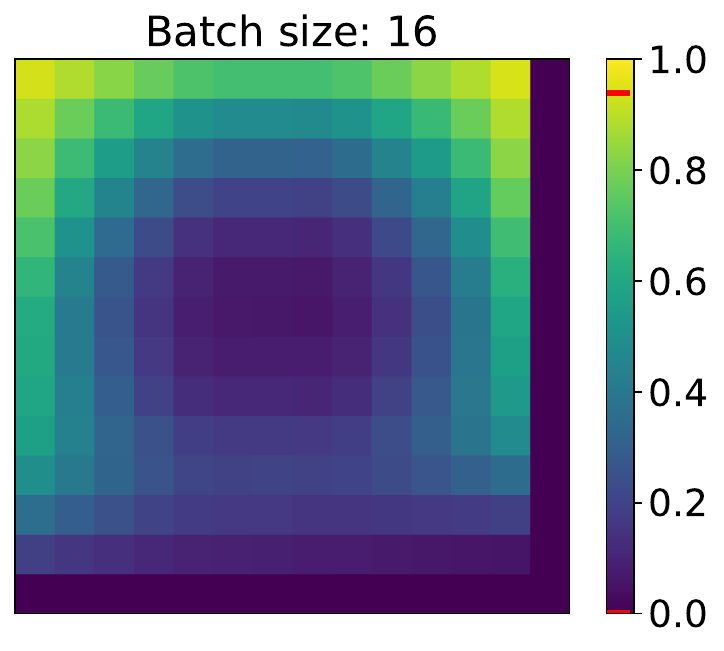}~~
    \includegraphics[scale=\iftoggle{arxiv}{0.24}{0.2}]{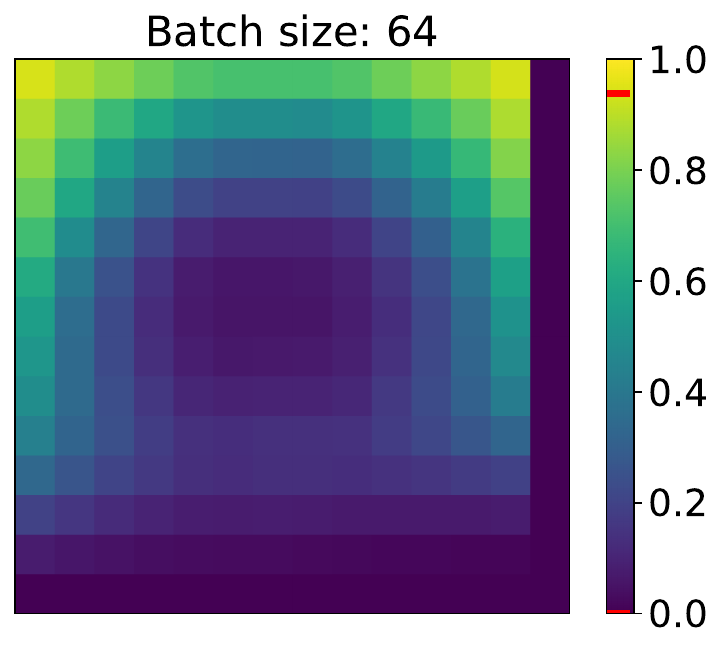}~~
    \includegraphics[scale=\iftoggle{arxiv}{0.24}{0.2}]{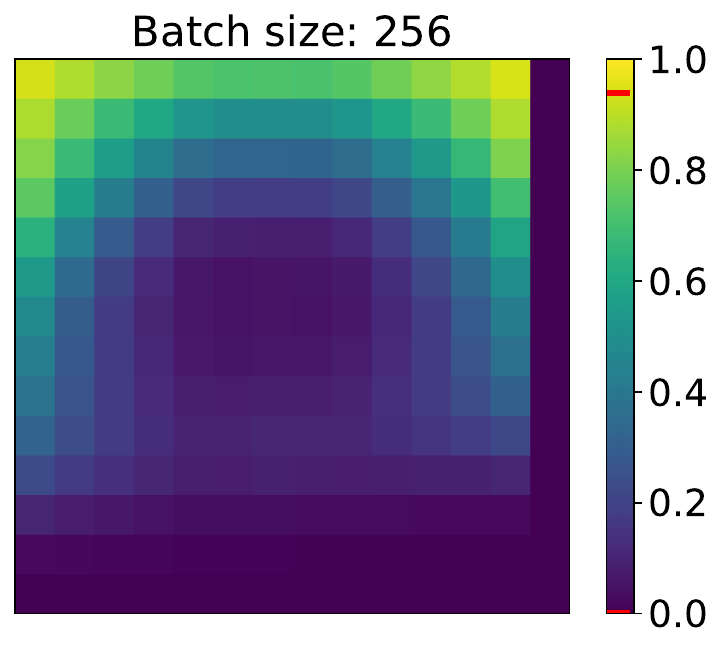}~~
    \includegraphics[scale=\iftoggle{arxiv}{0.24}{0.2}]{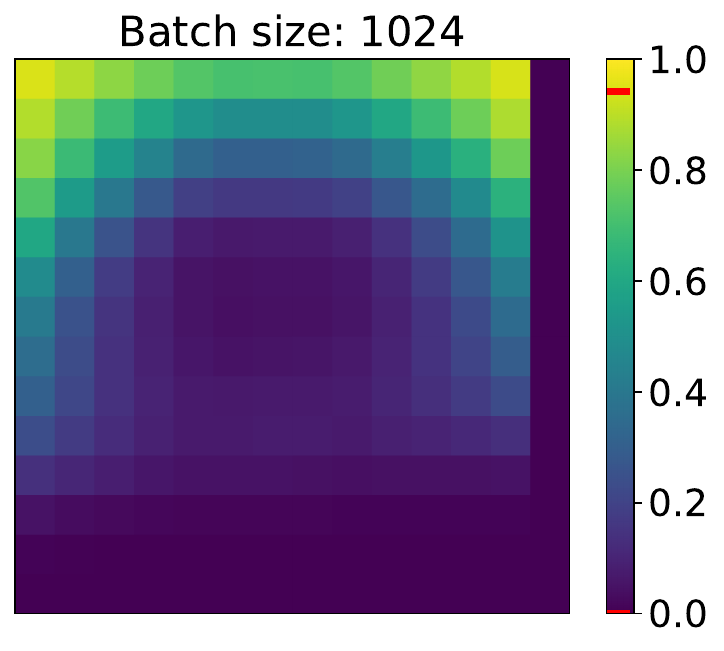} \\
    \includegraphics[scale=\iftoggle{arxiv}{0.24}{0.2}]{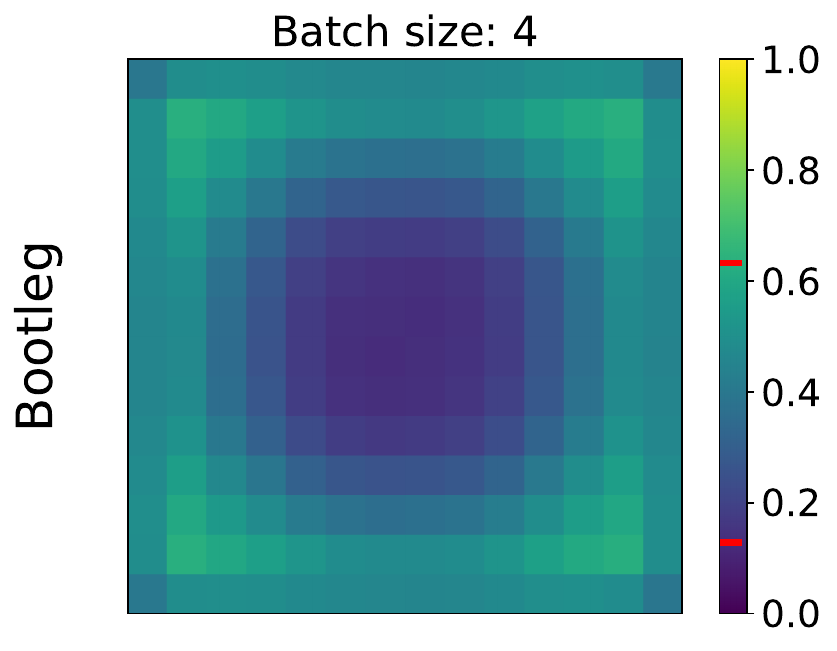}~~
    \includegraphics[scale=\iftoggle{arxiv}{0.24}{0.2}]{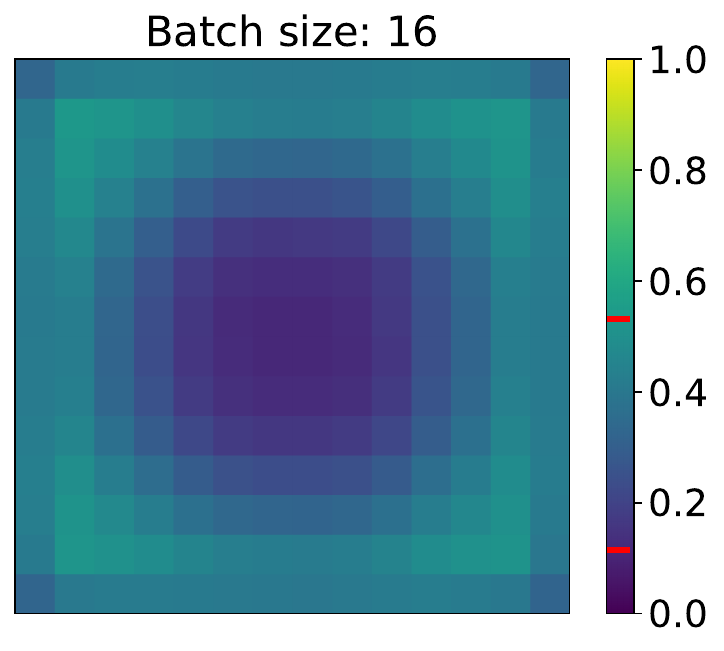}~~
    \includegraphics[scale=\iftoggle{arxiv}{0.24}{0.2}]{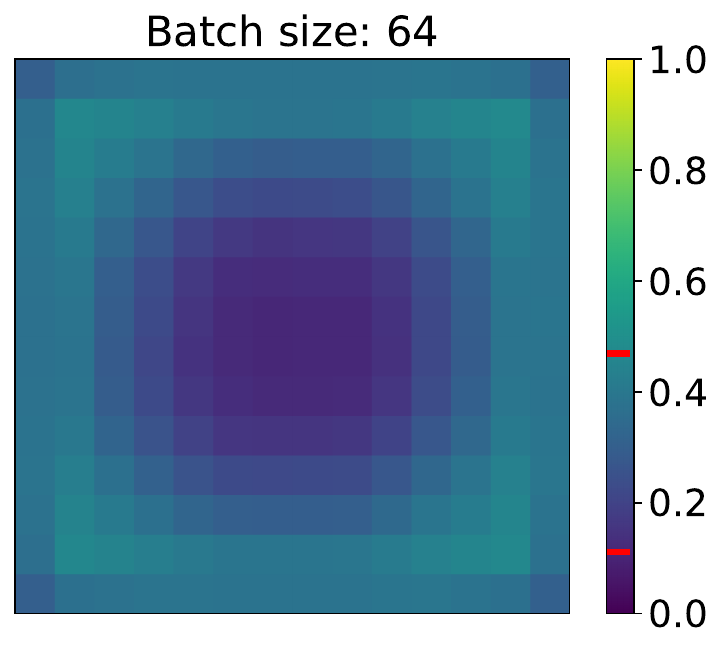}~~
    \includegraphics[scale=\iftoggle{arxiv}{0.24}{0.2}]{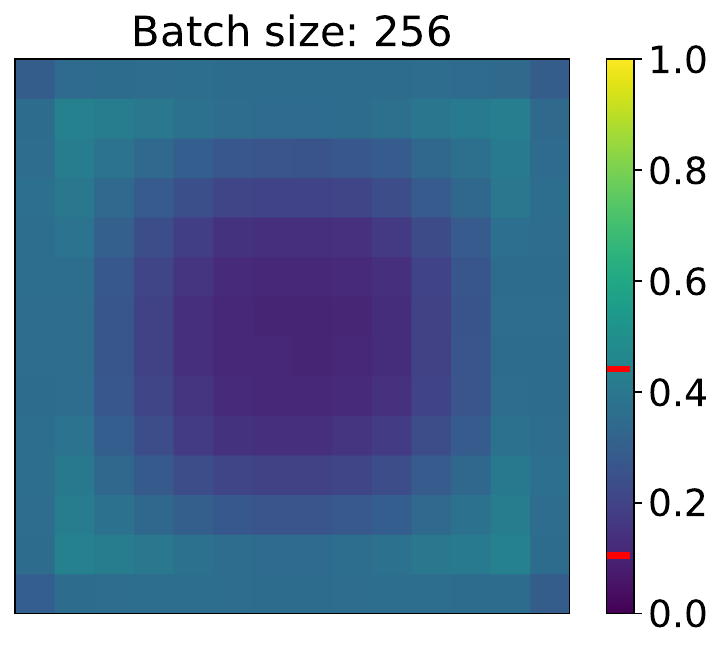}~~
    \includegraphics[scale=\iftoggle{arxiv}{0.24}{0.2}]{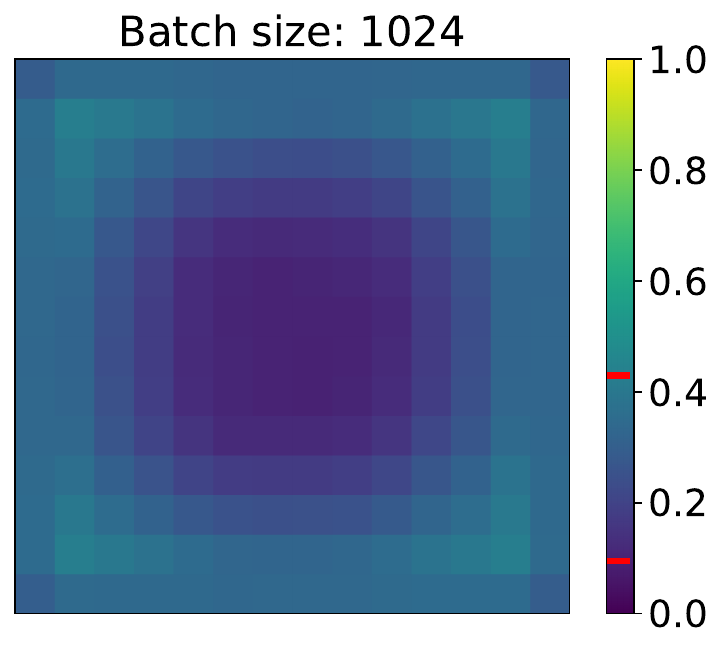}
    \caption{
    Rates at which each token position within the 14 \texttimes 14 grid is visible and presented to the student-encoder.
    \textit{Top row:} I-JEPA masking. \textit{Bottom row:} Bootleg masking.
    We show the change in visibility of the tokens as the per-GPU batch size is increased from 4 (left) to 1024 (right).
    Red marks on the colorbars indicate the least and most frequent any token is visible for a given plot.
    Note that token visibility is more consistent across space (a flatter distribution) for Bootleg masks.
    The I-JEPA masks omit the bottom row and right column entirely, rarely present the centre of the image, and last-in-first-out truncation leads to the bottom four rows being seldom presented when the per-GPU batch size is larger.
    }
    \label{fig:mask-enc-rates}
\end{figure*}

\begin{figure*}
    \centering
    \includegraphics[scale=\iftoggle{arxiv}{0.5}{0.475}]{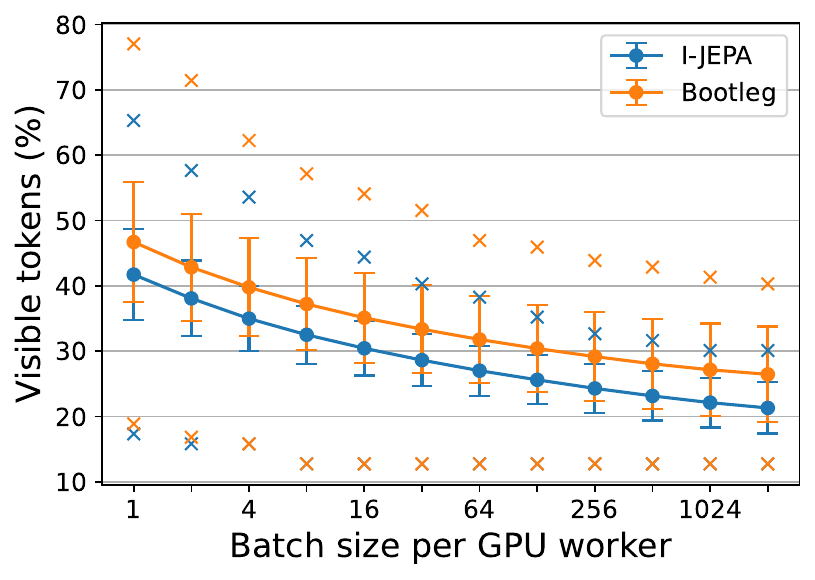} ~~~
    \includegraphics[scale=\iftoggle{arxiv}{0.5}{0.475}]{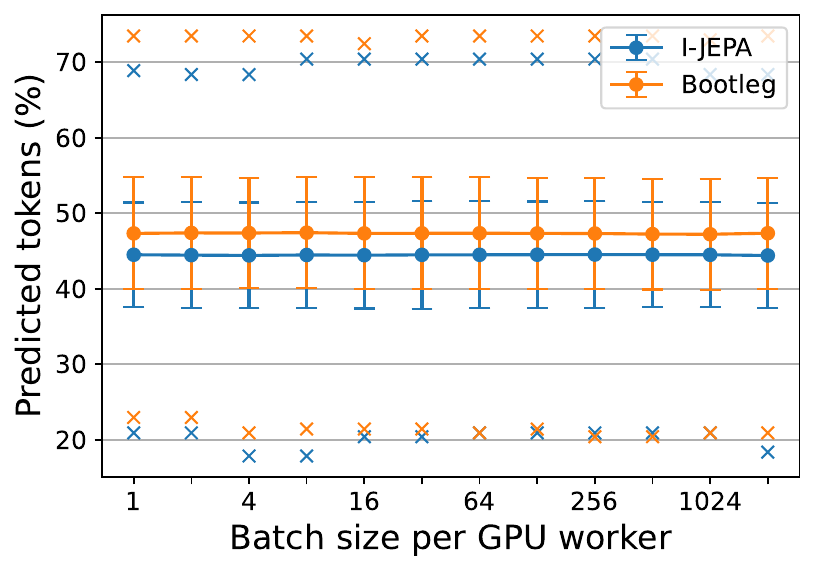} \\
    \includegraphics[scale=\iftoggle{arxiv}{0.5}{0.475}]{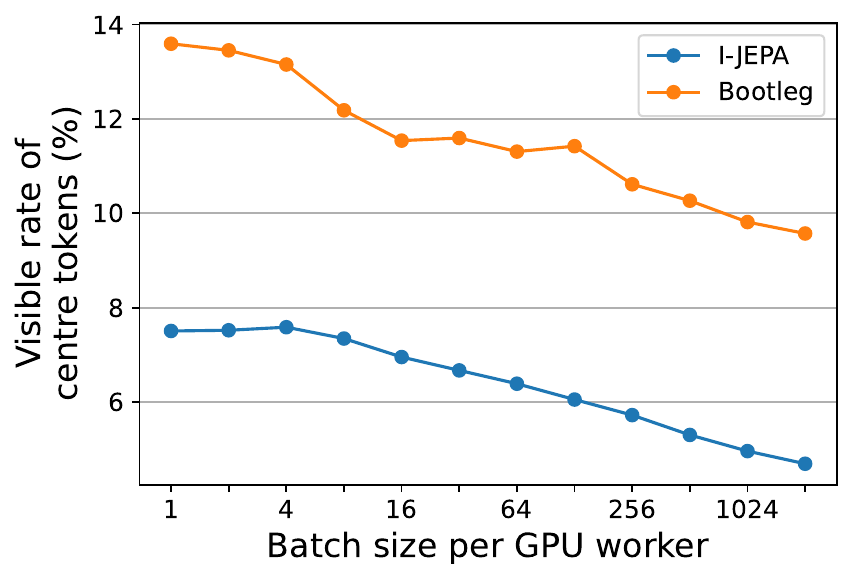} ~~~
    \includegraphics[scale=\iftoggle{arxiv}{0.5}{0.475}]{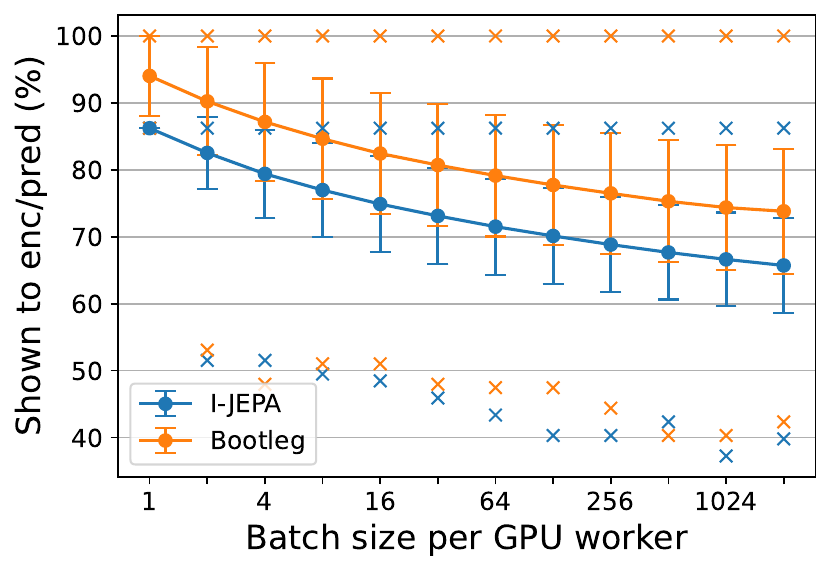}
    \caption{
    Token occurrence rates versus per-GPU batch size for I-JEPA and Bootleg masking strategies.
    Due to stochastic overlap of the predictor masks, the visible tokens need to be truncated to be the same length across all samples on the same GPU.
    \textit{Top left:} Fraction of tokens which are visible (shown to the encoder) decreases as the batch size increases. This is because the target length for the visible tokens is the minimum size of the complement of the predictor masks within over all samples in the batch for the GPU worker, and a larger per-GPU batch size leads to more opportunities to generate four predictor rectangles with minimal overlap.
    \textit{Top right:} Fraction of tokens predicted at least once across all masked out tokens per sample.
    \textit{Bottom left:} Average fraction across the four centre tokens for the rate at which they are visible. Centre tokens infrequently visible for I-JEPA and are visible twice as often for Bootleg.
    \textit{Bottom right:} Fraction of tokens within each image used either as a visible token passed to the encoder, or as a target token whose prediction loss is used to update the model. Due to random overlap and truncation, not all patch tokens are used for at least one task, reducing the efficiency of training.
    For each panel, the standard deviation is shown with error bars, and the min/max values with crosses.
    }
    \label{fig:mask-size-distribution}
\end{figure*}

\subsection{Seeding the RNG for the batch}
For reproducibility of experiments, we want to use a fixed RNG seed.
Note that there is a hierarchy to the processes: (1) a main GPU worker at rank 0, responsible for collecting gradients and saving checkpoints, (2) the set of GPU workers with ranks \eg{} 0--7, which may be running on the same or different nodes/machines to one another, (3) the CPU workers responsible for collating the batches of data, \eg{} 12 per GPU worker.

In I-JEPA's code, the global seed is set to 0.
This is used to sample the random weights for the model, and the seed is set to 0 on \href{https://github.com/facebookresearch/ijepa/blob/52c1ae95d05f743e000e8f10a1f3a79b10cff048/src/train.py#L58-L60}{every GPU worker}.
To ensure each CPU worker uses a different seed from the others for the operations in its batch, \texttt{multiprocessing.Value} is used so the seed is shared across the CPU workers.
However, since the GPU seed is the same, the set of seeds for the CPU workers for one GPU worker are the same as for every other GPU worker.
This results in redundancy in the generated masks across the total batch, distributed across the GPU workers.
(There is actually a race-condition in the dataloader workers, which leads to these shared-mask batches being desynchronized across steps, but the same set masks are used over a small number of steps equal to the number of CPU workers per GPU.)

In Bootleg's implementation, we use a different seed for each GPU worker, which ensures the random crop data transform used differs across GPU workers, unlike in I-JEPA's codebase.
For reproducibility across pre-empted runs, we map the global seed, sample index, and epoch number to a seed to use for each specific presentation stimulus.
Similarly, the CPU workers use seeds for their mask generation which are mapped from the global seed, global batch index, and epoch number.
This ensures the set of masks generated are unique for each batch on each GPU worker, across all global CPU workers.

\subsection{Sampling height/width of predictor masks}
The range of mask height and width values to sample are parametrized by a minimum/maximum block size, and minimum/maximum aspect ratio.
Due to a bug in the I-JEPA code, the \href{https://github.com/facebookresearch/ijepa/blob/52c1ae95d05f743e000e8f10a1f3a79b10cff048/src/masks/multiblock.py#L56-L63}{same random number} was used to select the aspect ratio as the block size, resulting in correlation between the two.
Consequently, small masks are always more landscape and large masks are always more portrait (see \cref{tab:mask-shapes}).

In Bootleg, we fix this bug so size and aspect ratio are sampled independently.
However, when fixing this, the distribution of mask sizes sampled in practice changed.
We thus refined the requested range of mask sizes from $[0.15, 0.20]$ (defined as a fraction of the image area) to a narrower range of $[0.16, 0.183]$, which better reflected the range actually being sampled by I-JEPA (see \cref{tab:mask-shapes} for a comparison of the sampled sizes with this parameter).

Additionally, the mask rectangles for Bootleg use alternating aspect ratios (by swapping the height and width after generating each mask) to increase diversity of masks used for each sample within the worker's batch.
This change also reduces the range of overlap amounts for the masks, since (when the aspect ratio is not square) it is impossible for all masks to overlap completely if two of them are oriented rotated \si{90\degree} from the two.
This increases the consistency of the mask generation and reduces the amount of truncation that needs to be applied to the visible tokens.

\begin{table}[tbh]
\centering
\caption{Distribution of shapes of prediction target mask rectangles for I-JEPA and Bootleg.
We assume a 14\texttimes14 grid of tokens (equivalent to a 224\texttimes 224 image with 16\texttimes16 patch size), and sample 1 million mask sizes.
}
\label{tab:mask-shapes}
\small
\begin{tabular}{cccrr}
\toprule
 & & & \multicolumn{2}{c}{Rate (\%)}\\
\cmidrule(r){4-5}
(h, w) & Area & Aspect & I-JEPA & Bootleg \\
\midrule
(5, 6) & 30 & 0.83 & 26.5 & 18.8 \\
(6, 5) & 30 & 1.20 & 12.6 & 18.7 \\
(5, 7) & 35 & 0.71 & \gray{0.0} & 20.0 \\
(7, 5) & 35 & 1.40 & 36.5 & 20.1 \\
(6, 6) & 36 & 1.00 & 21.8 & 22.4 \\
(5, 8) & 40 & 0.63 & \gray{0.0} &  \gray{0.0} \\
(8, 5) & 40 & 1.60 &  2.6 &  \gray{0.0} \\
\bottomrule
\end{tabular}
\end{table}

\subsection{Sampling height/width of background masks}
\label{s:masking:background}
In I-JEPA, the background mask is supposed to be randomly sampled from $[0.85, 1.0]$ of the total patches in the image, with an aspect ratio always set to 1.0.
For a 224\texttimes 224 image with 16\texttimes 16 pixels per patches, this means a grid of 14\texttimes 14 patch tokens, with the background mask being either all 14\texttimes 14, or a 13\texttimes 13 square within it.
However, due to an off-by-one error in the I-JEPA code, the background mask is capped at one token narrower than the image in both height and width, \href{https://github.com/facebookresearch/ijepa/pull/40#issuecomment-2187880299}{as pointed out by George Hu}, meaning the background mask is always a 13\texttimes 13 square.

In Bootleg, we fix this bug.
Our background mask is randomly sampled to be either a 13\texttimes 13 square of patch tokens, or the full 14\texttimes 14 square grid.

\subsection{Sampling the placement of masks}
In I-JEPA, the mask placement has an off-by-one error, meaning masks are randomly placed somewhere in the upper-left 13\texttimes 13 square of patch tokens, \href{https://github.com/facebookresearch/ijepa/pull/40}{as pointed out by Jack Wilkie}.
When combined with \iftoggle{useappendix}{\cref{s:masking:background},}{the previous item} \textit{``Sampling height/width of background masks''}, this means the background mask for I-JEPA is always the 13\texttimes 13 square in the top-left corner.
Consequently, the I-JEPA student-encoder never sees the right-most column or bottom-most row of image patches, which amounts to 14\% of the image, and the predictor is never asked to predict these tokens either.

In Bootleg, we fix this bug and sample mask locations uniformly across space while restricting the mask to be contiguous.
However, note that this sampling strategy still means prediction masks are less likely to be in the corners of the image and are more likely to be nearer the middle.
This is because there is only one valid position for a rectangular mask to be in any given corner of the image, but many positions allow it to overlap with the middle of the image.
Consequently, our student-encoder sees the edges of the image more often than it sees the middle of the image, and predicts the middle of the image more often than the edges.
It is unclear whether this centre/surround bias is an advantageous component of the masking strategy.

Additionally, note that because the four masks are always sized at least 5\texttimes6, random placement leaves them unlikely to be placed somewhere which does not overlap with the centre patch tokens within the image.
Increasing the background from the top-left 13\texttimes13 grid to 14\texttimes14 greatly increases the number of placements which do not overlap with the centre of the image, resulting in an increased rate at which these tokens are visible (see \cref{fig:mask-size-distribution}).

\subsection{Encoder mask truncation strategy}
We truncate the visible masks to length $V' = \min_i |v^{(i)}|$, the minimum length within the worker's batch.
The visible mask tokens are encoded as a list ordered by their index in row-column order, \ie{} first the top row (from left to right), then the second row (from left to right), \etc{}
In I-JEPA, the visible masks are truncated by \href{https://github.com/facebookresearch/ijepa/blob/52c1ae95d05f743e000e8f10a1f3a79b10cff048/src/masks/multiblock.py#L167}{keeping the first} $V'$ tokens within this sorted order.
Only a minority of samples within a worker's batch will have minimal overlap between the mask rectangles, so most samples undergo some amount of truncation to the list of visible tokens.
This means the context seen by the student-encoder is biased towards the top of the image (which is never truncated), with the student very rarely seeing any of the bottom three rows of image token patches (which are frequently truncated), as shown in \cref{fig:mask-enc-rates} and illustrated in \cref{fig:mask-samples}.

In Bootleg, we fix this issue by randomizing the edge from which the truncation is performed.
We randomly select one of the four corners of the grid, and randomly select an orientation (rows or columns) and truncate from that corner inwards.
The edge and orientation is independently selected for each sample within the worker's batch.

We considered other truncation strategies, such as removing a token with a uniform random distribution across all visible tokens or across all visible tokens at the edge of the image, but found it was better to remove tokens from one randomly selected edge instead.

\subsection{Impact of batch size effects on performance}

We explored the effect of batch-size related masking effects on model performance.
Holding the total batch size constant, we adjusted the batch size per GPU by altering the number of GPUs used to train the model.
As shown in \cref{tab:bs-perf}, both Bootleg and I-JEPA exhibit a decrease in performance when the batch size per GPU is taken outside its optimal range.
Due to truncation of the visible tokens to fit the batch neatly on the GPU (as described above), larger batch sizes per GPU ask the model to solve a harder training task (since fewer tokens are visible) than configurations which have a smaller batch size per GPU, even with the total batch size held constant.
The optimal configuration therefore may be one where tokens are ``wasted'' by discarding them instead of presenting them to either the encoder or the predictor in order to set the task difficulty optimally.

We find the effect of per-GPU batch size is much smaller for Bootleg than I-JEPA---for each metric the maximum delta in performance is between half and a fifth as large for Bootleg as it was for I-JEPA (see \cref{tab:bs-perf} rows marked $\Delta_\text{max}$).

\begin{table}[tbh]
\centering
\caption{Effect of per-GPU batch size (GPU-BS) on model performance for I-JEPA and Bootleg.
ViT-S, trained on IN-1k, 300 ep., (final hparams) with 2048 total batch size for all experiments.
We alter the number of GPU workers to control the batch size per GPU worker.
Rows $\Delta_\text{max}$ show the maximum difference in performance over the GPU batch sizes for each metric.
}
\label{tab:bs-perf}
\small
\begin{tabular}{lrccccccc}
\toprule
 &  &  &  &  & \multicolumn{3}{c}{IN-1k acc.} & \multicolumn{1}{c}{ADE20K mIoU} \\
\cmidrule(r){6-8} \cmidrule{9-9}
Method & GPU-BS & Seen (\%) & Target (\%) & Adj. (\%) & Patch & CLS & X-Blk & kNN \\
\midrule
I-JEPA   & 64 & $27.0 \pm 3.8$ & $44.5 \pm 2.3$ & $23.6 \pm 3.8$ & \sbest{55.2} & \na{} & \best{64.7} & \sbest{10.9} \\
         & 128 & $25.6 \pm 3.8$ & $44.5 \pm 2.2$ & $22.3 \pm 3.8$ & \best{55.6} & \na{} & \sbest{63.9} & 10.7 \\
\rowcolor{vlightgray}\cellcolor{white}
         & 256 & $24.3 \pm 3.7$ & $44.5 \pm 2.2$ & $21.1 \pm 3.8$ & 53.4 & \na{} & 62.3 & 10.5 \\
         & 512 & $23.1 \pm 3.7$ & $44.5 \pm 2.2$ & $19.9 \pm 3.8$ & 52.3 & \na{} & 62.6 & 10.1 \\
         & 1024 & $22.1 \pm 3.8$ & $44.5 \pm 2.1$ & $18.9 \pm 3.9$ & 52.7 & \na{} & 62.7 & \best{11.2} \\
\cmidrule(lr){2-2}
         & $\Delta_\text{max}$ & \decr{-4.9} & \decr{-0.0} & \decr{-4.7} & \decr{-3.2} & & \decr{-2.4} & \decr{-1.0} \\
\addlinespace
Bootleg  & 64 & $31.9 \pm 6.7$ & $47.3 \pm 2.4$ & $24.7 \pm 4.7$ & 67.3 & 68.4 & 74.1 & 21.7 \\
         & 128 & $30.5 \pm 6.7$ & $47.3 \pm 2.4$ & $23.6 \pm 4.7$ & 67.5 & \sbest{68.8} & \sbest{74.2} & 21.9 \\
\rowcolor{vlightgray}\cellcolor{white}
         & 256 & $29.3 \pm 6.7$ & $47.3 \pm 2.3$ & $22.6 \pm 4.9$ & \best{67.8} & \best{68.9} & \best{74.4} & \best{22.0} \\
         & 512 & $28.2 \pm 6.9$ & $47.3 \pm 2.3$ & $21.7 \pm 5.0$ & \sbest{67.6} & 68.5 & 74.1 & \sbest{22.0} \\
         & 1024 & $27.2 \pm 7.1$ & $47.3 \pm 2.3$ & $20.8 \pm 5.2$ & 66.9 & 67.7 & 73.8 & 21.4 \\
\cmidrule(lr){2-2}
         & $\Delta_\text{max}$ & \decr{-4.7} & \decr{-0.0} & \decr{-3.8} & \decr{-0.9} & \decr{-1.2} & \decr{-0.5} & \decr{-0.6} \\
\bottomrule
\end{tabular}
\end{table}

\FloatBarrier
\section{Compute requirements}
\label{s:compute}

All experiments were run on Slurm clusters of the Digital Research Alliance of Canada and the Vector Institute.
We describe here the setup used for the final versions of our experiments, with costs measured from our training logs.

\mypara{Pretraining.}
Each pretraining run uses two nodes, each with four NVIDIA H100 80\,GB GPUs and at least 48 CPU cores and 400\,GB of RAM.
We train with a global batch of 2048 images (256 per GPU) in bfloat16 mixed precision with PyTorch DDP, reading IN-1k from uncompressed tar archives.
\Cref{tab:compute-pretrain} gives the cost of one pretraining run at each model size.
The per-epoch time is the median over the run, as logged by our training loop; the totals exclude the online $k$-NN and feature-analysis monitoring, which we run periodically and which is not needed to reproduce the results.
Bootleg's additional self-distillation targets have a negligible effect on throughput (\cref{s:no-forward-mse}), so I-JEPA, MAE and CrossMAE pretraining at the same model size, epoch count and batch size cost about the same.

\begin{table}[tbh]
\centering
\caption{%
Cost of a single Bootleg pretraining run on IN-1k, on 8 \texttimes{} H100 GPUs (two nodes of 4) with a global batch size of 2048.
Per-epoch times are the median over the run; totals are per-epoch time $\times$ epochs.
}
\label{tab:compute-pretrain}
\small
\begin{tabular}{lrrrr}
\toprule
Encoder  & Time / epoch (s) & Epochs & Wall-clock (h) & GPU-hours \\
\midrule
ViT-S/16 & 78  & 300 & 6.5  & 52  \\
ViT-S/16 & 78  & 600 & 13\phantom{.0}   & 104 \\
ViT-B/16 & 134 & 600 & 22\phantom{.0}   & 179 \\
ViT-L/16 & 250 & 600 & 42\phantom{.0}   & 334 \\
\bottomrule
\end{tabular}
\end{table}

\mypara{Evaluation.}
Downstream evaluations use a frozen encoder and a single node, with one to four GPUs (NVIDIA L40S 48\,GB or H100 80\,GB).
Typical wall-clock times per pretrained model are as follows.
The IN-1k frozen-probe sweep trains all linear and attentive heads in parallel on one shared encoder forward pass, and takes 7--9\,h on one or two L40S GPUs.
The iNat21 frozen probe takes 5--7\,h on two (ViT-S/B) or four (ViT-L) H100 GPUs.
The I-JEPA linear-evaluation protocol (\cref{s:ijepa-linprobe}) takes 1--2\,h on two nodes of four H100 GPUs.
The ADE20K, Cityscapes and COCO-Stuff linear probes each take 1--9\,h on one or two L40S GPUs; the ADE20K linear probe takes 1.4\,h on four H100 GPUs.
The $k$-NN segmentation evaluations run on a single GPU in under 2\,h, but need up to 80\,GB (ViT-S), 160\,GB (ViT-B) or 240\,GB (ViT-L) of host RAM to hold the patch features.
Each of the 19 VTAB tasks takes about 0.5\,h on one L40S GPU.

\mypara{Total compute.}
The experiments reported in the paper are a small fraction of the compute used by the project as a whole.
The full project also included the hyperparameter searches (\cref{s:pretraining-hparams}), the ablations and preliminary experiments, the reproductions of baseline methods, and runs that were abandoned or did not reach the paper.
Many of these earlier runs were also made with less efficient versions of our code, or on hardware that was running below its nominal throughput, so their cost does not reflect the cost of reproducing our results.
The configuration files for every experiment in the paper are released in our code repository, \url{https://github.com/scottclowe/bootleg}.

\FloatBarrier
\section{No-forward MSE loss function}
\label{s:no-forward-mse}

Our hidden-self-distillation method, Bootleg, uses an increased number of training targets compared to MAE or I-JEPA.
We found an increased proportion of experiment run-time was devoted to computing the loss, which increased the more distillation targets were used.
For our proposed methodology, in which the output from every fourth transformer block is used as a self-distillation target, the computational overhead of the increased targets is not much (4 times as much as for I-JEPA; less than 2\% of total runtime).
But for some of our preliminary explorations (see \cref{s:target-choice}) using the output and residuals from every block (9 times more targets than Bootleg, 36 times more than I-JEPA), the loss computation was significant (11\% of total runtime), motivating us to find a solution.

Note that we do not need to know the loss value on every step---periodic evaluation and logging of the loss is sufficient for tracking training---we only need to compute the loss on each step so we can compute gradients.
So to reduce the amount of compute spent during training, we propose and implemented a no-forward version of the MSE loss, \texttt{MSELossNoForward}, shown in \cref{alg:mse_loss_no_forward}.
This method skips the forward computation of the loss, returning a dummy value, but will compute the gradient of the loss to return the same values in the backward pass as \texttt{MSELoss}.
It can be used in the compute graph like the conventional \texttt{MSELoss}, with downstream operations, such as scaling the loss by the world size or with a GradScaler, fed back into the backward pass of \texttt{MSELossNoForward}.

In practice, we accompany this with a separate forward computation of the MSE loss spaced at regular intervals (\eg{} every twenty steps) for monitoring purposes.

To measure the saving, we trained the same configuration (ViT-S/16 with 36 targets: the output and both residual streams of every block; 2 nodes of 4 L40S GPUs) with each loss, changing nothing else.
Computing the loss took 14.6\,s per epoch with the standard MSE loss and 6.0\,s with \texttt{MSELossNoForward}, a 59\% reduction, and the total epoch time fell from 215.1\,s to 202.6\,s ($-5.8\%$).
More broadly, across the 38 ViT-S runs in our logs with 12 to 50 targets and the standard MSE loss, the loss took a median 7.0\% of each epoch (12.1\,s), compared with 3.0\% (6.0\,s) for the runs using \texttt{MSELossNoForward}.

\newsavebox{\mselosslistingbox}
\begin{lrbox}{\mselosslistingbox}
\begin{minipage}{\iftoggle{arxiv}{}{\dimexpr1.1}\textwidth}
\begin{lstlisting}[
    language=Python,
    nolol,
]
class MSELossNoForward(torch.autograd.Function):
    @staticmethod
    def forward(
        ctx, input: torch.Tensor, target: torch.Tensor, reduction: str = "mean"
    ):
        # Compute the difference between input and target and save it.
        # This computation is only needed for the backward pass, but doing
        # it now halves the memory requirement we need in the context.
        ctx.save_for_backward(input - target)
        ctx.reduction = reduction
        # Return a dummy scalar tensor (no forward computation).
        # Match the device and dtype of input.
        dummy = input.new_zeros((), requires_grad=True)
        return dummy

    @staticmethod
    def backward(ctx, grad_output):
        # Recover difference between input and target, saved in forward pass
        difference = ctx.saved_tensors[0]
        reduction = ctx.reduction
        # Constant scaling of 2.0 comes from derivative of mean squared error.
        scaling = 2.0
        # If we are simulating taking the mean, scale against the number of
        # elements being averaged.
        if reduction == "mean":
            scaling /= difference.numel()
        # Usually grad_output is 1.0/world_size, but may differ if using a
        # GradScaler. Multiply grad_output by the constant term.
        grad_output *= scaling
        # Chain rule: multiply upstream gradient by (input - target).
        grad_input = grad_output * difference
        grad_target = None  # No gradient w.r.t. target
        grad_reduction = None
        return grad_input, grad_target, grad_reduction
\end{lstlisting}%
\end{minipage}
\end{lrbox}
\begin{figure*}
\captionof{lstlisting}{Pytorch implementation of no-forward, backward-only, mean squared error (MSE) loss function.}
\resizebox{\textwidth}{!}{\usebox{\mselosslistingbox}}
\label{alg:mse_loss_no_forward}
\end{figure*}\addtocounter{figure}{-1}

\FloatBarrier
\section{Evaluation methodology}
\label{a:eval-method}

For baseline comparisons, we use officially released model checkpoints where possible.
For I-JEPA, no ViT-S, -B, or -L checkpoints were publicly available; we trained these using our fork of the I-JEPA codebase.
For MAE \citep{he2022masked} and data2vec~2.0 \citep{baevski2023data2vec2}, no ViT-S checkpoints were publicly available; we trained these using our implementation of their respective methodologies.

\subsection{Image classification (IN-1k and iNat21)}

\subsubsection{Frozen probes}
\label{a:eval-method-classification-probe}

As described in \iftoggle{useappendix}{\cref{s:exp:in1k}}{the main text, Sec.~5.1}, we evaluated the pretrained networks by using probes of the frozen encoder.
Since the majority of compute is spent passing the images through the frozen encoder, which is the same for all probes, we perform a simultaneous hyperparameter sweep on each probe type, with 25 hyperparameters considered for each.

We consider four types of frozen probes.
Firstly, we consider the long-standing linear probe of the average patch embeddings (column ``\textbf{Patch}'' in tables).
We discard the CLS token and register tokens, take the average of the patch embeddings, apply batch norm, and a single learnable linear layer to predict the IN-1k classes.
Secondly, we consider the performance of a linear probe of the CLS token embedding (column ``\textbf{CLS}'' in tables).
We discard the patch tokens and register tokens, add a LayerNorm layer without trainable parameters, and train a single learnable linear layer on the CLS token embedding.
Thirdly, we consider the performance of a minimal attentive probe (column ``\textbf{X-Attn}'' in tables).
We keep the patch, CLS, and any register embeddings produced by the model; these are passed to a cross-attention block with a single learnable query vector, followed by a single linear head only.
We use the same width and number of attention heads as in the pretrained ViT.
Finally, we consider the performance of a non-linear attentive probe (column ``\textbf{X-Blk}'' in tables).
This was performed following the implementation in V-JEPA \citep{bardes2024vjepa, assran2025vjepa2}; the method is the same as X-Attn, except there is an MLP-block between the attention layer and the linear head.

In \cref{tab:main-results}, we show the maximum of the Patch and CLS linear probes (columns ``\textbf{Lin}''). 
For the full breakdown, see \cref{tab:main-results-classification}.

We train the probes with a batch size of 1024 on either the IN-1k train set or iNat21 train set \citep{inaturalist21} for 20 epochs.
(As iNat21 is around twice as large as IN-1k, this results in around twice as many training steps for its probes.)
Training data is augmented through (1) random crops, sized $(0.3, 1.0)$ of the original image and random stretch aspect ratio in the range 3:4 to 4:3, which are resized to 224\texttimes 224 pixels with bicubic interpolation; (2) random horizontal flips with $p=0.5$; (3) and random colour jitter of up to $0.4$ for brightness, contrast, and saturation.
See \cref{tab:hparams-in1k-probe-config}.

The training (learning rate and weight decay) hyperparameters we considered are a 5\texttimes5 grid of logarithmically spaced values:
$\eta_{\text{max}} \in \{0.0005,\allowbreak 0.001,\allowbreak 0.002,\allowbreak 0.004,\allowbreak 0.008\}$ and
$\lambda \in \{0.0005,\allowbreak 0.001,\allowbreak 0.002,\allowbreak 0.004,\allowbreak 0.008\}$, respectively.

The performance of each probe is evaluated on the IN-1k/iNat21 validation set, with an 87.5\% centre crop (resize to 256\texttimes 256, then crop to 224\texttimes 224).
For each of the four probe types, we report the best result across the 25 training hyperparameters considered.

\begin{table}[tbh]
\centering
\caption{%
Hyperparameter configuration for IN-1k classification probes.
}
\label{tab:hparams-in1k-probe-config}
\small
\begin{tabular}{lll}
\toprule
          & Hyperparameter    & Value \\
\midrule
Encoder   & Drop path         & \gray{0.0}          \\
\addlinespace
Data      & Input size        & 224\texttimes224    \\
          & Cropping          & RandCrop(0.3, 1.0)  \\
          & Stretch           & [\nicefrac{3}{4}, \nicefrac{4}{3}] \\
          & Interpolation     & Bicubic             \\
          & Transforms        & HFlip($p\!=\!0.5$)  \\
          &                   & ColorJitter(0.4)    \\
\addlinespace
Training  & Loss              & Cross-entropy       \\
          & Label smoothing   & \gray{0.0}          \\
          & Optimizer         & AdamW({\footnotesize$\beta\!=\!(0.9, 0.999)$}) \\
          & LR schedule       & Cosine decay        \\
          & LR schedule warmup& 5 epochs            \\
          & LR initial        & 0.0                 \\
          & LR maximum        & $[0.0005, 0.0080]$ \\
          & LR final          & 0.0                  \\
          & WD                & $[0.0005, 0.0080]$ \\
          & Batch size        & 1024            \\
          & Epochs            & 20              \\
\bottomrule
\end{tabular}
\end{table}

\cref{tab:main-results-classification} shows the performance on IN-1k and iNat21 with the Patch and CLS performance separated, instead of showing the maximum.

\begin{table*}[tb]
\centering
\caption{
\textbf{Image classification results}, extending \cref{tab:main-results}.
As with \cref{tab:main-results}, except patch average and CLS linear probes are shown separately instead of their maximum.
We show the top-1 accuracy (\%) for image classification on IN-1k and iNat21 using a probe on the frozen encoder (LP: Patch, CLS, X-Blk).
For IN-1k we also show $k$-NN classification accuracy (kNN) of the patch-averaged (Patch) and CLS embeddings of the last block (before the final LayerNorm, as for the probes), with cosine similarity, temperature $0.07$, and the best $k \in \{1, 2, 5, \ldots, 500\}$.
We highlight the \best{best} and \sbest{second best} SSL method for each group of models.
First block (black): single-view masked SSL methods trained on IN-1k (MAE, CrossMAE, data2vec~2.0, I-JEPA, Bootleg).
{\color{gray}Grey}:~models with a smaller patch size (14\texttimes14 rather than 16\texttimes16), and hence 30\% more patch tokens and more compute per image.
(dst): Distilled from pretrained ViT-g.
\na{}: Model lacks a CLS token to evaluate.
Second block (black): multi-view joint-embedding methods trained on IN-1k (DINO, iBOT, MoCo~v3, SimDINOv2), using the official checkpoints with our evaluation protocol.
MV:~multi-view (\cmark) or single-view (\xmark) pretraining.
$^{*}$:~trained by us (no official checkpoint available); all other baselines are evaluated from their official checkpoints.
}
\label{tab:main-results-classification}
\adjustbox{max width=\textwidth}{
{\setlength{\tabcolsep}{4pt}
\small
\begin{tabular}{llclr@{\hspace{0.5em}}cc@{\hspace{0.5em}}ccc@{\hspace{0.5em}}ccc}
\toprule
 &  & &  &  & \multicolumn{5}{c}{IN-1k acc.} & \multicolumn{3}{c}{iNat21 acc.} \\
\cmidrule(r){6-10} \cmidrule{11-13}
 &  & &  &  & \multicolumn{2}{c}{kNN} & \multicolumn{3}{c}{LP} & \multicolumn{3}{c}{LP} \\
\cmidrule(r){6-7} \cmidrule(r){8-10} \cmidrule{11-13}
Arch & Method & MV & Data & Ep. & Patch & CLS & Patch & CLS & X-Blk & Patch & CLS & X-Blk \\
\midrule
ViT-S/16 & MAE$^{*}$ & \xmark & IN-1k & 800 & 27.1 & \sbest{35.0} & 47.0 & 49.8 & 66.4 & 22.3 & 26.2 & 57.5 \\
 & CrossMAE & \xmark & IN-1k & 800 & 33.1 & 30.5 & 51.8 & \sbest{50.9} & \sbest{68.8} & 25.7 & \sbest{26.7} & \sbest{59.8} \\
 & data2vec 2.0$^{*}$ & \xmark & IN-1k & 200 & 20.5 & 17.9 & 39.9 & 41.7 & 62.2 & 15.4 & 14.6 & 47.4 \\
 & I-JEPA$^{*}$ & \xmark & IN-1k & 600 & \sbest{34.7} & \na{} & \sbest{52.4} & \na{} & 61.9 & \sbest{26.8} & \na{} & 48.4 \\
 & Bootleg (ours) & \xmark & IN-1k & 600 & \best{61.6} & \best{64.8} & \best{69.8} & \best{70.4} & \best{75.3} & \best{42.6} & \best{47.8} & \best{67.4} \\
\cmidrule(r){2-5}
 & DINO & \cmark & IN-1k & 800 & 43.8 & \sbest{74.6} & 62.4 & \sbest{76.1} & \sbest{76.3} & \sbest{42.6} & \best{60.9} & \sbest{68.8} \\
 & iBOT & \cmark & IN-1k & 800 & \best{52.1} & \best{75.2} & \best{68.8} & \best{77.0} & \best{77.5} & \best{44.0} & \sbest{60.8} & \best{69.7} \\
 & MoCo-v3 & \cmark & IN-1k & 300 & \sbest{51.2} & 69.3 & \sbest{66.0} & 73.1 & 74.6 & 40.7 & 52.4 & 64.5 \\
\cmidrule(r){2-5}
{\color{gray}ViT-S/14} & {\color{gray}DINOv2 (dst)} & {\color{gray}\cmark} & {\color{gray}LVD} & {\color{gray}Unk.} & {\color{gray}\best{70.4}} & {\color{gray}\best{78.9}} & {\color{gray}\best{77.9}} & {\color{gray}\best{80.6}} & {\color{gray}\best{81.5}} & {\color{gray}\best{64.2}} & {\color{gray}\best{74.0}} & {\color{gray}\best{78.9}} \\
\midrule
ViT-B/16 & MAE & \xmark & IN-1k & 1600 & 45.2 & 46.0 & 66.1 & \sbest{67.2} & \sbest{76.0} & 37.4 & \sbest{43.3} & \sbest{70.7} \\
 & CrossMAE & \xmark & IN-1k & 800 & 45.4 & \sbest{46.5} & 65.5 & 64.8 & 75.6 & 38.3 & 41.7 & 70.2 \\
 & data2vec 2.0 & \xmark & IN-1k & 200 & 23.3 & 41.2 & 62.2 & 61.0 & 73.7 & 29.7 & 31.3 & 61.3 \\
 & I-JEPA$^{*}$ & \xmark & IN-1k & 600 & \sbest{55.0} & \na{} & \sbest{67.0} & \na{} & 72.4 & \sbest{41.4} & \na{} & 63.0 \\
 & Bootleg (ours) & \xmark & IN-1k & 600 & \best{66.0} & \best{72.2} & \best{75.5} & \best{76.7} & \best{79.2} & \best{50.9} & \best{58.3} & \best{74.2} \\
\cmidrule(r){2-5}
 & DINO & \cmark & IN-1k & 400 & 54.6 & 76.1 & 71.4 & 77.5 & 77.7 & \sbest{53.3} & \sbest{63.7} & 70.1 \\
 & iBOT & \cmark & IN-1k & 400 & 55.6 & \sbest{77.2} & \sbest{73.7} & \sbest{78.8} & \best{79.2} & 50.4 & 63.4 & \sbest{72.5} \\
 & MoCo-v3 & \cmark & IN-1k & 300 & \sbest{57.6} & 72.1 & 71.7 & 76.2 & 77.3 & 49.2 & 57.3 & 68.4 \\
 & SimDINOv2 & \cmark & IN-1k & 100 & \best{64.8} & \best{77.8} & \best{75.7} & \best{79.3} & \best{79.2} & \best{59.0} & \best{66.3} & \best{73.0} \\
\cmidrule(r){2-5}
{\color{gray}ViT-B/14} & {\color{gray}Franca} & {\color{gray}\cmark} & {\color{gray}IN-22k} & {\color{gray}(90)} & {\color{gray}\sbest{66.4}} & {\color{gray}\sbest{78.7}} & {\color{gray}\sbest{78.7}} & {\color{gray}\sbest{81.7}} & {\color{gray}\sbest{83.0}} & {\color{gray}\sbest{63.2}} & {\color{gray}\sbest{73.6}} & {\color{gray}\sbest{81.0}} \\
{\color{gray}} & {\color{gray}DINOv2 (dst)} & {\color{gray}\cmark} & {\color{gray}LVD} & {\color{gray}Unk.} & {\color{gray}\best{74.8}} & {\color{gray}\best{82.0}} & {\color{gray}\best{82.2}} & {\color{gray}\best{83.7}} & {\color{gray}\best{84.2}} & {\color{gray}\best{71.4}} & {\color{gray}\best{80.4}} & {\color{gray}\best{84.3}} \\
\midrule
ViT-L/16 & MAE & \xmark & IN-1k & 1600 & 41.3 & \sbest{58.3} & \sbest{73.0} & \sbest{75.5} & 79.5 & \sbest{43.9} & \sbest{51.8} & \sbest{76.3} \\
 & CrossMAE & \xmark & IN-1k & 800 & 49.4 & 44.7 & 71.5 & 71.8 & 78.7 & 43.2 & 50.1 & 75.1 \\
 & data2vec 2.0 & \xmark & IN-1k & 200 & 33.9 & 51.2 & 70.5 & 72.7 & \sbest{80.0} & 38.6 & 41.5 & 72.7 \\
 & I-JEPA$^{*}$ & \xmark & IN-1k & 600 & \sbest{60.2} & \na{} & 68.4 & \na{} & 72.3 & 41.1 & \na{} & 61.3 \\
 & Bootleg (ours) & \xmark & IN-1k & 600 & \best{61.1} & \best{75.1} & \best{77.5} & \best{79.1} & \best{80.6} & \best{53.2} & \best{61.2} & \best{77.1} \\
\cmidrule(r){2-5}
 & iBOT & \cmark & IN-1k & 250 & \sbest{55.0} & \sbest{78.2} & \sbest{76.3} & \sbest{80.3} & \sbest{79.2} & \sbest{54.6} & \sbest{66.5} & \sbest{74.3} \\
 & SimDINOv2 & \cmark & IN-1k & 100 & \best{73.5} & \best{81.0} & \best{80.1} & \best{81.9} & \best{81.6} & \best{62.5} & \best{69.0} & \best{77.1} \\
\cmidrule(r){2-5}
{\color{gray}ViT-L/14} & {\color{gray}CAPI} & {\color{gray}\xmark} & {\color{gray}IN-1k} & {\color{gray}6394} & {\color{gray}62.0} & {\color{gray}\na{}} & {\color{gray}77.6} & {\color{gray}\na{}} & {\color{gray}83.2} & {\color{gray}47.6} & {\color{gray}\na{}} & {\color{gray}80.7} \\
{\color{gray}} & {\color{gray}Franca} & {\color{gray}\cmark} & {\color{gray}LAION} & {\color{gray}(3.2)} & {\color{gray}\sbest{73.3}} & {\color{gray}\sbest{80.3}} & {\color{gray}\sbest{82.5}} & {\color{gray}\sbest{83.4}} & {\color{gray}\sbest{84.4}} & {\color{gray}\sbest{71.5}} & {\color{gray}\sbest{77.0}} & {\color{gray}\sbest{86.2}} \\
{\color{gray}} & {\color{gray}DINOv2 (dst)} & {\color{gray}\cmark} & {\color{gray}LVD} & {\color{gray}Unk.} & {\color{gray}\best{76.1}} & {\color{gray}\best{83.4}} & {\color{gray}\best{84.1}} & {\color{gray}\best{85.4}} & {\color{gray}\best{85.9}} & {\color{gray}\best{76.3}} & {\color{gray}\best{83.2}} & {\color{gray}\best{87.3}} \\
\bottomrule
\end{tabular}
}
}
\end{table*}

\FloatBarrier
\subsection{VTAB classification}
\label{s:vtab}

We evaluate on the Visual Task Adaptation Benchmark (VTAB) \citep{Zhai2019_VTAB}, which comprises 19 diverse visual classification tasks grouped into three categories. \textbf{Natural}: 7 standard recognition tasks set in the natural world such as CIFAR-100, Caltech-101, and Flowers-102. \textbf{Specialized}: 4 domain-specific tasks including medical scans, microscopy, and satellite imagery. \textbf{Structured}: 8 tasks requiring geometric or spatial reasoning, such as object counting and orientation estimation, using synthetic images.

\subsubsection{Method}
\label{s:vtab-method}
Following the VTAB methodology, we train our probes for 50 epochs on a 1000 sample subset of the training set only, without data augmentation.
We use the same probe types as for IN-1k (Patch, CLS, X-Blk), evaluated on the full test set.

The training hyperparameters (learning rate and weight decay) are selected by training a model on a subset of 800 training samples, validated on 200 samples.
The best performing method is then trained again from scratch on the set 1000 training samples.
The evaluation hyperparameters considered are shown in \cref{tab:hparams-vtab}.
Due to the diversity of the tasks, we needed a wider range of learning rate and weight decay hyperparameters to attain optimal results across the models and dataset pairs; consequently, we sweep an 11\texttimes11 grid of learning rate and weight decay values for each head, using a wider spacing (factor of 3 spacing instead of 2):
$\eta_\text{max} \in \{3{\times}10^{-5},\allowbreak 1{\times}10^{-4},\allowbreak 3{\times}10^{-4},\allowbreak \cdots, 3.0\}$
$\lambda \in \{0,\allowbreak 1{\times}10^{-4},\allowbreak 3{\times}10^{-4},\allowbreak \cdots, 3.0\}$.

\begin{table}[tbh]
\centering
\caption{%
Hyperparameter configuration for VTAB frozen probes.
We sweep an 11\texttimes11 grid of learning rates and weight decay values for each probe type.
}
\label{tab:hparams-vtab}
\small
\begin{tabular}{lll}
\toprule
          & Hyperparameter    & Value \\
\midrule
Encoder   & Drop path         & \gray{0.0}          \\
\addlinespace
Data      & Input size        & 224\texttimes224    \\
          & Interpolation     & Bicubic             \\
          & Augmentations     & None                \\
\addlinespace
Training  & Loss              & Cross-entropy       \\
          & Optimizer         & AdamW               \\
          & LR schedule       & Cosine decay        \\
          & Warmup            & 5 epochs            \\
          & LR initial        & $2{\times}10^{-4}$  \\
          & LR max            & $[3{\times}10^{-5}, 3.0]$ \\
          & LR final          & 0.0                 \\
          & WD                & $[0, 3.0]$ \\
          & Batch size        & 64              \\
          & Training samples  & 1000            \\
          & Epochs            & 50              \\
\bottomrule
\end{tabular}
\end{table}

\subsubsection{Results}
\cref{tab:vtab-summary} presents category-average results across Natural, Specialized, and Structured task groups.
Among the single-view masked methods (first block), Bootleg achieves the highest overall accuracy across all three probe types at ViT-S and is best or second-best at ViT-B and ViT-L.
The gains are most pronounced on Natural tasks, where Bootleg leads by a wide margin (\eg{} 59.8\% vs.\ 47.9\% Patch at ViT-S), and on Specialized tasks, where Bootleg consistently achieves top scores.
On Structured tasks, which require spatial reasoning and counting, CrossMAE and data2vec~2.0 are more competitive. %
At ViT-L, Bootleg's X-Blk probe achieves the highest overall accuracy (68.7\%), with MAE and CrossMAE close behind on individual categories.

The grey baselines (DINOv2, Franca, CAPI), use a smaller patch size (ViT-S/14, ViT-B/14, ViT-L/14) which increases the effective sequence length and are trained on substantially larger curated datasets (LVD-142M, LAION) and/or for significantly longer (CAPI is trained on IN-1k for ten times more epochs than Bootleg).
Despite this advantage, Bootleg narrows the gap considerably on Natural and Specialized tasks.

\begin{table}[tbh]
\centering
\caption{VTAB benchmark results: category-average top-1 accuracy (\%) across Natural (7 tasks), Specialized (4 tasks), and Structured (8 tasks) categories.
We report category-averages for Patch, CLS, and X-Blk frozen probes.
MV:~multi-view (\cmark) or single-view (\xmark) pretraining.
$^{*}$:~trained by us (no official checkpoint available); all other baselines are evaluated from their official checkpoints.
}
\label{tab:vtab-summary}
\small
\adjustbox{max width=\textwidth}{%
{\setlength{\tabcolsep}{3pt}
\def\tablespec{llc@{\hspace{4\tabcolsep}}ccc@{\hspace{4\tabcolsep}}ccc@{\hspace{4\tabcolsep}}ccc@{\hspace{4\tabcolsep}}ccc}
\expandafter\tabular\expandafter{\tablespec}
\toprule
 &  & & \multicolumn{3}{c}{Natural} & \multicolumn{3}{c}{Specialized} & \multicolumn{3}{c}{Structured} & \multicolumn{3}{c}{Overall} \\
\cmidrule(r){4-6} \cmidrule(r){7-9} \cmidrule(r){10-12} \cmidrule(r){13-15}
Arch & Method & MV & Patch & CLS & X-Blk & Patch & CLS & X-Blk & Patch & CLS & X-Blk & Patch & CLS & X-Blk \\
\midrule
ViT-S/16 & MAE$^{*}$ & \xmark & 45.7 & 46.2 & 57.6 & 76.6 & \sbest{77.3} & 78.2 & 39.1 & \sbest{38.9} & 53.2 & 49.5 & 49.7 & 60.1 \\
 & CrossMAE & \xmark & \sbest{47.9} & \sbest{46.4} & \sbest{59.7} & \sbest{77.4} & 76.0 & \sbest{78.8} & \best{41.2} & \best{40.2} & \best{57.5} & \sbest{51.3} & \sbest{50.1} & \sbest{62.8} \\
 & data2vec 2.0$^{*}$ & \xmark & 38.8 & 37.4 & 48.4 & 74.6 & 75.6 & 74.1 & 40.1 & 38.4 & \sbest{54.7} & 46.9 & 45.8 & 56.5 \\
 & I-JEPA$^{*}$ & \xmark & 40.2 & \na{} & 45.8 & 74.3 & \na{} & 73.8 & 37.7 & \na{} & 43.9 & 46.3 & \na{} & 50.9 \\
\rowcolor{vlightgray}\cellcolor{white}
 & Bootleg (ours) & \xmark & \best{59.8} & \best{59.8} & \best{66.2} & \best{79.0} & \best{78.1} & \best{80.8} & \sbest{40.4} & 38.9 & 52.5 & \best{55.6} & \best{54.9} & \best{63.5} \\
\addlinespace
 & DINO & \cmark & 53.6 & \sbest{66.8} & \sbest{63.8} & \best{80.4} & \sbest{81.4} & \sbest{81.8} & \sbest{38.1} & 35.5 & \sbest{51.3} & 52.7 & \sbest{56.7} & \sbest{62.3} \\
 & iBOT & \cmark & \sbest{56.6} & \best{67.6} & \best{65.5} & \sbest{79.5} & \best{81.7} & 81.4 & \best{38.9} & \best{37.3} & \best{54.0} & \best{54.0} & \best{57.8} & \best{64.0} \\
 & MoCo-v3 & \cmark & \best{58.0} & 64.8 & 63.5 & \sbest{79.5} & 79.7 & \best{82.0} & 37.8 & \sbest{36.3} & 50.2 & \best{54.0} & 55.9 & 61.8 \\
\addlinespace
{\color{gray}ViT-S/14} & {\color{gray}DINOv2 (dst)} & {\color{gray}\cmark} & {\color{gray}\best{69.0}} & {\color{gray}\best{72.1}} & {\color{gray}\best{74.1}} & {\color{gray}\best{81.6}} & {\color{gray}\best{81.1}} & {\color{gray}\best{83.0}} & {\color{gray}\best{39.3}} & {\color{gray}\best{36.4}} & {\color{gray}\best{57.7}} & {\color{gray}\best{59.1}} & {\color{gray}\best{59.0}} & {\color{gray}\best{69.1}} \\
\midrule
ViT-B/16 & MAE & \xmark & 54.3 & 53.4 & 64.4 & 76.7 & \sbest{76.8} & 79.2 & 42.2 & 42.2 & \sbest{58.7} & 53.9 & 53.6 & 65.1 \\
 & CrossMAE & \xmark & \sbest{56.4} & \sbest{55.2} & \sbest{65.8} & \sbest{77.6} & 74.3 & \sbest{81.0} & \best{45.7} & \sbest{44.7} & \best{61.8} & \sbest{56.4} & \sbest{54.8} & \best{67.3} \\
 & data2vec 2.0 & \xmark & 52.6 & 43.9 & 54.6 & 76.6 & 71.6 & 72.4 & \sbest{43.7} & \best{44.9} & 53.8 & 53.9 & 50.2 & 58.0 \\
 & I-JEPA$^{*}$ & \xmark & 50.4 & \na{} & 55.1 & 77.2 & \na{} & 78.5 & 39.8 & \na{} & 49.8 & 51.6 & \na{} & 57.8 \\
\rowcolor{vlightgray}\cellcolor{white}
 & Bootleg (ours) & \xmark & \best{60.7} & \best{61.5} & \best{68.0} & \best{82.1} & \best{80.5} & \best{83.7} & 39.8 & 38.0 & 57.8 & \best{56.4} & \best{55.6} & \sbest{67.0} \\
\addlinespace
 & DINO & \cmark & 60.1 & \sbest{69.4} & 62.7 & \best{81.7} & \sbest{82.3} & 83.2 & 39.0 & 36.8 & 52.0 & \sbest{55.8} & 58.4 & 62.5 \\
 & iBOT & \cmark & 58.9 & 69.3 & 68.6 & \sbest{81.5} & 82.0 & 82.6 & \sbest{40.2} & \sbest{37.9} & \sbest{56.6} & \sbest{55.8} & \sbest{58.8} & \sbest{66.5} \\
 & MoCo-v3 & \cmark & \sbest{61.4} & 67.1 & \sbest{69.8} & 80.9 & \best{82.8} & \sbest{83.5} & 37.7 & 37.0 & 51.7 & 55.5 & 57.7 & 65.1 \\
 & SimDINOv2 & \cmark & \best{66.2} & \best{69.7} & \best{71.9} & \sbest{81.5} & 81.7 & \best{83.8} & \best{41.6} & \best{38.3} & \best{56.7} & \best{59.1} & \best{59.0} & \best{68.0} \\
\addlinespace
{\color{gray}ViT-B/14} & {\color{gray}Franca} & {\color{gray}\cmark} & {\color{gray}\sbest{66.5}} & {\color{gray}\sbest{71.3}} & {\color{gray}\sbest{75.1}} & {\color{gray}\sbest{82.1}} & {\color{gray}\sbest{79.7}} & {\color{gray}\sbest{83.3}} & {\color{gray}\sbest{38.6}} & {\color{gray}\sbest{32.8}} & {\color{gray}\sbest{57.3}} & {\color{gray}\sbest{58.1}} & {\color{gray}\sbest{56.9}} & {\color{gray}\sbest{69.3}} \\
 & {\color{gray}DINOv2 (dst)} & {\color{gray}\cmark} & {\color{gray}\best{72.8}} & {\color{gray}\best{74.6}} & {\color{gray}\best{78.1}} & {\color{gray}\best{83.1}} & {\color{gray}\best{81.7}} & {\color{gray}\best{84.2}} & {\color{gray}\best{41.3}} & {\color{gray}\best{37.2}} & {\color{gray}\best{58.3}} & {\color{gray}\best{61.8}} & {\color{gray}\best{60.3}} & {\color{gray}\best{71.1}} \\
\midrule
ViT-L/16 & MAE & \xmark & 56.2 & \sbest{60.1} & 65.7 & \sbest{79.4} & \sbest{80.2} & \sbest{81.6} & 45.4 & 44.7 & \sbest{62.1} & 56.6 & \best{57.9} & \sbest{67.5} \\
 & CrossMAE & \xmark & \sbest{59.1} & 57.6 & \sbest{68.4} & 78.8 & \best{80.5} & 81.4 & \sbest{46.0} & \sbest{46.5} & 54.1 & \best{57.7} & \sbest{57.8} & 65.1 \\
 & data2vec 2.0 & \xmark & 51.9 & 47.9 & 55.5 & 77.8 & 74.0 & 80.1 & \best{46.3} & \best{46.5} & \best{62.5} & 55.0 & 52.8 & 63.6 \\
 & I-JEPA$^{*}$ & \xmark & 54.0 & \na{} & 58.0 & 78.4 & \na{} & 79.1 & 38.0 & \na{} & 46.3 & 52.4 & \na{} & 57.5 \\
\rowcolor{vlightgray}\cellcolor{white}
 & Bootleg (ours) & \xmark & \best{60.7} & \best{64.9} & \best{70.9} & \best{81.9} & 79.7 & \best{84.2} & 41.5 & 39.7 & 58.9 & \sbest{57.1} & 57.4 & \best{68.7} \\
\addlinespace
 & iBOT & \cmark & \sbest{59.3} & \sbest{68.8} & \sbest{56.4} & \sbest{81.3} & \sbest{80.9} & \sbest{81.4} & \sbest{40.0} & \sbest{37.6} & \sbest{56.9} & \sbest{55.8} & \sbest{58.2} & \sbest{61.9} \\
 & SimDINOv2 & \cmark & \best{69.8} & \best{70.1} & \best{74.1} & \best{83.0} & \best{83.1} & \best{84.8} & \best{42.0} & \best{39.4} & \best{58.6} & \best{60.9} & \best{59.9} & \best{69.8} \\
\addlinespace
{\color{gray}ViT-L/14} & {\color{gray}CAPI} & {\color{gray}\xmark} & {\color{gray}62.4} & \na{} & {\color{gray}72.7} & {\color{gray}80.8} & \na{} & {\color{gray}83.5} & {\color{gray}\best{44.9}} & \na{} & {\color{gray}\best{63.1}} & {\color{gray}58.9} & \na{} & {\color{gray}70.9} \\
 & {\color{gray}Franca} & {\color{gray}\cmark} & {\color{gray}\sbest{71.2}} & {\color{gray}\sbest{68.7}} & {\color{gray}\sbest{78.6}} & {\color{gray}\sbest{82.2}} & {\color{gray}\sbest{78.9}} & {\color{gray}\sbest{84.5}} & {\color{gray}38.9} & {\color{gray}\sbest{32.2}} & {\color{gray}\sbest{60.1}} & {\color{gray}\sbest{59.9}} & {\color{gray}\sbest{55.5}} & {\color{gray}\best{72.1}} \\
 & {\color{gray}DINOv2 (dst)} & {\color{gray}\cmark} & {\color{gray}\best{72.9}} & {\color{gray}\best{75.5}} & {\color{gray}\best{79.6}} & {\color{gray}\best{83.8}} & {\color{gray}\best{84.4}} & {\color{gray}\best{84.7}} & {\color{gray}\sbest{42.5}} & {\color{gray}\best{37.7}} & {\color{gray}57.0} & {\color{gray}\best{62.4}} & {\color{gray}\best{61.5}} & {\color{gray}\sbest{71.2}} \\
\bottomrule
\endtabular
}
}
\end{table}

A break-down of results on individual datasets is shown for the patch probe in \cref{tab:vtab-patch}.

\begin{table}[tbh]
\centering
\caption{VTAB benchmark results: per-dataset top-1 accuracy (\%) using the Patch (linear) frozen probe.
C100 = CIFAR-100, Cal.\@ = Caltech-101, Flr.\@ = Flowers-102, Cam.\@ = Camelyon17, Eur.\@ = EuroSAT, Res.\@ = Resisc45, Ret.\@ = Retinopathy, Cl = Clevr, dS = dSprites, sN = SmallNORB, DM = DMLab, Kit.\@ = KITTI.
Grey: models with a smaller patch size (14\texttimes14 rather than 16\texttimes16).
MV:~multi-view (\cmark) or single-view (\xmark) pretraining.
$^{*}$:~trained by us (no official checkpoint available); all other baselines are evaluated from their official checkpoints.
}
\label{tab:vtab-patch}
\adjustbox{max width=\textwidth}{%
{\setlength{\tabcolsep}{3pt}
\def\tablespec{@{}llcccccccc@{\hspace{4\tabcolsep}}cccc@{\hspace{4\tabcolsep}}cccccccc@{}}
\expandafter\tabular\expandafter{\tablespec}
\toprule
 &  & & \multicolumn{7}{c}{Natural} & \multicolumn{4}{c}{Specialized} & \multicolumn{8}{c}{Structured} \\
\cmidrule(r){4-10} \cmidrule(r){11-14} \cmidrule(r){15-22}
 & Method & MV & C100 & Cal. & DTD & Flr. & Pets & Sun & SVH & Cam. & Eur. & Res. & Ret. & Cl\textsubscript{C} & Cl\textsubscript{D} & dS\textsubscript{L} & dS\textsubscript{O} & sN\textsubscript{A} & sN\textsubscript{E} & DM & Kit. \\
\midrule
\multicolumn{2}{l}{\textbf{ViT-S}} \\
 & MAE$^{*}$ & \xmark & 24.1 & 74.1 & 54.0 & 65.3 & 35.6 & \sbest{18.4} & 48.6 & \sbest{76.3} & \sbest{90.8} & 65.7 & \sbest{73.8} & 42.4 & 46.4 & 44.0 & \sbest{31.0} & 18.5 & \best{39.4} & 33.6 & 57.7 \\
 & CrossMAE & \xmark & \sbest{26.6} & \sbest{76.4} & \sbest{54.4} & \sbest{68.3} & 45.0 & 15.8 & \sbest{48.8} & \best{78.9} & 89.6 & \sbest{67.2} & 73.8 & \sbest{47.8} & \sbest{50.6} & 45.4 & \best{34.5} & \best{19.6} & 34.3 & 33.5 & \best{63.9} \\
 & data2vec 2.0$^{*}$ & \xmark & 20.3 & 59.9 & 47.1 & 60.4 & 24.4 & 15.2 & 44.3 & 73.5 & 88.1 & 63.4 & 73.5 & 44.3 & \best{51.9} & \best{54.4} & 18.9 & 16.5 & \sbest{39.4} & 33.2 & 62.0 \\
 & I-JEPA$^{*}$ & \xmark & 15.9 & 68.2 & 44.2 & 58.4 & \sbest{46.1} & 16.0 & 32.3 & 71.6 & 87.4 & 64.8 & 73.6 & 43.7 & 43.9 & \sbest{53.4} & 20.8 & 15.7 & 32.3 & \sbest{34.4} & 57.0 \\
\rowcolor{vlightgray}
 & Bootleg (ours) & \xmark & \best{35.8} & \best{87.4} & \best{62.9} & \best{80.2} & \best{70.9} & \best{32.0} & \best{49.1} & 73.3 & \best{92.3} & \best{75.9} & \best{74.4} & \best{49.5} & 48.0 & 40.9 & 31.0 & \sbest{18.9} & 35.1 & \best{36.2} & \sbest{63.4} \\
\addlinespace
 & DINO & \cmark & 32.0 & 74.0 & \sbest{64.6} & 78.5 & 58.4 & 23.0 & 44.6 & \best{81.8} & \best{91.4} & 73.9 & \sbest{74.5} & \sbest{45.5} & \best{49.3} & \sbest{35.3} & \best{27.9} & \best{16.4} & \sbest{33.0} & 34.9 & 62.3 \\
 & iBOT & \cmark & \sbest{36.6} & \best{80.9} & 64.3 & \sbest{79.7} & \sbest{59.8} & \best{29.1} & \sbest{45.5} & 77.5 & \sbest{91.2} & \best{74.3} & \best{74.8} & \best{50.5} & 46.9 & \best{40.5} & \sbest{25.7} & \sbest{15.8} & 32.3 & \sbest{36.5} & \sbest{63.3} \\
 & MoCo-v3 & \cmark & \best{39.5} & \sbest{79.9} & \best{65.5} & \best{82.1} & \best{63.8} & \sbest{28.7} & \best{46.3} & \sbest{81.1} & 87.9 & \best{74.3} & \sbest{74.5} & 42.3 & \sbest{49.2} & 31.7 & 24.4 & 14.3 & \best{33.8} & \best{37.0} & \best{69.9} \\
\addlinespace
 & {\color{gray}DINOv2 (dst)} & {\color{gray}\cmark} & {\color{gray}\best{54.1}} & {\color{gray}\best{89.1}} & {\color{gray}\best{74.1}} & {\color{gray}\best{97.4}} & {\color{gray}\best{77.9}} & {\color{gray}\best{46.6}} & {\color{gray}\best{43.6}} & {\color{gray}\best{82.0}} & {\color{gray}\best{91.0}} & {\color{gray}\best{78.3}} & {\color{gray}\best{75.0}} & {\color{gray}\best{48.7}} & {\color{gray}\best{48.7}} & {\color{gray}\best{34.3}} & {\color{gray}\best{38.2}} & {\color{gray}\best{19.5}} & {\color{gray}\best{32.3}} & {\color{gray}\best{35.2}} & {\color{gray}\best{57.7}} \\
\midrule
\multicolumn{2}{l}{\textbf{ViT-B}} \\
 & MAE & \xmark & 27.1 & \sbest{84.5} & \sbest{60.5} & 73.8 & 62.7 & \sbest{25.0} & 46.4 & \sbest{79.5} & 90.2 & \sbest{71.2} & 65.8 & \sbest{50.5} & 48.5 & 43.8 & \sbest{34.2} & 20.3 & 36.8 & \sbest{34.5} & \sbest{69.5} \\
 & CrossMAE & \xmark & 32.6 & 83.9 & 59.2 & \sbest{75.1} & 62.1 & 24.0 & \sbest{57.7} & 77.1 & \sbest{90.8} & 68.8 & 73.6 & 49.5 & \sbest{52.6} & \sbest{48.8} & \best{46.2} & \best{24.5} & \best{39.5} & 34.3 & \best{70.6} \\
 & data2vec 2.0 & \xmark & \sbest{32.7} & 80.3 & 56.1 & 66.3 & 44.6 & 24.1 & \best{64.5} & 75.2 & 90.2 & 67.0 & 73.9 & \best{50.7} & \best{56.5} & \best{50.0} & 29.0 & \sbest{23.9} & 37.6 & 32.6 & 69.1 \\
 & I-JEPA$^{*}$ & \xmark & 20.9 & 78.1 & 49.5 & 69.5 & \best{78.9} & 21.3 & 34.8 & 78.0 & 88.7 & 67.9 & \sbest{74.1} & 47.3 & 45.4 & 46.6 & 26.5 & 20.5 & 35.5 & 32.8 & 64.0 \\
\rowcolor{vlightgray}
 & Bootleg (ours) & \xmark & \best{35.1} & \best{87.8} & \best{64.0} & \best{82.7} & \sbest{75.2} & \best{34.9} & 45.2 & \best{82.4} & \best{92.5} & \best{78.5} & \best{74.9} & 49.0 & 48.1 & 40.6 & 33.3 & 15.8 & \sbest{39.0} & \best{36.0} & 56.8 \\
\addlinespace
 & DINO & \cmark & 37.4 & \sbest{83.7} & \sbest{68.5} & \sbest{85.7} & 70.2 & 27.6 & 47.7 & 78.1 & \best{92.9} & \sbest{81.1} & 74.6 & 45.2 & \best{50.1} & 33.9 & 30.0 & 18.9 & 33.4 & \sbest{37.7} & 62.7 \\
 & iBOT & \cmark & 42.1 & 83.5 & \sbest{68.5} & 81.2 & 58.9 & 30.1 & 48.0 & \sbest{80.2} & 92.5 & 78.3 & \best{74.8} & \best{50.2} & 49.0 & \sbest{35.0} & \sbest{30.4} & \sbest{20.1} & \sbest{35.7} & 36.7 & \sbest{64.3} \\
 & MoCo-v3 & \cmark & \sbest{45.0} & 74.8 & 68.0 & 85.3 & \sbest{72.7} & \sbest{31.6} & \sbest{52.1} & \best{80.4} & \sbest{92.8} & 78.6 & 71.7 & 47.1 & 45.5 & 33.0 & 26.9 & 16.8 & \best{36.7} & 37.5 & 57.7 \\
 & SimDINOv2 & \cmark & \best{47.7} & \best{87.7} & \best{69.6} & \best{89.7} & \best{79.8} & \best{34.7} & \best{54.3} & 76.8 & 92.7 & \best{81.7} & \sbest{74.7} & \sbest{50.0} & \sbest{49.9} & \best{38.8} & \best{31.5} & \best{20.9} & 33.7 & \best{39.7} & \best{68.4} \\
\addlinespace
 & {\color{gray}Franca} & {\color{gray}\cmark} & {\color{gray}\sbest{49.6}} & {\color{gray}\sbest{87.2}} & {\color{gray}\sbest{72.5}} & {\color{gray}\sbest{94.8}} & {\color{gray}\sbest{71.5}} & {\color{gray}\sbest{42.9}} & {\color{gray}\best{47.0}} & {\color{gray}\best{83.7}} & {\color{gray}\best{91.7}} & {\color{gray}\sbest{77.8}} & {\color{gray}\sbest{75.4}} & {\color{gray}\sbest{48.8}} & {\color{gray}\best{49.2}} & {\color{gray}\best{35.7}} & {\color{gray}\sbest{36.0}} & {\color{gray}\sbest{14.8}} & {\color{gray}\sbest{33.0}} & {\color{gray}\sbest{36.2}} & {\color{gray}\sbest{55.4}} \\
 & {\color{gray}DINOv2 (dst)} & {\color{gray}\cmark} & {\color{gray}\best{63.0}} & {\color{gray}\best{91.1}} & {\color{gray}\best{78.5}} & {\color{gray}\best{97.9}} & {\color{gray}\best{81.3}} & {\color{gray}\best{52.7}} & {\color{gray}\sbest{45.5}} & {\color{gray}\sbest{82.5}} & {\color{gray}\sbest{91.6}} & {\color{gray}\best{82.8}} & {\color{gray}\best{75.6}} & {\color{gray}\best{49.4}} & {\color{gray}\sbest{48.8}} & {\color{gray}\sbest{34.2}} & {\color{gray}\best{44.1}} & {\color{gray}\best{17.5}} & {\color{gray}\best{34.3}} & {\color{gray}\best{41.2}} & {\color{gray}\best{61.3}} \\
\midrule
\multicolumn{2}{l}{\textbf{ViT-L}} \\
 & MAE & \xmark & 28.5 & 84.8 & \sbest{62.7} & 75.5 & 62.3 & 26.4 & 53.5 & \sbest{80.7} & 91.1 & \sbest{72.9} & 73.0 & 52.6 & 49.9 & 48.1 & \sbest{43.4} & 23.8 & 39.5 & 34.5 & \best{71.6} \\
 & CrossMAE & \xmark & 36.4 & \sbest{85.5} & \best{63.0} & \sbest{76.5} & 66.6 & 27.5 & \best{58.0} & \best{81.6} & 91.8 & 68.3 & 73.4 & \sbest{52.8} & \sbest{50.3} & \best{51.9} & \best{49.4} & \sbest{24.0} & \sbest{39.8} & 33.2 & 66.4 \\
 & data2vec 2.0 & \xmark & \sbest{37.5} & 82.4 & 54.0 & 67.3 & 41.8 & 23.0 & \sbest{57.8} & 80.2 & 90.3 & 69.7 & 71.0 & \best{57.3} & \best{51.3} & \sbest{50.2} & 38.4 & \best{24.7} & \best{44.4} & \sbest{36.5} & \sbest{67.8} \\
 & I-JEPA$^{*}$ & \xmark & 29.4 & 78.9 & 54.1 & 69.6 & \best{78.8} & \sbest{28.1} & 39.5 & 76.6 & \sbest{91.9} & 71.6 & \sbest{73.5} & 43.6 & 46.0 & 44.8 & 25.3 & 20.0 & 34.2 & 33.1 & 57.0 \\
\rowcolor{vlightgray}
 & Bootleg (ours) & \xmark & \best{39.2} & \best{88.0} & 58.8 & \best{85.1} & \sbest{69.5} & \best{34.5} & 49.9 & 78.8 & \best{92.8} & \best{80.9} & \best{75.2} & 49.0 & 50.1 & 43.2 & 34.8 & 20.8 & 38.6 & \best{37.2} & 58.4 \\
\addlinespace
 & iBOT & \cmark & \sbest{44.3} & \sbest{83.5} & \sbest{68.4} & \sbest{79.2} & \sbest{61.1} & \sbest{30.5} & \sbest{48.5} & \best{83.0} & \sbest{89.5} & \sbest{78.4} & \sbest{74.3} & \sbest{47.8} & \best{49.7} & \sbest{36.5} & \sbest{26.3} & \sbest{17.7} & \best{35.9} & \best{40.6} & \sbest{65.0} \\
 & SimDINOv2 & \cmark & \best{55.1} & \best{88.1} & \best{70.5} & \best{91.7} & \best{84.3} & \best{40.7} & \best{58.3} & \sbest{80.2} & \best{93.5} & \best{83.5} & \best{74.9} & \best{49.3} & \sbest{46.3} & \best{38.1} & \best{38.0} & \best{20.0} & \sbest{34.7} & \sbest{39.7} & \best{70.2} \\
\addlinespace
 & {\color{gray}CAPI} & {\color{gray}\xmark} & {\color{gray}42.3} & {\color{gray}86.3} & {\color{gray}69.3} & {\color{gray}84.6} & {\color{gray}62.0} & {\color{gray}38.0} & {\color{gray}\best{54.6}} & {\color{gray}77.9} & {\color{gray}\sbest{92.4}} & {\color{gray}78.1} & {\color{gray}\sbest{75.0}} & {\color{gray}\sbest{51.5}} & {\color{gray}\best{49.5}} & {\color{gray}\best{41.4}} & {\color{gray}37.1} & {\color{gray}\sbest{18.9}} & {\color{gray}\best{50.2}} & {\color{gray}37.5} & {\color{gray}\best{73.6}} \\
 & {\color{gray}Franca} & {\color{gray}\cmark} & {\color{gray}\sbest{64.9}} & {\color{gray}\best{90.7}} & {\color{gray}\sbest{72.6}} & {\color{gray}\sbest{98.0}} & {\color{gray}\sbest{80.4}} & {\color{gray}\sbest{48.4}} & {\color{gray}\sbest{43.2}} & {\color{gray}\sbest{81.1}} & {\color{gray}91.0} & {\color{gray}\sbest{80.3}} & {\color{gray}\best{76.2}} & {\color{gray}43.5} & {\color{gray}\sbest{49.3}} & {\color{gray}\sbest{36.1}} & {\color{gray}\best{44.1}} & {\color{gray}\best{19.2}} & {\color{gray}\sbest{34.5}} & {\color{gray}\sbest{40.5}} & {\color{gray}44.6} \\
 & {\color{gray}DINOv2 (dst)} & {\color{gray}\cmark} & {\color{gray}\best{69.0}} & {\color{gray}\sbest{90.7}} & {\color{gray}\best{78.1}} & {\color{gray}\best{99.0}} & {\color{gray}\best{82.1}} & {\color{gray}\best{50.0}} & {\color{gray}41.2} & {\color{gray}\best{83.7}} & {\color{gray}\best{92.5}} & {\color{gray}\best{86.1}} & {\color{gray}73.0} & {\color{gray}\best{52.7}} & {\color{gray}48.6} & {\color{gray}36.0} & {\color{gray}\sbest{43.5}} & {\color{gray}18.3} & {\color{gray}33.6} & {\color{gray}\best{42.1}} & {\color{gray}\sbest{64.8}} \\
\bottomrule
\endtabular
}
}
\end{table}

\subsection{Semantic segmentation (ADE20K, Cityscapes, COCO-Stuff)}
\label{s:method-semseg}

\subsubsection{Frozen probes --- kNN}

We evaluated the pretrained models' representation of semantic information at the patch level with probes on the pretrained models using the kNN methodology from CAPI \citep{darcet2025capi}.
For this probe (``\textbf{kNN}''), images from the training partition are encoded without augmentations.
The embeddings and pixel-wise segmentation labels for each patch token form the kNN training set.
Inference is performed by finding the $k$ nearest neighbours for each patch in the image, and then for each pixel in the patch predicting the most common label across that pixel location in the $k$ neighbours.
We sweep 4 neighbourhood sizes $k \in \{1, 3, 10, 30\}$, using both cosine and L2 distance metrics.
Because the kNN training set holds the embedding and pixel labels of every training patch in memory, its memory requirement grows with the size of the training set; for COCO-Stuff, whose 118k training images are about six times as many as ADE20K's, this exceeded the RAM available on our compute nodes, so we omit the kNN probe for COCO-Stuff.

\subsubsection{Frozen probes --- Lin and Blk}

We evaluated the pretrained models' representation of semantic information at the pixel level using probes on the pretrained models with a Segmenter \citep{strudel2021segmenter} head trained atop the frozen backbone.
We consider two heads, trained separately.
``\textbf{Lin}'', a bi-linear projector from the patch embeddings to the Segmenter; or ``\textbf{Blk}'', with one randomly initialized transformer block before the projector.
Our training strategy is based on the BEiT evaluation methodology \citep{bao2021beit}; we train for 128 epochs with a batch size of 64, and augment the data with random resized cropping, horizontal flips, cut-out, and colour jitter.
For the linear projector, we halve the magnitude of the colour jitter.
For COCO-Stuff, we reduce the training duration to 20 epochs, which is approximately the same number of steps as for 128 epochs of ADE20K training.

We sweep 4 LRs for each head type, and report the best performance.
For Lin probes, we use
$\eta_\text{max} \in \{2.5{\times}10^{-4},\allowbreak 7.5{\times}10^{-4},\allowbreak 2.5{\times}10^{-3},\allowbreak 7.5{\times}10^{-3}\}$;
for Blk we use
$\eta_\text{max} \in \{2.5{\times}10^{-5},\allowbreak 7.5{\times}10^{-5},\allowbreak 2.5{\times}10^{-4},\allowbreak 7.5{\times}10^{-4}\}$.
We empirically found these LR ranges were in the optimal range for the models considered.
See \cref{tab:hparams-ade20k-probe}.

All frozen evaluation was performed using images at 224\texttimes224 resolution.
As the pretrained model uses sinusoidal position embeddings at 224\texttimes224 resolution (14\texttimes14 patches), it lacks the flexibility to correctly process other image sizes without retraining.

For the full breakdown of results, see \cref{tab:main-results-semseg}.
This shows the performance on ADE20K, Cityscapes, and COCO-Stuff with kNN (ADE20K and Cityscapes only), Bilinear (Lin), and Blk probes.

\begin{table}[tbh]
\centering
\caption{%
Hyperparameter configuration for semantic segmentation frozen probes.
We sweep 4 learning rates for each head type.
}
\label{tab:hparams-ade20k-probe}
\small
\adjustbox{max width=\textwidth}{%
\begin{tabular}{llll}
\toprule
          & Hyperparameter      & Lin & Blk \\
\midrule
Decoder   & Type                & MaskTransformer & MaskTransformer \\
          & Depth               & 0 blocks & 1 block \\
\addlinespace
Data      & Input size          & 224   & 224 \\
          & Cropping            & RandScale(0.5, 2.0) & RandScale(0.5, 2.0) \\
          & Stretch             & \gray{[1, 1]} & \gray{[1, 1]} \\
          & HFlip               & $p\!=\!0.5$ & $p\!=\!0.5$ \\
          & Cutout              & $p\!=\!0.5$ & $p\!=\!0.5$ \\
          & ColorJitter         & $p\!=\!0.5$ (half mag.) & $p\!=\!0.5$ \\
          & \quad Brightness    & 0.0625 & 0.125 \\
          & \quad Contrast      & 0.25   & 0.5 \\
          & \quad Saturation    & 0.25   & 0.5 \\
          & \quad Hue           & 0.035  & 0.07 \\
\addlinespace
Training  & Loss                & Cross-entropy    & Cross-entropy \\
          & Label smoothing     & 0.0   & 0.0 \\
          & Optimizer           & AdamW & AdamW \\
          & LR schedule         & Cosine & Cosine \\
          & Warmup              & 5 epochs  & 5 epochs \\
          & LR initial          & 0.0   & 0.0 \\
          & LR max (per 256 samp.) & $[2.5{\times}10^{-4}, 7.5{\times}10^{-3}]$ & $[2.5{\times}10^{-5}, 7.5{\times}10^{-4}]$ \\
          & LR final            & 0.0   & 0.0 \\
          & Weight decay        & 0.0001 & 0.05 \\
          & Batch size          & 64    & 64 \\
          & Epochs              & 128   & 128 \\
\bottomrule
\end{tabular}
}
\end{table}

\begin{table*}[tb]
\centering
\caption{
\textbf{Semantic segmentation results}, extending \cref{tab:main-results}.
We show mean-IoU (\%) on ADE20K, Cityscapes, and COCO-Stuff using a frozen encoder, with three decoders: a $k$-NN classifier on the patch features (kNN), a bilinear decoder (Lin; the decoder reported in \cref{tab:main-results}), and a decoder with one transformer block (Blk). $k$-NN is omitted for COCO-Stuff due to its memory requirements (\cref{s:method-semseg}).
We highlight the \best{best} and \sbest{second best} SSL method for each group of models.
First block (black): single-view masked SSL methods trained on IN-1k (MAE, CrossMAE, data2vec~2.0, I-JEPA, Bootleg).
Second block (black): multi-view joint-embedding methods trained on IN-1k (DINO, iBOT, MoCo~v3, SimDINOv2), using the official checkpoints with our evaluation protocol.
{\color{gray}Grey}:~models with a smaller patch size (14\texttimes14 rather than 16\texttimes16), and hence 30\% more patch tokens and more compute per image.
(dst): Distilled from pretrained ViT-g.
MV:~multi-view (\cmark) or single-view (\xmark) pretraining.
$^{*}$:~trained by us (no official checkpoint available); all other baselines are evaluated from their official checkpoints.
}
\label{tab:main-results-semseg}
\adjustbox{max width=\textwidth}{
{\setlength{\tabcolsep}{4pt}
\small
\begin{tabular}{llclr@{\hspace{0.5em}}ccc@{\hspace{0.5em}}ccc@{\hspace{0.5em}}cc}
\toprule
 &  & &  &  & \multicolumn{3}{c}{ADE20K mIoU} & \multicolumn{3}{c}{Cityscapes mIoU} & \multicolumn{2}{c}{COCO mIoU} \\
\cmidrule(r){6-8} \cmidrule(r){9-11} \cmidrule{12-13}
Arch & Method & MV & Data & Ep. & kNN & Lin & Blk & kNN & Lin & Blk & Lin & Blk \\
\midrule
ViT-S/16 & MAE$^{*}$ & \xmark & IN-1k & 800 & 10.2 & 14.1 & 28.7 & \sbest{26.9} & 24.7 & 29.3 & 15.6 & 28.9 \\
 & CrossMAE & \xmark & IN-1k & 800 & \sbest{10.5} & \sbest{15.2} & \sbest{31.2} & 25.8 & \sbest{25.7} & \sbest{30.8} & \sbest{15.8} & \sbest{30.4} \\
 & data2vec 2.0$^{*}$ & \xmark & IN-1k & 200 & \phantom{0}9.5 & \phantom{0}9.6 & 25.4 & 23.3 & 22.2 & 25.8 & \phantom{0}9.5 & 25.2 \\
 & I-JEPA$^{*}$ & \xmark & IN-1k & 600 & \phantom{0}8.2 & 11.8 & 21.1 & 17.9 & 19.8 & 24.5 & 13.8 & 22.3 \\
 & Bootleg (ours) & \xmark & IN-1k & 600 & \best{23.5} & \best{26.6} & \best{33.9} & \best{29.4} & \best{32.1} & \best{35.1} & \best{28.4} & \best{34.6} \\
\cmidrule(r){2-5}
 & DINO & \cmark & IN-1k & 800 & 16.2 & 20.4 & \sbest{34.4} & \sbest{29.5} & 25.4 & \sbest{31.5} & 20.8 & \sbest{35.4} \\
 & iBOT & \cmark & IN-1k & 800 & \best{20.5} & \best{26.9} & \best{36.0} & \best{30.4} & \best{28.8} & \best{33.6} & \best{28.8} & \best{36.0} \\
 & MoCo-v3 & \cmark & IN-1k & 300 & \sbest{18.3} & \sbest{23.4} & 32.4 & 29.1 & \sbest{25.6} & 28.9 & \sbest{24.5} & 32.6 \\
\cmidrule(r){2-5}
{\color{gray}ViT-S/14} & {\color{gray}DINOv2 (dst)} & {\color{gray}\cmark} & {\color{gray}LVD} & {\color{gray}Unk.} & {\color{gray}\best{34.0}} & {\color{gray}\best{39.7}} & {\color{gray}\best{42.7}} & {\color{gray}\best{39.0}} & {\color{gray}\best{46.4}} & {\color{gray}\best{47.1}} & {\color{gray}\best{37.5}} & {\color{gray}\best{41.3}} \\
\midrule
ViT-B/16 & MAE & \xmark & IN-1k & 1600 & 17.6 & \sbest{24.7} & 35.8 & \sbest{28.0} & 30.5 & 34.9 & \sbest{26.1} & \sbest{36.3} \\
 & CrossMAE & \xmark & IN-1k & 800 & 14.9 & 24.1 & \sbest{36.9} & 27.4 & \sbest{31.2} & \sbest{36.2} & 24.6 & 36.0 \\
 & data2vec 2.0 & \xmark & IN-1k & 200 & \sbest{17.6} & 22.5 & 34.7 & 26.6 & 27.8 & 32.5 & 22.6 & 34.6 \\
 & I-JEPA$^{*}$ & \xmark & IN-1k & 600 & 13.1 & 19.3 & 28.8 & 20.3 & 24.3 & 29.8 & 20.4 & 29.9 \\
 & Bootleg (ours) & \xmark & IN-1k & 600 & \best{25.1} & \best{30.9} & \best{38.8} & \best{30.4} & \best{35.9} & \best{39.7} & \best{32.2} & \best{38.5} \\
\cmidrule(r){2-5}
 & DINO & \cmark & IN-1k & 400 & 19.5 & 27.1 & 36.6 & 31.1 & 30.5 & 33.6 & 27.8 & 37.2 \\
 & iBOT & \cmark & IN-1k & 400 & \best{23.9} & \best{32.6} & \best{39.4} & \best{32.8} & \best{33.8} & \best{37.2} & \best{33.2} & \best{38.4} \\
 & MoCo-v3 & \cmark & IN-1k & 300 & 21.5 & 29.1 & 36.1 & \sbest{31.5} & 27.7 & 31.2 & 29.6 & 35.7 \\
 & SimDINOv2 & \cmark & IN-1k & 100 & \sbest{23.0} & \sbest{31.4} & \sbest{38.8} & 31.0 & \sbest{32.7} & \sbest{35.6} & \sbest{32.0} & \sbest{37.7} \\
\cmidrule(r){2-5}
{\color{gray}ViT-B/14} & {\color{gray}Franca} & {\color{gray}\cmark} & {\color{gray}IN-22k} & {\color{gray}(90)} & {\color{gray}\sbest{30.4}} & {\color{gray}\sbest{38.0}} & {\color{gray}\sbest{43.1}} & {\color{gray}\sbest{38.3}} & {\color{gray}\sbest{46.1}} & {\color{gray}\sbest{49.8}} & {\color{gray}\sbest{38.4}} & {\color{gray}\sbest{42.9}} \\
{\color{gray}} & {\color{gray}DINOv2 (dst)} & {\color{gray}\cmark} & {\color{gray}LVD} & {\color{gray}Unk.} & {\color{gray}\best{36.0}} & {\color{gray}\best{42.8}} & {\color{gray}\best{45.8}} & {\color{gray}\best{41.4}} & {\color{gray}\best{49.6}} & {\color{gray}\best{51.1}} & {\color{gray}\best{40.1}} & {\color{gray}\best{44.6}} \\
\midrule
ViT-L/16 & MAE & \xmark & IN-1k & 1600 & 20.0 & 28.8 & 40.9 & 27.2 & 34.7 & 38.3 & 28.8 & 40.4 \\
 & CrossMAE & \xmark & IN-1k & 800 & 18.2 & 28.8 & 39.8 & 28.6 & \sbest{36.7} & 39.5 & \sbest{28.8} & 39.4 \\
 & data2vec 2.0 & \xmark & IN-1k & 200 & \sbest{23.4} & \sbest{30.0} & \best{43.0} & \sbest{28.7} & 35.5 & \best{44.3} & 27.7 & \best{41.4} \\
 & I-JEPA$^{*}$ & \xmark & IN-1k & 600 & 15.6 & 21.8 & 28.6 & 18.4 & 25.5 & 30.4 & 23.6 & 30.8 \\
 & Bootleg (ours) & \xmark & IN-1k & 600 & \best{27.2} & \best{34.7} & \sbest{41.2} & \best{29.1} & \best{39.1} & \sbest{42.8} & \best{34.4} & \sbest{40.5} \\
\cmidrule(r){2-5}
 & iBOT & \cmark & IN-1k & 250 & \sbest{24.7} & \sbest{33.2} & \sbest{40.4} & \sbest{33.1} & \sbest{35.0} & \sbest{36.2} & \sbest{34.0} & \sbest{40.0} \\
 & SimDINOv2 & \cmark & IN-1k & 100 & \best{30.4} & \best{36.6} & \best{41.6} & \best{36.6} & \best{37.5} & \best{40.8} & \best{36.4} & \best{40.6} \\
\cmidrule(r){2-5}
{\color{gray}ViT-L/14} & {\color{gray}CAPI} & {\color{gray}\xmark} & {\color{gray}IN-1k} & {\color{gray}6394} & {\color{gray}28.6} & {\color{gray}38.9} & {\color{gray}46.1} & {\color{gray}32.9} & {\color{gray}43.7} & {\color{gray}48.1} & {\color{gray}37.4} & {\color{gray}44.0} \\
{\color{gray}} & {\color{gray}Franca} & {\color{gray}\cmark} & {\color{gray}LAION} & {\color{gray}(3.2)} & {\color{gray}\sbest{33.6}} & {\color{gray}\sbest{41.7}} & {\color{gray}\sbest{46.4}} & {\color{gray}\sbest{38.5}} & {\color{gray}\sbest{48.1}} & {\color{gray}\sbest{50.6}} & {\color{gray}\best{40.3}} & {\color{gray}\sbest{44.8}} \\
{\color{gray}} & {\color{gray}DINOv2 (dst)} & {\color{gray}\cmark} & {\color{gray}LVD} & {\color{gray}Unk.} & {\color{gray}\best{34.6}} & {\color{gray}\best{43.4}} & {\color{gray}\best{47.3}} & {\color{gray}\best{42.4}} & {\color{gray}\best{50.9}} & {\color{gray}\best{52.5}} & {\color{gray}\sbest{40.2}} & {\color{gray}\best{45.1}} \\
\bottomrule
\end{tabular}
}
}
\end{table*}

\FloatBarrier
\section{Additional evaluations}
\label{a:extra-evals}

\subsection{Variance across random seeds}
\label{s:variance-over-rng}

We ran experiments with three different seeds for model initialization and dataloader to investigate the stability of our model's performance and significance of the results.
For this experiment, all models were trained for 300 epochs.
As shown in \cref{tab:seed-variance}, the differences between methods are significant, and Bootleg has much more consistency in its frozen probe performance than I-JEPA, indicating our training method is more stable.

\begin{table}[tbh]
\centering
\caption{Variance under change in random seed. We trained MAE, I-JEPA, and Bootleg (final hparams) ViT-S models for 300 ep. on IN-1k with three random seeds for both init and dataloader. Results are shown as mean $\pm$ stdev.}
\label{tab:seed-variance}
\small
\adjustbox{max width=\textwidth}{%
\begin{tabular}{lcccccc}
\toprule
 & \multicolumn{3}{c}{IN-1k acc.} & \multicolumn{3}{c}{ADE20K mIoU} \\
\cmidrule(r){2-4} \cmidrule(l){5-7}
Method & Patch & CLS & X-Blk & kNN & Lin & Blk \\
\midrule
MAE & 43.2 \tpm{0.2} & \sbest{35.9} \tpm{1.1} & \sbest{64.5} \tpm{0.0} & \phantom{0}\sbest{9.6} \tpm{0.3} & \sbest{13.5} \tpm{0.2} & \sbest{26.8} \tpm{0.3} \\
I-JEPA & \sbest{51.5} \tpm{2.9} & \na{} & 62.1 \tpm{2.3} & \phantom{0}9.0 \tpm{1.9} & 12.7 \tpm{2.2} & 22.4 \tpm{2.1} \\
\rowcolor{vlightgray}
Bootleg (ours) & \best{67.2} \tpm{0.6} & \best{68.3} \tpm{0.6} & \best{74.0} \tpm{0.3} & \best{21.8} \tpm{0.3} & \best{26.3} \tpm{0.3} & \best{33.8} \tpm{0.8} \\
\bottomrule
\end{tabular}
}
\end{table}

\FloatBarrier

\subsection{I-JEPA linear-probe protocol}
\label{s:ijepa-linprobe}

Our frozen-probe protocol (\cref{a:eval-method-classification-probe}) differs from the linear evaluation used by I-JEPA \citep{assran2023ijepa}, which leaves open whether the gap between I-JEPA and Bootleg is partly an artefact of the evaluation.
We therefore re-implemented I-JEPA's linear-evaluation protocol (their App.~A.2) and evaluated every model in \cref{tab:main-results} with it, together with the official I-JEPA ViT-H/14 checkpoint.
Following I-JEPA, we train for 50 epochs with batch size 16384 using LARS, with the learning rate divided by 10 every 15 epochs and minimal augmentation (random resized crop and horizontal flip only).
We sweep the same grid as I-JEPA in parallel over a single frozen-encoder pass: representation (last-layer average-pool, or the concatenation of the last 4 layers), head (linear, or non-affine BatchNorm followed by linear), reference learning rate $\{0.01, 0.05, 0.001\}$ (scaled linearly with batch size, following the MAE convention which I-JEPA's reference values adopt), and weight decay $\{0.0005, 0\}$, and report the best of the 24 heads.
I-JEPA probes patch-averaged features because its encoder has no CLS token; for models with a CLS token we also report the same sweep on CLS-token readouts (``CLS'').

To validate the implementation, we evaluated checkpoints for which I-JEPA reports results with this protocol.
Our numbers reproduce the published values for MAE ViT-B/16 (68.9 vs.\ 68.0) and ViT-L/16 (76.9 vs.\ 76.0), and the official I-JEPA ViT-H/14 checkpoint (79.1 vs.\ 79.3) (\cref{tab:ijepa-linprobe}).

Under I-JEPA's own protocol, our I-JEPA reproductions score 59.2\%, 69.5\%, and 69.7\% (70.2\% for a second seed) for ViT-S/B/L, versus 71.7\%, 76.9\%, and 79.2\% for Bootleg.
That is, the advantage of Bootleg over I-JEPA is not a product of our evaluation protocol: it persists (+12.5, +7.4, and +9.0 percentage points at ViT-S/B/L) when both are evaluated in the manner I-JEPA was designed for.
The I-JEPA ViT-B/L numbers are also consistent with those obtained with our own probe in \cref{tab:main-results}. For ViT-B/16 and ViT-L/16, \citet{assran2023ijepa} report 72.9\% and 77.5\% with their own (unreleased) 600-epoch models, 3.4 and 7.3--7.8 percentage points above our reproductions; since our pipeline reproduces the published values for MAE and for the released I-JEPA ViT-H/14 checkpoint, this gap is not due to the evaluation pipeline, and Bootleg exceeds even the reported I-JEPA values at both sizes (76.9\% and 79.2\%).

\begin{table}[tbh]
\centering
\caption{%
IN-1k top-1 accuracy (\%) using the linear-evaluation protocol of I-JEPA \citep{assran2023ijepa} (LARS, 50 ep., best of 24 heads; see \cref{s:ijepa-linprobe}).
Patch: average-pooled patch tokens (I-JEPA's readout). CLS: CLS-token readout. Reported: value reported by \citet{assran2023ijepa} (their Table~1) with this protocol, for the same checkpoint, or, for I-JEPA ViT-B/16 and ViT-L/16, for the authors' unreleased 600-epoch models whose recipe our reproductions follow.
\na{}: model has no CLS token. \missing{}: not reported.
Block structure as in \cref{tab:main-results}; we highlight the \best{best} and \sbest{second best} within each block.
MV:~multi-view (\cmark) or single-view (\xmark) pretraining.
$^{*}$:~trained by us (no official checkpoint available); all other baselines are evaluated from their official checkpoints.
}
\label{tab:ijepa-linprobe}
\small
\begin{tabular}{llcrrr}
\toprule
Arch & Method & MV & Patch & CLS & Reported \\
\midrule
ViT-S/16 & MAE$^{*}$ & \xmark & 50.5 & \sbest{53.0} & \missing{} \\
 & CrossMAE & \xmark & 56.3 & 51.9 & \missing{} \\
 & data2vec 2.0$^{*}$ & \xmark & 48.0 & 45.2 & \missing{} \\
 & I-JEPA$^{*}$ & \xmark & \sbest{59.2} & \na{} & \missing{} \\
 & Bootleg (ours) & \xmark & \best{70.9} & \best{71.7} & \missing{} \\
\cmidrule(r){2-3}
 & DINO & \cmark & 73.1 & 76.4 & \missing{} \\
 & iBOT & \cmark & \sbest{73.3} & \best{77.4} & \missing{} \\
 & MoCo~v3 & \cmark & 69.0 & 73.7 & \missing{} \\
\cmidrule(r){2-3}
{\color{gray}ViT-S/14} & {\color{gray}DINOv2 (dst)} & {\color{gray}\cmark} & {\color{gray}\best{78.8}} & {\color{gray}\best{81.1}} & \missing{} \\
\midrule
ViT-B/16 & MAE & \xmark & 68.9 & \sbest{68.9} & 68.0 \\
 & CrossMAE & \xmark & 67.7 & 64.3 & \missing{} \\
 & data2vec 2.0 & \xmark & 68.1 & 67.8 & \missing{} \\
 & I-JEPA$^{*}$ & \xmark & \sbest{69.5} & \na{} & 72.9 \\
 & Bootleg (ours) & \xmark & \best{76.1} & \best{76.9} & \missing{} \\
\cmidrule(r){2-3}
 & DINO & \cmark & 74.8 & 77.7 & \missing{} \\
 & iBOT & \cmark & \sbest{75.9} & \sbest{79.3} & \missing{} \\
 & MoCo~v3 & \cmark & 73.6 & 76.5 & \missing{} \\
 & SimDINOv2 & \cmark & \best{76.8} & \best{79.6} & \missing{} \\
\cmidrule(r){2-3}
{\color{gray}ViT-B/14} & {\color{gray}Franca} & {\color{gray}\cmark} & {\color{gray}\sbest{79.6}} & {\color{gray}\sbest{82.3}} & \missing{} \\
 & {\color{gray}DINOv2 (dst)} & {\color{gray}\cmark} & {\color{gray}\best{82.5}} & {\color{gray}\best{84.3}} & \missing{} \\
\midrule
ViT-L/16 & MAE & \xmark & 75.0 & 76.9 & 76.0 \\
 & CrossMAE & \xmark & 72.3 & 71.1 & \missing{} \\
 & data2vec 2.0 & \xmark & \sbest{77.9} & \sbest{77.5} & \missing{} \\
 & I-JEPA$^{*}$ & \xmark & 69.7 & \na{} & \missing{} \\
 & I-JEPA (seed 0)$^{*}$ & \xmark & 70.2 & \na{} & 77.5 \\
 & Bootleg (ours) & \xmark & \best{78.0} & \best{79.2} & \missing{} \\
\cmidrule(r){2-3}
 & iBOT & \cmark & \sbest{77.3} & \sbest{80.9} & 81.0 \\
 & SimDINOv2 & \cmark & \best{80.7} & \best{82.3} & \missing{} \\
\cmidrule(r){2-3}
{\color{gray}ViT-L/14} & {\color{gray}CAPI} & {\color{gray}\xmark} & {\color{gray}79.1} & {\color{gray}\na{}} & \missing{} \\
 & {\color{gray}Franca} & {\color{gray}\cmark} & {\color{gray}\sbest{83.1}} & {\color{gray}\sbest{84.6}} & \missing{} \\
 & {\color{gray}DINOv2 (dst)} & {\color{gray}\cmark} & {\color{gray}\best{84.6}} & {\color{gray}\best{86.1}} & \missing{} \\
\midrule
ViT-H/14 & I-JEPA (official) & \xmark & \best{79.1} & \na{} & 79.3 \\
\bottomrule
\end{tabular}
\end{table}

\FloatBarrier
\subsection{Fine-tuned models}
\label{s:results-ft}

We fine-tuned SSL-pretrained models at IN-1k classification using fine-tuning on 100\% of the training data (full fine-tuning), 1\% of the data (low-shot, full fine-tuning), and 1\% of the data using LoRA (low-shot, low-rank fine-tuning), as described in \cref{s:in1k-ft}.
We additionally fine-tuned on ADE20K segmentation, as described in \cref{s:ade20k-ft}.

\subsubsection{IN-1k fine-tuning method}
\label{s:in1k-ft}

We fine-tuned the pretrained models on IN-1k using either the full labels, or a 1\% subset of the labels as described below.

Our methodology for full fine-tuning on IN-1k follows that recommended in the GitHub repository of MAE \citep{he2022masked} and used by CrossMAE \citep{fu2025crossmae}: heavy augmentation (RandAugment, Mixup, Cutmix, RandomErasure, drop path), average patch embeddings with layer norm and linear head, label-smoothed cross-entropy, LR layer decay of 0.65 (ViT-S,-B) or 0.75 (ViT-L), trained for 50 (ViT-L) or 100 epochs (ViT-S,-B).
The fine-tuning hyperparameters are detailed in \cref{tab:hparams-in1k-ft}.

\begin{table*}[tbh]
\centering
\caption{%
Hyperparameter configuration for IN-1k classification fine-tuning.
Learning rate (LR) reference values are shown relative to a batch size of 256 and linearly scaled for the actual batch size.
For 1\% fine-tuning, we sweep five LR values and report the best; the swept ranges are shown.
}
\label{tab:hparams-in1k-ft}
\small
\adjustbox{max width=\textwidth}{%
\begin{tabular}{llllll}
\toprule
          &                     & \multicolumn{2}{c}{100\% Full FT} & \multicolumn{2}{c}{1\% Low-shot FT} \\
\cmidrule(r){3-4} \cmidrule(r){5-6}
          & Hyperparameter      & ViT-S,-B & ViT-L & Full & LoRA \\
\midrule
Encoder   & Drop path           & 0.1   & 0.2   & \gray{0.0}  & \gray{0.0} \\
\addlinespace
Data      & Input size          & 224   & 224   & 224  & 224 \\
          & Interpolation       & Bicubic & Bicubic & Bicubic & Bicubic \\
          & HFlip               & $p\!=\!0.5$ & $p\!=\!0.5$ & $p\!=\!0.5$ & $p\!=\!0.5$ \\
          & Transforms          & RandAug & RandAug & RandAug & RandAug \\
          & RandErasure         & $p\!=\!0.25$ & $p\!=\!0.25$ & \gray{$p\!=\!0$} & \gray{$p\!=\!0$} \\
          & Subset              & 100\% & 100\% & 1\%  & 1\% \\
\addlinespace
Training  & Loss                & Cross-entropy & Cross-entropy & Cross-entropy & Cross-entropy \\
          & Label smoothing     & 0.1   & 0.1   & 0.1  & 0.1 \\
          & MixUp               & 0.8   & 0.8   & \gray{0.0}  & \gray{0.0} \\
          & MixUp / CutMix      & 1.0   & 1.0   & \gray{0.0}  & \gray{0.0} \\
          & Optimizer           & AdamW & AdamW & AdamW & AdamW \\
          & LR schedule         & Cosine & Cosine & Cosine & Cosine \\
          & Warmup              & 5 ep  & 5 ep  & 5 ep & 5 ep \\
          & LR initial          & 0.0   & 0.0   & 0.0  & 0.0 \\
          & LR max (per 256 samp.) & $5{\times}10^{-4}$ & $1{\times}10^{-3}$ & $[5{\times}10^{-6},\, 5{\times}10^{-4}]$ & $[1.5{\times}10^{-6},\, 1.5{\times}10^{-3}]$ \\
          & LR final            & $2.5{\times}10^{-7}$ & $2.5{\times}10^{-7}$ & 0.0 & 0.0 \\
          & LR layer decay      & 0.65  & 0.75  & 0.75 & 1.0 \\
          & Weight decay        & 0.05  & 0.05  & 0.05 & 0.05 \\
          & Batch size          & 1024  & 1024  & 512  & 512 \\
          & Epochs              & 100   & 50    & 50   & 50 \\
\addlinespace
LoRA      & Rank                & \gray{full} & \gray{full} & \gray{full}  & 8 \\
          & $\alpha$            & \na{} & \na{} & \na{}  & 16 \\
          & Dropout             & \na{} & \na{} & \na{}  & 0.05 \\
          & Targets             & \gray{all} & \gray{all} & \gray{all}  & qkv, proj, mlp \\
          & Train norms         & \gray{\cmark{}} & \gray{\cmark{}} & \gray{\cmark{}} & \cmark{} \\
\bottomrule
\end{tabular}
}
\end{table*}

We performed low-shot fine-tuning on 1\% of IN-1k using the subset proposed by SimCLR \citep{chen2020simple} and the methodology from I-JEPA \citep{assran2023ijepa}.
We adapted the methodology to sweep over five maximum learning rates, $\eta_\text{max} \in \{5{\times}10^{-6},\allowbreak 1.5{\times}10^{-5},\allowbreak 5{\times}10^{-5},\allowbreak 1.5{\times}10^{-4},\allowbreak 5{\times}10^{-4}\}$, and report the best performance over these learning rates.

We additionally evaluated the models at low-rank (parameter efficient) fine-tuning.
For this we use the LoRA methodology \citep{Hu2022_LoRA} with $\text{rank} = 8$ and $\alpha = 16$.
We fine-tune the attention QKV weights, attention projector, and MLP weights from every block on the 1\% low-shot subset of IN-1k.
Learning rates were swept over $\eta_\text{max} \in \{1.5{\times}10^{-5},\allowbreak 5{\times}10^{-5},\allowbreak 1.5{\times}10^{-4},\allowbreak 5{\times}10^{-4},\allowbreak 1.5{\times}10^{-3}\}$, and we report the best performance over these learning rates.

All fine-tuned models were evaluated on IN-1k validation set, with an 87.5\% centre crop (resize to 256\texttimes 256, then crop to 224\texttimes 224).
Results are shown in \cref{tab:finetuning}.

\subsubsection{ADE20K fine-tuning method}
\label{s:ade20k-ft}

For ADE20K fine-tuning, we add a single-block MaskTransformer decoder \citep{strudel2021segmenter} to the encoder.
Both the encoder and decoder are fine-tuned end-to-end for 128 epochs with a batch size of 32, using the AdamW optimizer.
Training hyperparameters are detailed in \cref{tab:hparams-ade20k-ft}.
The hyperparameters used, including learning rate, were selected on a prototyping model and not tuned for our final model(s) on which we report the final results.

Fine-tuning evaluation was performed using images at 512\texttimes512 resolution.
We interpolate the position embeddings to accommodate the processing of 512\texttimes512 px images, a change which the model is quickly able to adapt to with fine-tuning.

For evaluation, images are simply resized to 512\texttimes512, with no other transformations applied.

\begin{table}[tbh]
\centering
\caption{%
Hyperparameter configuration for ADE20K segmentation fine-tuning.
Learning rate (LR) reference values are shown relative to a batch size of 256 and linearly scaled for the actual batch size (32).
}
\label{tab:hparams-ade20k-ft}
\small
\begin{tabular}{llll}
\toprule
          & Hyperparameter      & ViT-S, -B & ViT-L \\
\midrule
Encoder   & Drop path           & 0.1   & 0.2 \\
\addlinespace
Decoder   & Type                & MaskTransformer & MaskTransformer \\
          & Depth               & 1 block & 1 block \\
\addlinespace
Data      & Input size          & 512   & 512 \\
          & Cropping            & RandScale(0.5, 2.0) & RandScale(0.5, 2.0) \\
          & Stretch             & \gray{[1, 1]} & \gray{[1, 1]} \\
          & HFlip               & $p\!=\!0.5$ & $p\!=\!0.5$ \\
          & Cutout              & $p\!=\!0.5$ & $p\!=\!0.5$ \\
          & ColorJitter         & $p\!=\!0.5$ & $p\!=\!0.5$ \\
\addlinespace
Training  & Loss                & Cross-entropy    & Cross-entropy \\
          & Optimizer           & AdamW & AdamW \\
          & LR schedule         & Cosine & Cosine \\
          & Warmup              & 10 epochs & 10 epochs \\
          & LR initial          & 0.0   & 0.0 \\
          & LR max (per 256 samp.) & $7.5{\times}10^{-3}$ & $2.5{\times}10^{-3}$ \\
          & LR final            & 0.0   & 0.0 \\
          & LR layer decay      & 0.65  & 0.75 \\
          & Weight decay        & 0.05  & 0.05 \\
          & Batch size          & 32    & 32 \\
          & Epochs              & 128   & 128 \\
\bottomrule
\end{tabular}
\end{table}

\subsubsection{Results}
As shown in \cref{tab:finetuning}, Bootleg achieves the best or near-best fine-tuning performance across all settings.
On IN-1k 100\% fine-tuning, Bootleg outperforms all baselines at every model size, with particularly large margins at ViT-S (+0.9 over the next best) and ViT-L (+0.7).
The advantage is most pronounced in the low-data regime: on 1\% fine-tuning, Bootleg surpasses the next-best method by +8.6 (ViT-B) and +12.5 (ViT-S) points, demonstrating that hidden-self-distillation produces representations that are especially effective when labelled data is scarce.
LoRA fine-tuning follows a similar pattern, with Bootleg matching or exceeding full fine-tuning of I-JEPA despite updating far fewer parameters.

On ADE20K fine-tuning, Bootleg leads at ViT-S and ViT-B; at ViT-L, data2vec~2.0 and MAE are stronger.

Data2vec~2.0 full fine-tuning was sometimes unstable at the learning rates used for these experiments (collapsed model at 0.1\% acc.) and would benefit from a reduced learning rate.

\begin{table}[tbh]
\centering
\caption{%
Fine-tuning results on IN-1k classification (top-1 accuracy, \%) and ADE20K semantic segmentation (mIoU, \%).
IN-1k results include full fine-tuning on 100\% and 1\% of training data, and LoRA fine-tuning on 1\%.
Ep: number of pretraining epochs.
}
\label{tab:finetuning}
\small
\adjustbox{max width=\textwidth}{%
\begin{tabular}{lllrcccc}
\toprule
 &  &  &  & \multicolumn{3}{c}{IN-1k acc.} & \multicolumn{1}{c}{ADE20K mIoU} \\
\cmidrule(r){5-7} \cmidrule(r){8-8}
 &  &  &  & \multicolumn{2}{c}{Full-FT} & \multicolumn{1}{c}{LoRA} & \multicolumn{1}{c}{Full-FT} \\
\cmidrule(r){5-6} \cmidrule(r){7-7} \cmidrule(r){8-8}
Arch & Method & Data & Ep. & 100\% & 1\% & 1\% & 100\% \\
\midrule
ViT-S/16 & MAE & IN-1k & 800 & 78.2 & 37.1 & 33.9 & 39.2 \\
 & CrossMAE & IN-1k & 800 & 79.8 & 40.4 & 37.5 & \sbest{42.2} \\
 & data2vec 2.0 & IN-1k & 200 & 79.9 & 39.3 & 36.8 & 40.5 \\
 & I-JEPA & IN-1k & 600 & \sbest{79.9} & \sbest{48.6} & \sbest{47.3} & 36.1 \\
\rowcolor{vlightgray}\cellcolor{white}
 & Bootleg (ours) & IN-1k & 600 & \best{80.8} & \best{61.1} & \best{60.4} & \best{44.3} \\
\midrule
ViT-B/16 & MAE & IN-1k & 1600 & 82.7 & 55.9 & 53.5 & 44.0 \\
 & CrossMAE & IN-1k & 800 & 82.9 & 52.3 & 49.6 & \sbest{45.0} \\
 & data2vec 2.0 & IN-1k & 200 & \sbest{83.3} & 58.9 & 58.1 & \gray{\phantom{0}0.1} \\
 & I-JEPA & IN-1k & 600 & 82.7 & \sbest{61.9} & \sbest{62.2} & 29.8 \\
\rowcolor{vlightgray}\cellcolor{white}
 & Bootleg (ours) & IN-1k & 600 & \best{83.9} & \best{70.5} & \best{69.9} & \best{46.6} \\
\midrule
ViT-L/16 & MAE & IN-1k & 1600 & \sbest{84.7} & 67.7 & 67.5 & \sbest{50.7} \\
 & CrossMAE & IN-1k & 800 & 84.3 & 62.7 & 61.7 & 49.6 \\
 & data2vec 2.0 & IN-1k & 200 & \gray{\phantom{0}0.1} & \sbest{73.2} & \best{74.1} & \best{51.8} \\
 & I-JEPA & IN-1k & 600 & 82.0 & 63.8 & 63.8 & 31.9 \\
\rowcolor{vlightgray}\cellcolor{white}
 & Bootleg (ours) & IN-1k & 600 & \best{85.4} & \best{73.4} & \sbest{73.2} & 48.3 \\
\bottomrule
\end{tabular}
}
\end{table}

\FloatBarrier
\section{Additional ablations}

\subsection{Interpreting the I-JEPA-to-Bootleg ablation}
\label{s:config-interpolation}

\Cref{tab:config-interpolation} departs from the conventional ablation, in which changes are stacked one after another and each is credited with the gain observed when it is added.
Because gains in accuracy shrink as improvements accumulate, that convention favours whichever change is added first, and its attributions depend on the chosen order.
We therefore measure each component at the two extremes: added first to the I-JEPA configuration (``Only with''), and added last to complete the Bootleg configuration (``Only without''); the right-most column averages the resulting deltas.
By this measure, the masking improvements and the multi-layer targets tie for the largest effect (+3.6\,pp each).
The multi-layer targets give the largest gain of any component when added first (+7.0\,pp IN-1k X-Blk), but a small one when added last (+0.4\,pp).
This is consistent with their role as a stabilizer: Bootleg is less sensitive than I-JEPA to the training seed, the per-GPU batch size, and the masking and optimizer hyperparameters, so once the masking implementation is fixed and the other hyperparameters are tuned, there is less instability left for the multi-layer targets to remove.

\subsection{Preliminary hyperparameters}
\label{s:prelim}

As described in \cref{s:pretraining-hparams}, our hyperparameter sweep was conducted in multiple rounds.
Some ablations were conducted using a prototype hyperparameter configuration, before the final hyperparameters were established.
The prototype and final configurations are compared in \cref{tab:hparams-prelim}.

Additionally, note that tables using prototype hyperparameter values use a weaker implementation of the IN-1k probe methodology (too large a range in crop sizes, too small a range in LR values), and are not comparable with the main tables but are self-consistent.

\begin{table*}[tbh]
\centering
\caption{Preliminary and final ablation hyperparameter configurations, each used for some of the ablation experiments presented (ViT-S, 300 ep.).}
\label{tab:hparams-prelim}
\small
\adjustbox{max width=\textwidth}{
\begin{tabular}{llll}
\toprule
          & Hyperparameter    & Bootleg (prototype)           & Bootleg (final)               \\
\midrule
Encoder   & Architecture      & ViT-S/16                      & ViT-S/16                      \\
          & Depth             & 12                            & 12                            \\
          & Width             & 384                           & 384                           \\
          & Attention heads   & 6                             & 6                             \\
          & Patch size        & 16                            & 16                            \\
          & CLS tokens        & 1                             & {1}                           \\
          & Register tokens   & 4                             & {4}                           \\
\addlinespace
Predictor & Depth             & 8                             & {10}                          \\
          & Width             & 256                           & {192}                         \\
          & Attention heads   & 8                             & {16}                          \\
          & Register tokens   & 4                             & {4}                           \\
\addlinespace
Data      & Transforms        & RandCrop(0.2, 1.0)\,+\,Hflip  & RandCrop(0.35, 1.0)\,+\,Hflip \\
          & Interpolation     & Bicubic                       & Bicubic                       \\
          & Input size        & 224\texttimes 224             & 224\texttimes 224             \\
\addlinespace
Masking   & Strategy          & {4 rectangular blocks}        & {4 rectangular blocks}        \\
          & Mask seen rate    & 0.292 $\pm$ 0.069             & 0.292 $\pm$ 0.069             \\
          & Mask target rate  & 0.475 $\pm$ 0.076             & 0.475 $\pm$ 0.076             \\
\addlinespace
Training  & Target(s)         & Block 1, 4, 8, 12 outputs     & Block 1, 4, 8, 12 outputs     \\
     & Target standardization & Z-score (patch \& feat. dims) & Z-score (patch \& feat. dims) \\
          & Target length     & 1536                          & 1536                          \\
          & Loss              & {Mean squared error}          & {Mean squared error}          \\
          & Optimizer         & AdamW({\footnotesize$\beta\!=\!(0.9, 0.95)$}) & AdamW({\footnotesize$\beta\!=\!(0.9, 0.95)$}) \\
          & EMA initial       & 0.998                         & 0.9985                        \\
          & EMA final         & 1.0                           & 0.9985                        \\
          & LR schedule       & Cosine annealing              & Cosine annealing              \\
          & LR schedule warmup& 40 epochs                     & 69 epochs                     \\
          & LR initial        & 0.0                           & 0.0001                        \\
          & LR maximum        & 0.0012                        & {0.003}                       \\
          & LR final          & 0.0                           & 0.0001                        \\
          & WD initial        & 0.05                          & 0.05                          \\
          & WD final          & 0.05                          & {0.05}                        \\
          & Schedule stretch  & 1.25                          & 1.0                           \\
          & Batch size        & 2048                          & 2048                          \\
          & Num epochs        & 300                           & 300                           \\
\bottomrule
\end{tabular}
}
\end{table*}

\FloatBarrier
\subsection{Choice of target layers}

To accompany \iftoggle{useappendix}{\cref{s:ablation:target}}{Sec.~6.1 of the main text}, we show additional results exploring which hidden layers to use as targets for hidden-self-distillation.

\subsubsection{Number of targets}
\label{s:target-choice}

As shown in \cref{tab:change-target}, we explore the effect of varying how many hidden layers are targeted for self-distillation, the type of intermediate representation used (block outputs, mid-block representations, residual terms), and whether to include the tokenizer output or raw image pixels.
For ViT-S, we find two peaks in performance: one when using 3--4 equispaced block outputs as targets, and another when targeting all residuals and outputs from every block.
For ViT-B and ViT-L, the equispaced block outputs consistently outperform the denser target configurations.
Overall, targeting the output of every fourth block provides best or near-best performance across all three architecture sizes.

\begin{table*}[tbh]
\centering
\caption{%
Selection of which hidden layer(s) to predict during pretraining.
Block sets are denoted \texttt{start:step:end}, \eg{} \texttt{1:4:12} = \{1, 4, 8, 12\} and \texttt{1:1:12} denotes all blocks.
Experiments trained for 300 ep. on IN-1k (using \textit{prototype} hparams) and evaluated with frozen probe.
We also show the number of distillation targets per mask token.
We find using the output of every fourth block (highlighted) provides best or near best performance across architecture sizes.
}
\label{tab:change-target}
\small
\adjustbox{max width=\textwidth}{
\begin{tabular}{llrrrccc}
\toprule
 & & \multicolumn{3}{c}{\# Distillation targets} & \multicolumn{3}{c}{IN-1k Probe} \\
\cmidrule(l){3-5} \cmidrule(l){6-8}
Arch & Targets & Blocks & Layers & Dim & Patch & CLS & X-Blk \\
\midrule
ViT-S & Block 12 output                          &  1 &  1 &   384 &        54.1  &        53.8  &        68.5  \\
      & Block 1:6:12 output                      &  3 &  3 &  1152 &  \best{62.9} & \sbest{62.8} &        72.5  \\
\rowcolor{vlightgray}\cellcolor{white}
      & Block 1:4:12 output                      &  4 &  4 &  1536 &        62.2  &        62.4  & \sbest{73.0} \\
      & Block 1:3:12 output                      &  5 &  5 &  1920 &        60.7  &        61.3  &        72.8  \\
      & Block 2:2:12 output                      &  6 &  6 &  2304 &        59.8  &        61.5  &        72.8  \\
      & Block 1:4:12 output and residuals        &  4 & 12 &  4608 &        57.6  &        54.3  &        70.2  \\
      & Block 1:1:12 output                      & 12 & 12 &  4608 &        58.5  &        56.3  &        72.2  \\
      & Block 1:1:12 residuals                   & 12 & 24 &  9216 &        55.6  &        54.8  &        70.0  \\
      & Block 1:1:12 mid and output              & 12 & 24 &  9216 &        59.3  &        59.9  &        72.8  \\
      & Block 1:1:12 residuals and output        & 12 & 36 & 13824 & \sbest{62.4} &        61.9  &  \best{73.1} \\
      & Block 1:1:12 residuals, mid, output      & 12 & 48 & 18432 &        61.7  &  \best{63.1} & \sbest{73.0} \\
      & Img; Tok; Block 1:1:12 res., mid, output & 2+12 & 50 & 19584 &        60.5  &        61.7  &        72.9  \\
\addlinespace
ViT-B & Block 1:6:12 output                      &  3 &  3 &  2304 & \sbest{72.3} & \sbest{73.4} &        78.2  \\
\rowcolor{vlightgray}\cellcolor{white}
      & Block 1:4:12 output                      &  4 &  4 &  3072 &  \best{73.4} &  \best{74.4} &  \best{78.7} \\
      & Block 1:3:12 output                      &  5 &  5 &  3840 &        72.1  &        71.7  & \sbest{78.3} \\
      & Block 1:2:12 output                      &  6 &  6 &  4608 &        70.2  &        70.8  &        78.0  \\
      & Block 1:1:12 output                      & 12 & 12 &  9216 &        70.4  &        68.2  &        77.7  \\
      & Block 1:1:12 residuals and output        & 12 & 36 & 27648 &        70.3  &        70.7  &        77.8  \\
\addlinespace
ViT-L & Block 1:8:24 output                      &  4 &  4 &  4096 &  \best{77.9} & \sbest{77.4} & \sbest{81.2} \\
\rowcolor{vlightgray}\cellcolor{white}
      & Block 1:4:24 output                      &  7 &  7 &  7168 & \sbest{77.9} &  \best{78.0} &  \best{81.2} \\
      & Block 1:1:24 residuals and output        & 24 & 72 & 73728 &        77.1  &        76.1  &        80.8  \\
\bottomrule
\end{tabular}
}
\end{table*}

\subsubsection{Whether to use the tokenizer block}
We considered whether to include the tokenizer, the least processed layer after the raw image pixels, as a self-distillation target.
This layer's output has an advantage over the raw image pixels in being the same shape as the transformer block outputs, which (though not strictly necessary) does make processing easier.
As shown in \cref{tab:target-tokenizer}, we found using the tokenizer's output instead of the first block when using spaced out targets was detrimental.
For target configurations where all transformer blocks were being predicted, including the tokenizer layer as a self-distillation target was generally negative on IN-1k performance.

\begin{table}[tbh]
\centering
\caption{Choosing whether to predict the tokenizer's output.
We consider whether to predict the tokenizer output (Tok.) instead of the first block for spaced targets, or in addition to all blocks.
We find it is generally disadvantageous to use the tokenizer as a self-distillation target in these scenarios.
ViT-S, trained on IN-1k, 300 ep., \textit{prototype} hparams.
}
\label{tab:target-tokenizer}
\small
\begin{tabular}{lrccc}
\toprule
 & \multicolumn{1}{c}{\# tgt} & \multicolumn{3}{c}{IN-1k Probe} \\
\cmidrule(r){2-2} \cmidrule(r){3-5}
Distillation targets & Layer & Patch & CLS & X-Blk \\
\midrule
\rowcolor{vlightgray}
Block 1:4:12 out                    &  4 & \sbest{62.2} &  \best{62.4} & \sbest{73.0} \\
Block 4:4:12 out \& Tok.            &  4 &        60.3  &        61.6  &        72.4  \\
\addlinespace
Block 1:1:12 out                    & 12 &        58.5  &        56.3  &        72.2  \\
Block 1:1:12 out \& Tok.            & 13 &        59.2  &        59.2  &        72.4  \\
\addlinespace
Block 1:1:12 mid, out               & 24 &        59.3  &        59.9  &        72.8  \\
Block 1:1:12 mid, out \& Tok.       & 25 &        58.9  &        58.9  &        72.3  \\
\addlinespace
Block 1:1:12 res., out              & 36 &  \best{62.4} & \sbest{61.9} &  \best{73.1} \\
Block 1:1:12 res., out \& Tok.      & 37 &        60.4  &        58.7  &        72.5  \\
\bottomrule
\end{tabular}
\end{table}

\subsection{Masking hyperparameter sensitivity}
\label{s:ablation-masking-hparams}

We ablate the masking hyperparameters of Bootleg: the size of each mask region (\cref{tab:mask-size}), the number of mask regions (\cref{tab:mask-count}), and the aspect ratio range of mask regions (\cref{tab:mask-aspect}).

For each ablation, we report the average fraction of patch tokens which were seen by the encoder, and the fraction of patch tokens used as a target by at least one mask.
We also show the fraction of targets which were adjacent to a seen token.

\subsubsection{Mask size}

\cref{tab:mask-size} shows the effect of scaling mask region size while keeping $K{=}4$ regions fixed.
The default mask scale (100\%) yields the best results across all metrics.
The performance falls off similarly as the mask size is increased or decreased.
Smaller masks (80\%) leave too much context visible (40.3\% seen), reducing the difficulty of the prediction task, while larger masks (110--140\%) occlude too much of the image, limiting the encoder's access to informative context.
Performance holds out well across the range of mask sizes explored, despite the mean seen rate ranging from 22\% to 40\%.

\begin{table}[tbh]
\centering
\caption{Effect of mask region size. The mask scale is varied as a percentage relative to the default (highlighted), keeping the number of mask regions fixed at $K{=}4$. ViT-S, 300 ep. on IN-1k, \textit{final} hparams.}
\label{tab:mask-size}
\small
\adjustbox{max width=\textwidth}{%
\begin{tabular}{llccccccc}
\toprule
 & & & & & \multicolumn{3}{c}{IN-1k acc.} & \multicolumn{1}{c}{ADE20K mIoU} \\
\cmidrule(r){6-8} \cmidrule{9-9}
$K$ & Mask scale & Seen (\%) & Target (\%) & Adj. (\%) & Patch & CLS & X-Blk & kNN \\
\midrule
4 & $\phantom{0}70\% \rightarrow [0.112,\, 0.128]$ & $46.5 \pm 7.7$ & $36.9 \pm 2.6$ & $39.5 \pm 5.5$ & 62.4 & 62.0 & 72.4 & 18.9 \\
4 & $\phantom{0}80\% \rightarrow [0.128,\, 0.146]$ & $40.4 \pm 7.7$ & $40.6 \pm 2.7$ & $33.4 \pm 5.8$ & 65.3 & 64.9 & 73.2 & 19.7 \\
4 & $\phantom{0}90\% \rightarrow [0.144,\, 0.165]$ & $33.9 \pm 7.1$ & $44.5 \pm 2.5$ & $27.0 \pm 5.2$ & \sbest{67.2} & \sbest{68.3} & \sbest{74.2} & 21.1 \\
\rowcolor{vlightgray}
4 & $100\% \rightarrow [0.160,\, 0.183]$ & $29.3 \pm 6.7$ & $47.3 \pm 2.3$ & $22.6 \pm 4.9$ & \best{67.8} & \best{68.9} & \best{74.4} & \best{22.0} \\
4 & $110\% \rightarrow [0.176,\, 0.201]$ & $25.5 \pm 6.4$ & $49.7 \pm 2.1$ & $19.1 \pm 4.6$ & 66.7 & 67.7 & 73.6 & \sbest{21.7} \\
4 & $120\% \rightarrow [0.192,\, 0.220]$ & $21.8 \pm 6.3$ & $52.3 \pm 2.3$ & $15.8 \pm 4.6$ & 65.1 & 66.2 & 73.1 & 21.2 \\
4 & $140\% \rightarrow [0.224,\, 0.256]$ & $15.9 \pm 5.3$ & $56.9 \pm 2.0$ & $11.0 \pm 3.5$ & 62.4 & 62.9 & 71.4 & 20.2 \\
\bottomrule
\end{tabular}
}
\end{table}

\subsubsection{Number of masks}

\cref{tab:mask-count} varies the number of mask regions, $K$, while adjusting their size to keep the total masked area approximately constant.
The default $K{=}4$ achieves the best results, with $K{=}3$ and $K{=}5$ close behind.
Fewer, larger masks ($K{=}1$, $K{=}2$) produce contiguous masked regions that occlude a large area, elevating the task.
Many small masks ($K{=}8$) scatter targets across the image, reducing task difficulty slightly due to the reduced size of each mask.
The results suggest that a moderate number of mask regions provides the best balance between prediction difficulty and spatial diversity.
Performance holds out well across the range of number of mask regions.

\begin{table}[tbh]
\centering
\caption{Effect of changing the number of mask regions, $K$. The mask scale is adjusted to keep the seen fraction of the image approximately constant. The mask scale percentage is relative to the default ($K{=}4$). ViT-S, 300 ep. on IN-1k, \textit{final} hparams.}
\label{tab:mask-count}
\small
\adjustbox{max width=\textwidth}{%
\begin{tabular}{llccccccc}
\toprule
 & & & & & \multicolumn{3}{c}{IN-1k acc.} & \multicolumn{1}{c}{ADE20K mIoU} \\
\cmidrule(r){6-8} \cmidrule{9-9}
$K$ & Mask scale & Seen (\%) & Target (\%) & Adj. (\%) & Patch & CLS & X-Blk & kNN \\
\midrule
1 & $379\% \rightarrow [0.575,\, 0.726]$ & $28.4 \pm 8.0$ & $64.8 \pm 5.0$ & $19.5 \pm 7.1$ & 67.2 & 67.5 & 73.5 & 21.2 \\
2 & $208\% \rightarrow [0.281,\, 0.432]$ & $29.1 \pm 7.7$ & $52.0 \pm 8.0$ & $16.1 \pm 7.0$ & 67.1 & 67.6 & 73.6 & 21.4 \\
3 & $131\% \rightarrow [0.196,\, 0.254]$ & $28.7 \pm 7.2$ & $47.3 \pm 7.8$ & $16.0 \pm 8.0$ & \sbest{67.5} & \sbest{68.6} & 74.0 & 21.7 \\
\rowcolor{vlightgray}
4 & $100\% \rightarrow [0.160,\, 0.183]$ & $29.3 \pm 6.7$ & $47.3 \pm 7.3$ & $17.7 \pm 8.3$ & \best{67.8} & \best{68.9} & \best{74.4} & \best{22.0} \\
5 & $\phantom{0}86\% \rightarrow [0.126,\, 0.168]$ & $28.6 \pm 7.4$ & $48.9 \pm 7.6$ & $18.2 \pm 9.0$ & 67.3 & 68.1 & \sbest{74.1} & \sbest{21.8} \\
6 & $\phantom{0}72\% \rightarrow [0.109,\, 0.138]$ & $30.0 \pm 6.9$ & $48.9 \pm 7.1$ & $20.8 \pm 9.1$ & 66.9 & 67.7 & 73.8 & 21.1 \\
8 & $\phantom{0}57\% \rightarrow [0.083,\, 0.111]$ & $29.2 \pm 7.0$ & $50.8 \pm 6.9$ & $22.1 \pm 9.6$ & 65.5 & 66.1 & 73.5 & 20.4 \\
\bottomrule
\end{tabular}
}
\end{table}

\subsubsection{Mask aspect ratio}

\cref{tab:mask-aspect} compares different aspect ratio ranges for mask regions.
The default range $[0.667, 1.5]$ achieves the best performance across all metrics.
There is only a small impact on the performance of the model when changing the mask aspect ratio.

Using square masks slightly increases the difficulty in the task because more targets are further from seen components of the image.
Similarly, wider aspect ratio ranges ($[0.5, 2]$ and $[0.333, 3]$) increases the task difficulty slightly as highly elongated mask rectangles allow the model to see context closer to the target patches.
The default aspect ratio provides the best balance in task complexity, though the effect is not large.

\begin{table}[tbh]
\centering
\caption{Effect of mask aspect ratio range. The default aspect ratio (highlighted) allows moderate variation; $[1,\,1]$ constrains masks to be square. ViT-S, $K{=}4$, 300 ep. on IN-1k, \textit{final} hparams. Although the mask scale parameter is held constant, the seen rate varies due to changes in the distribution of possible mask areas for a given aspect ratio range.}
\label{tab:mask-aspect}
\small
\adjustbox{max width=\textwidth}{%
\begin{tabular}{cccccccc}
\toprule
 & & & & \multicolumn{3}{c}{IN-1k acc.} & \multicolumn{1}{c}{ADE20K mIoU} \\
\cmidrule(r){5-7} \cmidrule{8-8}
Aspect ratio & Seen (\%) & Target (\%) & Adj. (\%) & Patch & CLS & X-Blk & kNN \\
\midrule
$[1,\, 1]$ & $25.1 \pm 5.6$ & $49.4 \pm 7.5$ & $14.0 \pm 7.0$ & \sbest{67.2} & 68.0 & 73.6 & 21.5 \\
\rowcolor{vlightgray}
$[0.667,\, 1.5]$ & $29.3 \pm 6.7$ & $47.3 \pm 7.3$ & $17.7 \pm 8.3$ & \best{67.8} & \best{68.9} & \best{74.4} & \best{22.0} \\
$[0.5,\, 2]$ & $30.1 \pm 6.9$ & $47.2 \pm 7.2$ & $18.8 \pm 8.8$ & 67.2 & \sbest{68.3} & \sbest{74.2} & \sbest{21.6} \\
$[0.333,\, 3]$ & $30.4 \pm 7.0$ & $47.8 \pm 7.2$ & $19.8 \pm 9.4$ & 67.1 & 67.7 & 73.9 & 21.3 \\
\bottomrule
\end{tabular}
}
\end{table}

\subsection{Removing EMA}
\label{s:no-ema}

We explored the effect on Bootleg training of removing the EMA teacher and performing hidden-self-distillation on the student directly.
For these experiments, the teacher-encoder is set to be a clone of the student on every step (EMA momentum 0) and the method is otherwise unchanged.
Thus the ``teacher'' forward pass sees the whole image, whereas the ``student'' forward pass sees only the visible context.
We retain a stop-gradient on the outputs of the ``teacher'' forward pass, so the student encoder and predictor are updated to better predict the embeddings of hidden layers at the masked positions, but those embeddings are not themselves updated to be better predicted.

As shown in \cref{tab:no-ema}, removing the EMA teacher causes training to collapse for I-JEPA, for Bootleg with its four distillation targets, and for a Bootleg variant distilling only the final block, with IN-1k accuracy falling by 45--62 percentage points in each case.
Hence, while multiple distillation targets stabilize training relative to a single deep target (\cref{s:training-dynamics:collapse}), they do not remove the need for the EMA teacher: an EMA (or another collapse-prevention mechanism) remains necessary for latent self-distillation.

\begin{table}[tbh]
\centering
\caption{Performance when ablating the EMA teacher: self-distillation is applied directly to (a stop-gradient copy of) the student encoder.
ViT-S/16, trained on IN-1k for 300 ep., evaluated with frozen probes.
I-JEPA uses its published hyperparameters, and Bootleg uses ours (\cref{s:pretraining-hparams}), with its four hidden blocks or only the final block (b12) as targets.
The I-JEPA rows come from a different pretraining run, evaluated with an earlier version of our probe protocol, than the I-JEPA baseline in \cref{tab:add-targets-to-others,tab:config-interpolation} (50.3 vs.\ 53.4 Patch); compare them with each other only.
}
\label{tab:no-ema}
\small
\begin{tabular}{llr@{\hspace{0.3em}}rr@{\hspace{0.3em}}rr@{\hspace{0.3em}}r}
\toprule
        &          & \multicolumn{6}{c}{IN-1k Probe} \\
\cmidrule(r){3-8}
Method  & Ablation & \multicolumn{2}{l}{Patch} & \multicolumn{2}{l}{CLS} & \multicolumn{2}{l}{X-Blk} \\
\midrule
I-JEPA                    & Baseline & 50.3 &               &  \na &               & 62.1 &               \\
                          & No EMA   &  4.6 & \adecr{-45.7} &  \na &               & 14.0 & \adecr{-48.1} \\
Bootleg (b1, b4, b8, b12) & Baseline & 67.8 &               & 68.9 &               & 74.4 &               \\
                          & No EMA   &  8.5 & \adecr{-59.3} &  6.8 & \adecr{-62.1} & 24.2 & \adecr{-50.2} \\
Bootleg (b12 only)        & Baseline & 56.7 &               & 54.6 &               & 68.6 &               \\
                          & No EMA   &  5.0 & \adecr{-51.7} &  3.9 & \adecr{-50.7} & 17.8 & \adecr{-50.7} \\
\bottomrule
\end{tabular}
\end{table}

\subsection{Number of pretraining epochs}
\label{sec:abl:epochs}

We compare Bootleg (multi-target, blocks $\{1, 4, 8, 12\}$) against a block-12-only ablation when sweeping the number of pretraining epochs from 50 to 1200.
The block-12-only ablation removes Bootleg's hidden self-distillation, distilling only from the penultimate teacher block.
Both variants use the final hyperparameter configuration (\cref{tab:hparams-prelim}) at ViT-S/16 with the IN-1k probe and ADE20K kNN evaluation.
\Cref{tab:abl-epochs} reports the results, and \cref{fig:abl-epochs-xblk} plots the IN-1k X-Blk top-1 accuracy as a function of training duration.

The two variants behave very differently as training duration grows.
Bootleg's accuracy continues to improve through 600 epochs and then plateaus, with X-Blk top-1 reaching 75.3\% at 600 epochs.
The block-12-only ablation, by contrast, peaks early (around 100 epochs) and then \emph{degrades} with longer training, dropping by more than 9 points on X-Blk by 1200 epochs.
This collapse-with-duration is consistent with the well-known instability of single-final-layer self-distillation, and demonstrates that hidden self-distillation provides a clear stability benefit at long training horizons.

\begin{table}[tbh]
\centering
\caption{Effect of pretraining duration.
ViT-S/16, IN-1k, \textit{final} hyperparameters.
}
\label{tab:abl-epochs}
\small
\begin{tabular}{llcccc}
\toprule
 & & \multicolumn{3}{c}{IN-1k Probe} & ADE20K \\
\cmidrule(r){3-5} \cmidrule(r){6-6}
Epochs & Targets & Patch & CLS & X-Blk & kNN \\
\midrule
50 & b1, 4, 8, 12         & 52.9 & 53.0 & 66.3 & 14.3 \\
   & b12 & 55.0 & 53.8 & 67.0 & 15.9 \\
\addlinespace
100 & b1, 4, 8, 12         & 59.0 & 59.4 & 70.0 & 17.7 \\
    & b12 & \best{61.0} & \best{60.2} & \best{69.9} & \best{17.1} \\
\addlinespace
200 & b1, 4, 8, 12         & 65.5 & 66.3 & 73.2 & 20.9 \\
    & b12 & \sbest{59.3} & \sbest{57.1} & \sbest{69.6} & \sbest{16.2} \\
\addlinespace
300 & b1, 4, 8, 12         & 67.8 & 68.9 & 74.4 & 22.0 \\
    & b12 & 56.7 & 54.6 & 68.6 & 13.1 \\
\addlinespace
600 & b1, 4, 8, 12         & \best{69.8} & \best{70.4} & \best{75.3} & \best{23.5} \\
    & b12 & 47.9 & 46.4 & 64.4 & \phantom{0}7.5 \\
\addlinespace
900 & b1, 4, 8, 12         & 69.0 & 69.9 & 75.0 & 22.4 \\
    & b12 & 46.8 & 47.7 & 63.5 & \phantom{0}6.2 \\
\addlinespace
1200 & b1, 4, 8, 12         & \sbest{69.4} & \sbest{70.2} & \sbest{75.0} & \sbest{22.5} \\
     & b12 & 43.8 & 41.2 & 60.0 & \phantom{0}5.5 \\
\bottomrule
\end{tabular}
\end{table}

\begin{figure}[tbh]
\centering
\includegraphics[width=\iftoggle{arxiv}{0.6}{0.7}\linewidth]{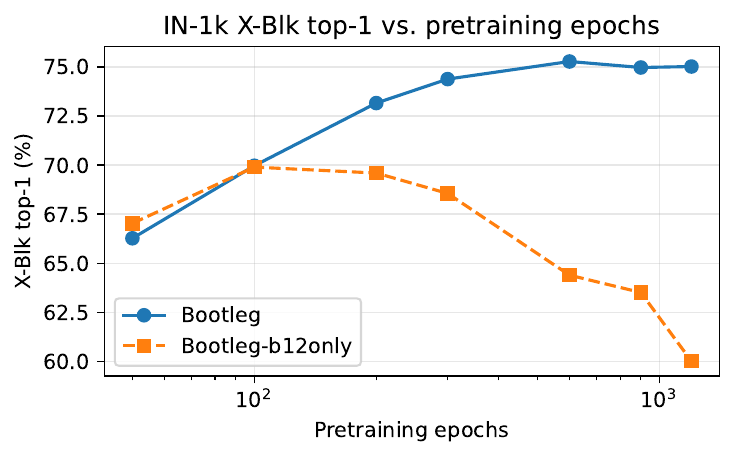}
\caption{IN-1k X-Blk top-1 accuracy as a function of pretraining duration for Bootleg (multi-target, blocks $\{1, 4, 8, 12\}$) and the block-12-only ablation.
ViT-S/16, IN-1k, \textit{final} hyperparameters.
Each datapoint is a separate run, trained from scratch to the budgeted number of epochs.
Bootleg continues to improve through 600 epochs and plateaus; the block-12-only variant peaks near 100 epochs and then degrades with longer training.
}
\label{fig:abl-epochs-xblk}
\end{figure}

\subsection{Teacher EMA momentum sensitivity}
\label{sec:abl:ema}

We sweep the teacher EMA momentum $\tau$ around its default value of $0.9985$ and compare Bootleg against the block-12-only ablation.
\Cref{tab:abl-ema} reports the results.
Bootleg is largely insensitive to $\tau$ over the swept range: probe accuracies vary by less than 1.3 points across $\tau \in [0.988, 0.9996]$.
The block-12-only ablation, in contrast, is highly sensitive: shifting $\tau$ from $0.988$ to $0.9996$ changes Patch probe accuracy by 20 points (42.1\% $\to$ 62.2\%) and ADE20K kNN by 15 points (1.8\% $\to$ 17.2\%), with higher $\tau$ (slower teacher) clearly preferred.
This is again consistent with the block-12-only variant relying more critically on a stable teacher to avoid representational drift, while Bootleg's multi-layer distillation provides robustness to teacher dynamics.

\begin{table}[tbh]
\centering
\caption{Teacher EMA momentum $\tau$ sensitivity.
ViT-S/16, 300 ep.\ on IN-1k, \textit{final} hyperparameters.
The default $\tau = 0.9985$ is highlighted.
}
\label{tab:abl-ema}
\small
\begin{tabular}{llcccc}
\toprule
 & & \multicolumn{3}{c}{IN-1k Probe} & ADE20K \\
\cmidrule(r){3-5} \cmidrule(r){6-6}
$\tau$ & Targets & Patch & CLS & X-Blk & kNN \\
\midrule
$0.988$ & b1, 4, 8, 12         & 66.6 & 67.1 & 73.6 & 21.2 \\
        & b12 & 42.1 & 41.3 & 60.6 & \phantom{0}1.8 \\
\addlinespace
$0.997$ & b1, 4, 8, 12         & \sbest{67.5} & \sbest{68.3} & \sbest{74.1} & \sbest{21.5} \\
        & b12 & 51.5 & 49.3 & 66.5 & \phantom{0}9.7 \\
\addlinespace
\rowcolor{vlightgray}%
$0.9985$ & b1, 4, 8, 12         & \best{67.8} & \best{68.9} & \best{74.4} & \best{22.0} \\
         & b12 & \sbest{56.7} & \sbest{54.6} & \sbest{68.6} & \sbest{13.1} \\
\addlinespace
$0.9996$ & b1, 4, 8, 12         & 66.6 & 68.0 & 73.9 & \sbest{21.5} \\
         & b12 & \best{62.2} & \best{61.8} & \best{71.0} & \best{17.2} \\
\bottomrule
\end{tabular}
\end{table}

\subsection{Optimizer $\beta_2$ sensitivity}
\label{sec:abl:beta2}

We sweep the AdamW second-moment decay $\beta_2$ around its default value of $0.95$.
\Cref{tab:abl-beta2} shows that both Bootleg and the block-12-only ablation are relatively insensitive to $\beta_2$ over $\{0.9, 0.95, 0.999\}$, with Bootleg varying by less than 1 point on all probe heads.
The block-12-only ablation prefers smaller $\beta_2$ (here $0.9$ wins on every head), while Bootleg is best at the default $0.95$ on the probe heads and at $0.999$ on ADE20K kNN.
We retain the default $0.95$.

\begin{table}[tbh]
\centering
\caption{Optimizer $\beta_2$ sensitivity.
ViT-S/16, 300 ep.\ on IN-1k, final hyperparameters.
The default $\beta_2 = 0.95$ is highlighted.
}
\label{tab:abl-beta2}
\small
\begin{tabular}{llcccc}
\toprule
 & & \multicolumn{3}{c}{IN-1k Probe} & ADE20K \\
\cmidrule(r){3-5} \cmidrule(r){6-6}
$\beta_2$ & Targets & Patch & CLS & X-Blk & kNN \\
\midrule
$0.9$ & b1, 4, 8, 12         & 67.1 & 68.2 & \sbest{74.0} & 21.3 \\
      & b12 & \best{57.6} & \best{54.7} & \best{68.9} & \best{13.2} \\
\addlinespace
\rowcolor{vlightgray}\cellcolor{white}%
$0.95$ & b1, 4, 8, 12         & \best{67.8} & \best{68.9} & \best{74.4} & \sbest{22.0} \\
       & b12 & \sbest{56.7} & \sbest{54.6} & \sbest{68.6} & \sbest{13.1} \\
\addlinespace
$0.999$ & b1, 4, 8, 12         & \sbest{67.2} & \sbest{68.5} & 74.0 & \best{22.2} \\
        & b12 & 53.1 & 50.6 & 66.9 & 11.5 \\
\bottomrule
\end{tabular}
\end{table}

\FloatBarrier
\section{Additional experiments}

In this section, we present additional experiments exploring design choices in the Bootleg framework.
Unless otherwise noted, experiments use ViT-S pretrained on IN-1k for 300 epochs with the final hyperparameter configuration (\cref{tab:hparams-prelim}).

\subsection{Predicting register tokens}

In Bootleg, the embedding of the CLS token is passed from the encoder to the predictor, along with embeddings of the visible patches.
However, the register tokens are not passed from the encoder to predictor.
Instead, the predictor has its own register tokens.

Since the register tokens are not visible to the predictor, estimating these would be non-trivial for the predictor.
We explored whether including the hidden embeddings of the register tokens as targets (in addition to the hidden-layer embeddings of the masked-out tokens) would improve the performance of the network.
As shown in \cref{tab:target-reg}, we found including register tokens as targets has negligible impact in training performance.

\begin{table}[tbh]
\centering
\caption{Effect of using hidden embeddings of register tokens as self-distillation targets, in addition to hidden embeddings of patch tokens.
ViT-S, trained 300ep on IN-1k, \textit{prototype} hparams.
}
\label{tab:target-reg}
\small
\begin{tabular}{llcccc}
\toprule
        &  & \multicolumn{4}{c}{IN-1k Probe} \\
\cmidrule(r){3-6}
Arch    & Distil. method & Patch & CLS & X-Attn & X-Blk \\
\midrule
\rowcolor{vlightgray}\cellcolor{white}
ViT-S & Baseline         &  \best{62.2} & \sbest{62.4} &  \best{71.1} &  \best{73.0} \\
      & +target 4 reg.   & \sbest{62.2} &  \best{62.5} & \sbest{70.9} & \sbest{72.7} \\
\bottomrule
\end{tabular}
\end{table}

\begin{table*}[tbh]
\centering
\caption{Effect of splitting targets across mask rectangles.
We show the number of target layers per mask rectangle (/) and over all masks ($\Sigma$).
ViT-S, trained 300ep on IN-1k, \textit{prototype} hparams.
}
\label{tab:target-groups}
\small
\adjustbox{max width=\textwidth}{
\begin{tabular}{llrrcccc}
\toprule
 & & \multicolumn{2}{c}{\# targets} & \multicolumn{4}{c}{IN-1k Probe} \\
\cmidrule(r){3-4} \cmidrule(r){5-8}
Grouping method & Distillation targets per mask rectangle & / & $\Sigma$ & Patch & CLS & X-Attn & X-Blk \\
\midrule
\rowcolor{vlightgray}
Same (all) for each rectangle     & Blocks $\{1, 4, 8, 12\}$ outputs          & 4 & 4 & \sbest{62.2} &  \best{62.4} & \sbest{71.1} & \sbest{73.0} \\
Split as single target per rect.  & Block $g$ out; for $g = 1, 4, 8, 12$     & 1 & 4 &        60.7  &        61.7  &        70.8  &        72.7  \\
\addlinespace
Same (all) for each rectangle     & Blocks 1:1:12 res. \& out              & 36 & 36 &  \best{62.4} & \sbest{61.9} &  \best{71.4} &  \best{73.1} \\
Split as contiguous blocks        & Blocks $\{g, g+1, g+2\}$ res. \& out; for $g=1,4,7,10$   & 12 & 36 &        60.2  &        60.3  &        70.9  &        72.7  \\
Split as interleaved blocks       & Blocks $\{g, g+4, g+8\}$ res. \& out; for $g=1,2,3,4$   & 12 & 36 &        60.7  &        57.9  &        70.5  &        72.6  \\
Split as interleaved targets      & $\{r^\text{Attn}_l, r^\text{MLP}_{l+1}, x_{l+2}\}$ for $l\!\in\!\{1, 5, 9\}$, \etc{} & 12 & 36 &        60.9  &        60.6  &        70.9  &        72.8  \\
\bottomrule
\end{tabular}
}
\end{table*}

\subsection{Using different targets for each mask}

To train ViT-S and ViT-B, our Bootleg method uses four self-distillation targets (the outputs of the teacher's blocks 1, 4, 8, and 12), and predicts these targets for all masked out tokens within four blocking mask rectangles.
Each rectangle is processed separately with self-attention between the encoder outputs and the mask tokens.

This symmetry presents an opportunity: what if we break up the distillation targets and ask the predictor to predict only a quarter of the hidden layers for each pass through the predictor?
In the standard Bootleg config, this means each mask rectangle can devote its processing to predicting a single distillation target.
Intuitively, this may yield an advantage if the predictor is ``overloaded'' in its capacity to predict all the targets.

We implemented this variant of Bootleg by using a different learnable mask token for each of the four mask regions.
Since mask placement is symmetric, we let the first mask rectangle correspond to the first mask token and let its targets always be the output of block 1; similarly the second mask rectangle always predicts the output of block 4; \etc{}
We retain the full dimensionality of the read-out head (the final linear layer of the predictor network) as if the module were predicting all four targets, but select only the dimensions corresponding to the $i$-th target; we found this was necessary to best facilitate the training of the predictor.
As shown in \cref{tab:target-groups}, we found splitting up the target layers across the mask rectangles lead to a slight reduction in performance.

We also experimented with this method when using 36 targets: the output and residuals of each block.
As this is a much larger set of targets, it is more likely the predictor is ``overloaded'' when trying to complete all prediction tasks simultaneously.
For this set of targets, we explored multiple options of how to group the 36 targets into 4 sets of 12: (1) predict all 36 targets for every rectangle (default); (2) group blocks contiguously as $\{1, 2, 3\}$, $\{4, 5, 6\}$, $\{7, 8, 9\}$, $\{10, 11, 12\}$; (3) use interleaved blocks $\{1, 5, 9\}$, $\{2, 6, 10\}$, $\{3, 7, 11\}$, $\{4, 8, 12\}$; or (4) interleave targets entirely
$\{r^\text{Attn}_1, r^\text{MLP}_2, x_3, r^\text{Attn}_5, r^\text{MLP}_6, x_7, r^\text{Attn}_9, r^\text{MLP}_{10}, x_{11}\}$,
$\{r^\text{MLP}_1, x_2, r^\text{Attn}_4, r^\text{MLP}_5, x_6, r^\text{Attn}_8, r^\text{MLP}_9, x_{10}, r^\text{Attn}_{12}\}$,
\etc{}
In this case, we found (\cref{tab:target-groups}) using all targets for each of the 4 rectangles worked best, followed by interleaving the targets as much as possible.

These results indicate that using every hidden distillation as a target for every masked out token is the superior strategy.

\subsection{Components of V-JEPA~2.1}
\label{s:vjepa21}

V-JEPA~2.1 \citep{murlabadia2026vjepa21} shares Bootleg's use of multiple hidden layers as self-distillation targets (its ``deep self-supervision''), and adds two further components: a dense predictive loss on the \textit{context} tokens (weighted by distance to the masked region and ramped in early in training), and a multi-level predictor input that feeds the concatenated, normalized outputs of several encoder layers to the predictor.
It also uses a different masking strategy, in which the union of 10 blocks forms a single prediction target.
We evaluated each component added to Bootleg in isolation, as well as the full V-JEPA~2.1 objective trained with Bootleg's recipe (optimizer, schedule, EMA, augmentation, epochs; \cref{s:pretraining-hparams}).
The multi-level predictor input uses Bootleg's own target levels (b1, b4, b8 and the final output).

As shown in \cref{tab:vjepa21}, feeding the predictor multi-level inputs improves performance at image classification (most notably $+1.8$ Patch, a smaller $+0.7$ increase for CLS, and only $+0.2$ X-Blk), at the cost of a small reduction in semantic segmentation performance ($-0.5$ ADE20K mIoU).
Combining it with the dense context loss gives mixed results ($+0.8$ Patch, but $-0.2$ to $-1.0$ on the other metrics), and is worse than the multi-level input alone.
The dense context loss on its own reduces performance on all metrics, as does V-JEPA~2.1's masking strategy.
We note that, as in the reference implementation, V-JEPA~2.1's masks are truncated to the shortest mask in the per-GPU batch, so the effective context and target sizes depend on the per-GPU batch size; at our per-GPU batch of 256, the retained masks average ${\sim}25$ context and ${\sim}92$ target tokens of 196.
The full V-JEPA~2.1 objective at Bootleg's training recipe performs substantially worse than Bootleg, indicating that its gains are tied to its own (video) training recipe rather than transferring directly to IN-1k image pretraining at this scale.

\begin{table}[tbh]
\centering
\caption{%
Adding components of V-JEPA~2.1 \citep{murlabadia2026vjepa21} to Bootleg, each in isolation (deltas vs.\ the Bootleg baseline).
ViT-S/16, IN-1k, 300 ep., {final} hyperparameters. IN-1k top-1 accuracy (\%) with frozen probes, and ADE20K mIoU (\%) with a frozen linear decoder.
\na{}: no CLS token.
}
\label{tab:vjepa21}
\small
\adjustbox{max width=\textwidth}{%
\begin{tabular}{lr@{\hspace{0.3em}}rr@{\hspace{0.3em}}rr@{\hspace{0.3em}}rr@{\hspace{0.3em}}r}
\toprule
 & \multicolumn{6}{c}{IN-1k Probe} & \multicolumn{2}{c}{ADE20K} \\
\cmidrule(r){2-7} \cmidrule(l){8-9}
Model & \multicolumn{2}{l}{Patch} & \multicolumn{2}{l}{CLS} & \multicolumn{2}{l}{X-Blk} & \multicolumn{2}{l}{Lin} \\
\midrule
Bootleg (baseline) & 67.8 &  & 68.9 &  & 74.4 &  & 26.6 &  \\
\midrule
+ multi-level predictor input & 69.6 & \aincr{+1.8} & 69.5 & \aincr{+0.7} & 74.6 & \aincr{+0.2} & 26.1 & \adecr{-0.5} \\
+ dense context loss & 58.0 & \adecr{-9.8} & 61.9 & \adecr{-7.0} & 70.4 & \adecr{-4.0} & 20.8 & \adecr{-5.8} \\
+ both & 68.6 & \aincr{+0.8} & 68.7 & \adecr{-0.2} & 74.0 & \adecr{-0.4} & 25.6 & \adecr{-1.0} \\
+ V-JEPA~2.1 masking & 58.4 & \adecr{-9.4} & 58.2 & \adecr{-10.7} & 68.6 & \adecr{-5.7} & 20.9 & \adecr{-5.7} \\
V-JEPA~2.1 (at Bootleg recipe) & 54.8 & \adecr{-13.0} & \na{} &  & 64.3 & \adecr{-10.0} & 19.4 & \adecr{-7.2} \\
\bottomrule
\end{tabular}
}
\end{table}

\subsection{Using a cross-attention predictor with Bootleg}

To accompany the \textit{``Bootleg edition of CrossMAE''} in \iftoggle{useappendix}{\cref{s:ablation:bootleg-others}}{Sec.~6.2 of the main paper}, we considered whether Bootleg itself would perform better when using the cross-attention predictor of CrossMAE instead of a self-attention predictor network.
As shown in \cref{tab:xattn}, we found that using the cross-attention predictor on the output of the encoder only (XA without WFM) greatly reduced performance (62.4 $\to$ 54.7 patch probe).
Using cross-attention with the weighted feature map (XA with WFM) was overall slightly worse than using self-attention.
However, we did not increase the depth of the predictor network, as recommended by \citet{fu2025crossmae}; it is possible doing so would improve the results when training Bootleg with a cross-attention predictor.

\begin{table}[tbh]
\centering
\caption{Performance when using cross-attention predictor and weighted feature map (CrossMAE config) instead of self-attention (MAE config).
Performance when using either the MAE/I-JEPA method of self-attention (SA) for the predictor network, or the CrossMAE method of cross-attention (XA) against a weighted feature map (WFM).
The WFM uses a learnable weighted average of the outputs from the tokenizer and outputs of each transformer block in the encoder as the input to the predictor.
ViT-S, 300 ep., \textit{prototype} hparams.
}
\label{tab:xattn}
\small
\begin{tabular}{lllccc}
\toprule
 & & & \multicolumn{3}{c}{IN-1k Probe} \\
\cmidrule(r){4-6}
Arch & Distillation targets & Predictor & Patch & CLS & X-Blk \\
\midrule
\rowcolor{vlightgray}\cellcolor{white}
ViT-S %
 & Block 1:4:12 out       & SA      &        62.2  &  \best{62.4} & \sbest{73.0} \\
 & Block 1:4:12 out       & XA, WFM &        62.2  &        61.1  &        72.8  \\
 & Block 1:1:12 res., out & SA      & \sbest{62.4} & \sbest{61.9} &  \best{73.1} \\
 & Block 1:1:12 res., out & XA      &        54.7  &        46.5  &        69.7  \\
 & Block 1:1:12 res., out & XA, WFM &  \best{62.6} &        59.1  &        72.6  \\
\addlinespace
\rowcolor{vlightgray}\cellcolor{white}
ViT-B %
 & Block 1:4:12 out       & SA      &  \best{73.4} &  \best{74.4} &  \best{78.7} \\
 & Block 1:4:12 out       & XA, WFM & \sbest{72.9} & \sbest{72.6} & \sbest{78.7} \\
\bottomrule
\end{tabular}
\end{table}

\subsection{Jointly standardizing targets}

In Bootleg, our targets are standardized on a per-patch, per-hidden-layer basis by subtracting the mean and dividing by the standard deviation as measured over the embedding dimension (taking the z-score over the embedding dimension).
This standardization process is the same as in I-JEPA (which only targets one layer, the output of the teacher-encoder), and originates in MAE \citep{he2022masked}, in which analysis found using the z-score of the RGB values, referred to as the ``normalized pixel values'', was a better training target than the raw RGB values.

In self-distillation literature \citep{grill2020bootstrap, Chen2021simsiam, assran2023ijepa} taking the z-score of the targets---often as a z-score over the batch dimension, implemented by applying a BatchNorm module without an affine-transformation---is motivated by the need to prevent collapse of the representations.
This representation collapse can be seen to motivate our own target standardization, but in this case it would be sufficient to standardize the full target vector, rather than standardizing each embedding vector.

We investigated this by considering two options to handle the preparation of the target vector for a given mask patch token: (1) first separately standardize each embedding vector from blocks 1, 4, 8, 12; then concatenate the standardized embeddings to create the target vector; (2) first concatenate the embedding vectors from blocks 1, 4, 8, 12; then jointly standardize the concatenated embedding vector.
As shown in \cref{tab:jointly-standardize}, we find separate standardization outperforms jointly standardizing the embeddings, though joint standardization does not result in a collapse in the model.

\begin{table}[tbh]
\centering
\caption{%
We explored whether to standardize the targets with a z-score separately applied to each target embedding vector from each hidden layer used, or with a z-score taken across the concatenation of all targets.
ViT-S, trained on IN-1k, 300 ep. \textit{prototype} hparams.
}
\label{tab:jointly-standardize}
\small
\begin{tabular}{lcccc}
\toprule
 & \multicolumn{1}{c}{\# targets} & \multicolumn{3}{c}{IN-1k Probe} \\
\cmidrule(r){2-2} \cmidrule(r){3-5}
Target standardization  & Layers & Patch & CLS & X-Blk \\
\midrule
\rowcolor{vlightgray}
Separately standardized &  4 &  \best{62.2} &  \best{62.4} &  \best{73.0} \\
Jointly standardized    &  4 &  {57.8} & {57.2} & {69.8} \\
\bottomrule
\end{tabular}
\end{table}

\subsection{Target normalization}
\label{s:target-norm}

Bootleg standardizes each target token with a parameter-free layer-norm (z-normalization) before computing the MSE loss.
Motivated by recent work reporting benefits from averaging normalized representations \citep{singh2026raev2}, we compared this against L2 normalization of the targets, changing only the target normalization (\cref{tab:target-norm}).
Naive L2 normalization is clearly worse.
However, this is an optimization artefact rather than a property of the normalization: L2 normalization produces targets with unit norm, whereas z-normalization produces targets with norm ${\approx}\sqrt{d}$, so the MSE loss and the gradient reaching the predictor shrink by a factor of $d$ ($d=384$ for ViT-S; we measure a $433\times$ drop in the final loss).
Rescaling the L2-normalized targets by $\sqrt{d}$---which changes nothing about the direction or geometry of the targets---recovers the baseline performance.
The two schemes are therefore equivalent for our method once the target scale is matched.

\begin{table}[tbh]
\centering
\caption{Effect of the target normalization. ViT-S/16, IN-1k, 300 ep., preliminary experiments; IN-1k top-1 accuracy (\%) with frozen probes.
}
\label{tab:target-norm}
\small
\begin{tabular}{lccc}
\toprule
Target normalization & Target norm & Linear & Attentive \\
\midrule
z-normalization (ours) & ${\approx}\sqrt{d}$ & 65.4 & 73.8 \\
L2 normalization & 1 & 57.6 & 70.4 \\
L2 normalization, $\times\sqrt{d}$ & $\sqrt{d}$ & 65.4 & 73.6 \\
\bottomrule
\end{tabular}
\end{table}

\FloatBarrier
\section{Representational analysis}
\label{sec:rep-analysis}

To better understand the structure of Bootleg's learned representations, we analyze how embeddings relate across transformer blocks and spatial positions.
We compute two inter-layer similarity measures---element-wise Pearson correlation and Centered Kernel Alignment (CKA)~\cite{kornblith2019similarity}---as well as spatial autocorrelation within each layer.
All analyses are computed over 10{,}000 ImageNet-1k validation images using ViT-S/16 and ViT-B/16 Bootleg encoders trained for 600 epochs.

\subsection{Feature visualizations}
\label{s:feature-pca}

We visualize the patch features of frozen encoders following the protocol of CAPI \citep{darcet2025capi}.
For each image and layer, we fit a whitened PCA to the patch tokens (prefix CLS and register tokens are excluded), rescale the projections by the standard deviation of the first component, flip the sign of each component so that its skewness is positive, and map the first three components to RGB over the fixed range $[-2.3, 2.3]$.
This avoids per-image min--max scaling and arbitrary component signs, and involves no foreground masking.
We show the output of the blocks (before the final norm; every block for ViT-S, and block~1 and every second block for ViT-B/L) and the final output of the encoder (after its final norm, \ie{} the representation passed to the predictor/decoder, and the self-distillation target of I-JEPA).
Images are resized to $224{\times}224$, the resolution at which all models were trained; at higher resolutions, models whose position embeddings were trained for a $14{\times}14$ grid (including all of our own models) receive out-of-distribution position embeddings.
For an unselected set of images, we take 16 images evenly spaced through CAPI's default subset of the ADE20K training set \citep{ade20k} (every 19th image of the sorted training set, then every 64th of these), and add the demonstration images of DINO and DINOv2; the figures show a subset of eight.

\begin{figure}[tbh]
\centering
\includegraphics[width=\linewidth]{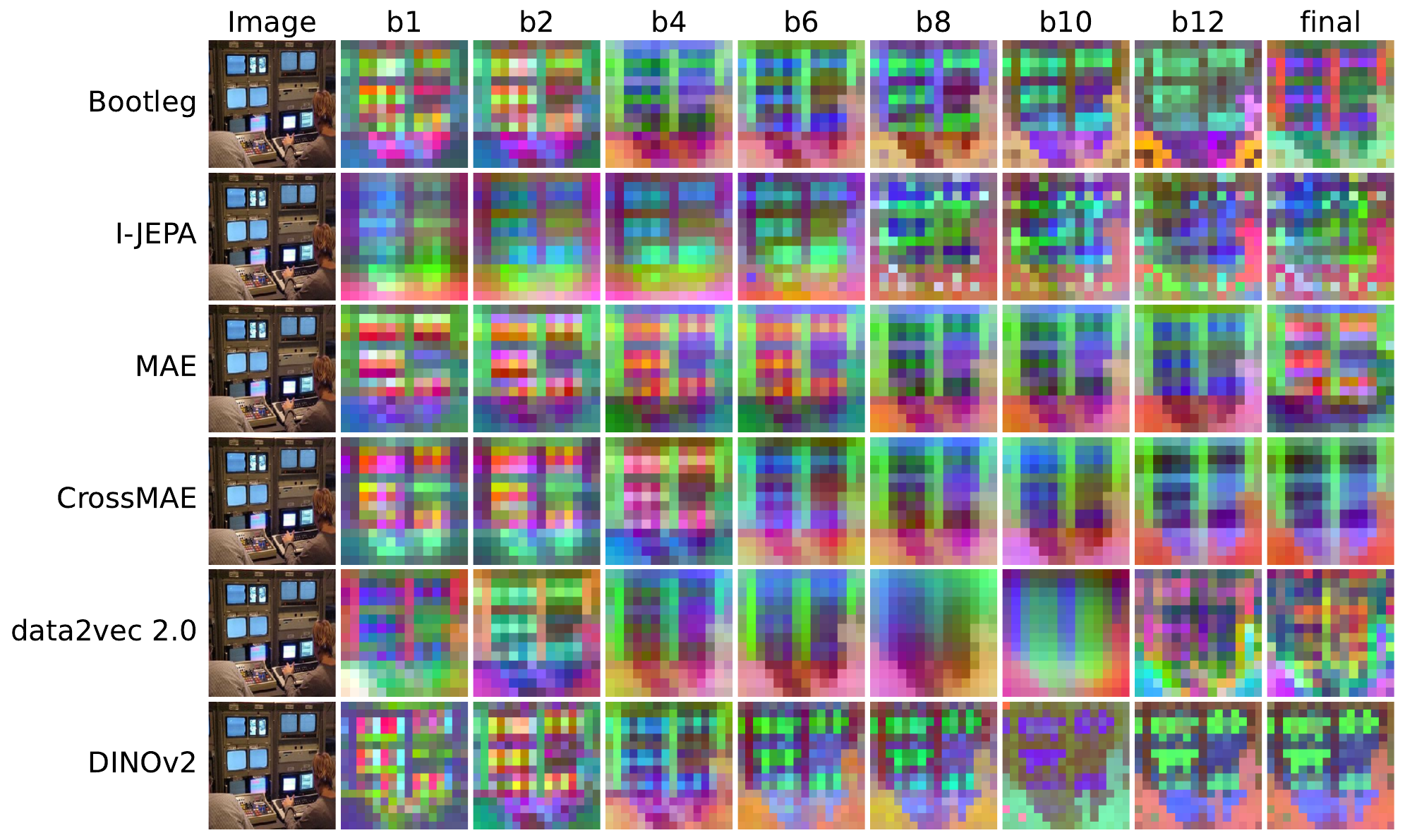}
\caption{PCA of the patch features at block~1, every second block, and the final output of ViT-B/16 encoders (B/14 for DINOv2), on an image from ADE20K, following the protocol of CAPI \citep{darcet2025capi} (per-image whitened PCA, first three components as RGB).}
\label{fig:pca-vitb-layers}
\end{figure}

\begin{figure}[tbh]
\centering
\includegraphics[width=\linewidth]{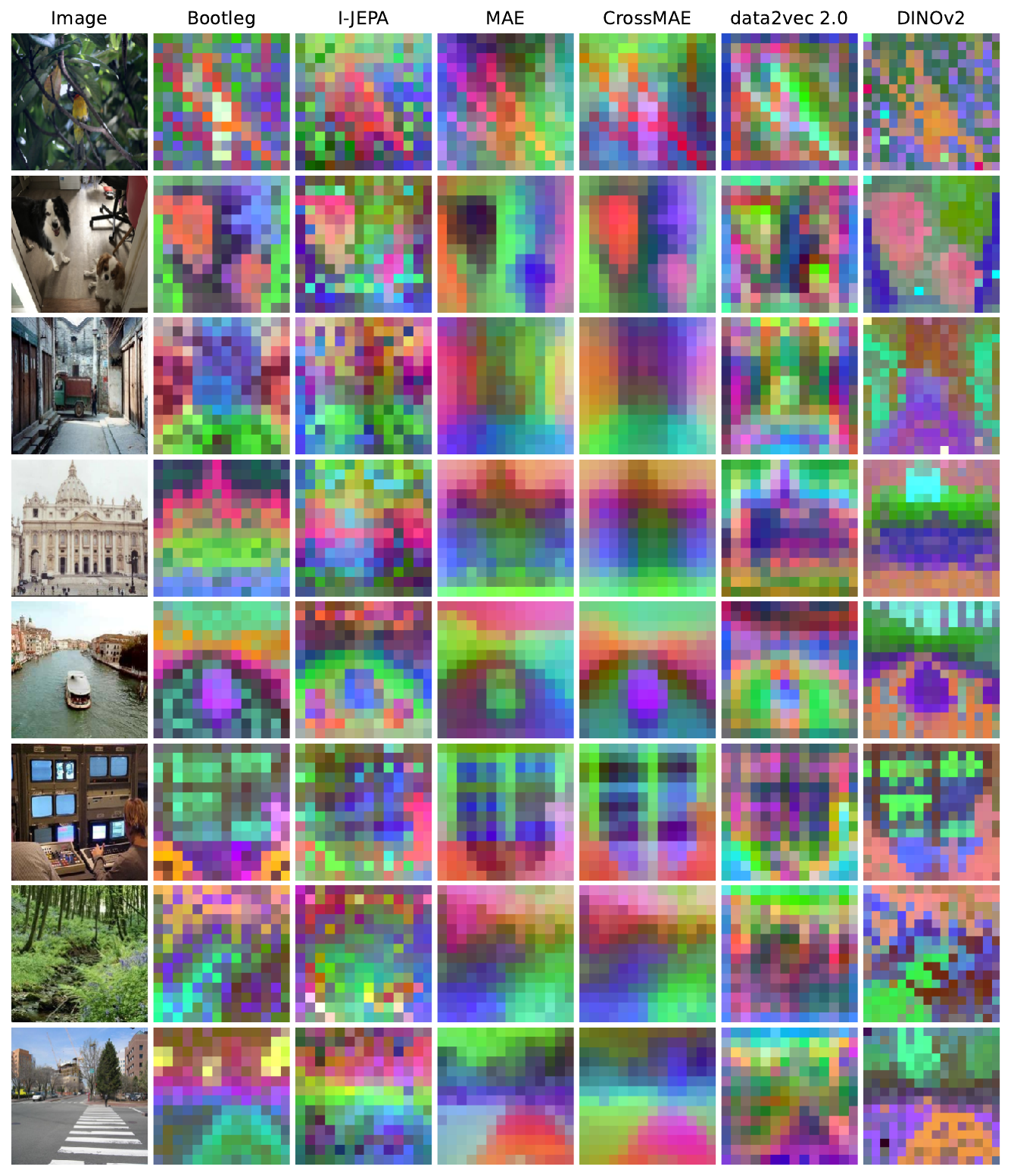}
\caption{PCA of the output of the last block (b12, before the final norm) of ViT-B encoders, for the demonstration images of DINO and DINOv2 (top two rows) and six images from ADE20K (per-image PCA, CAPI protocol).}
\label{fig:pca-vitb-final}
\end{figure}

\begin{figure}[tbh]
\centering
\includegraphics[width=\linewidth]{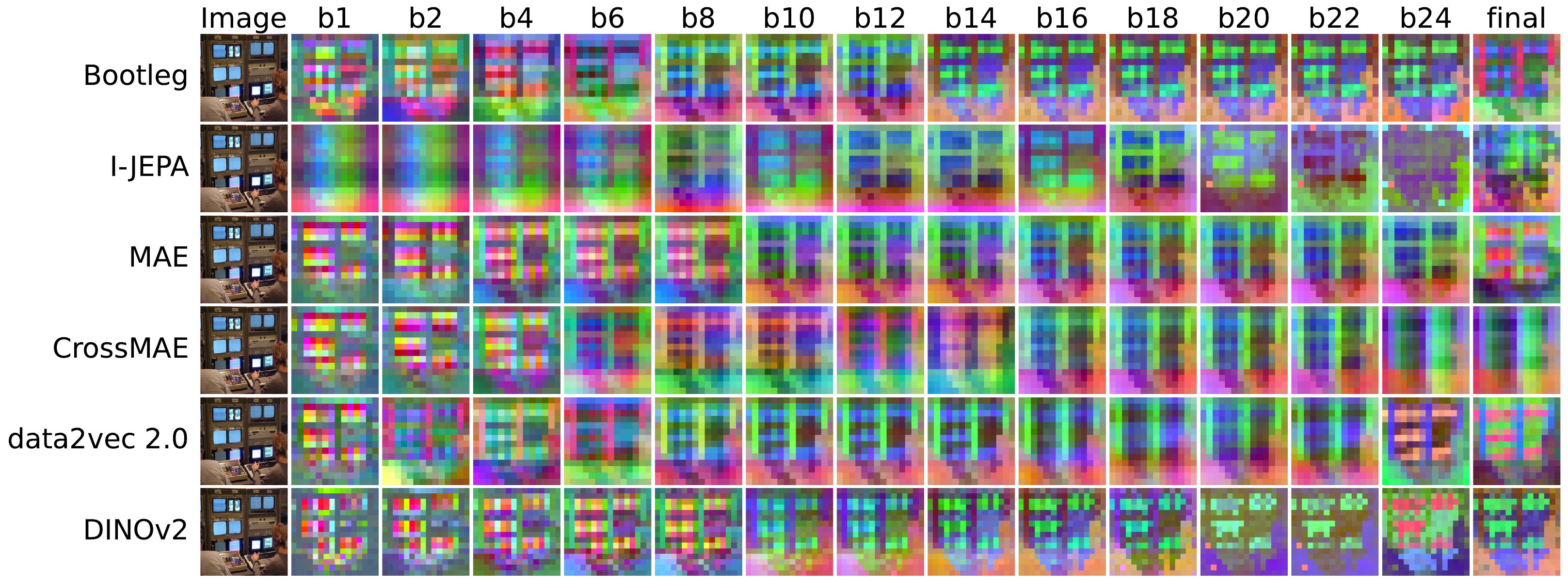}
\caption{PCA of the patch features at block~1, every second block, and the final output of ViT-L/16 encoders (L/14 for DINOv2), on the image of \cref{fig:pca-vitb-layers} (per-image PCA, CAPI protocol). In I-JEPA (blocks~20--24) and DINOv2 (blocks~18--23), a few high-norm artifact tokens dominate the principal components, and the spatial structure of the scene is lost.}
\label{fig:pca-vitl-layers}
\end{figure}

\begin{figure}[tbh]
\centering
\includegraphics[width=\linewidth]{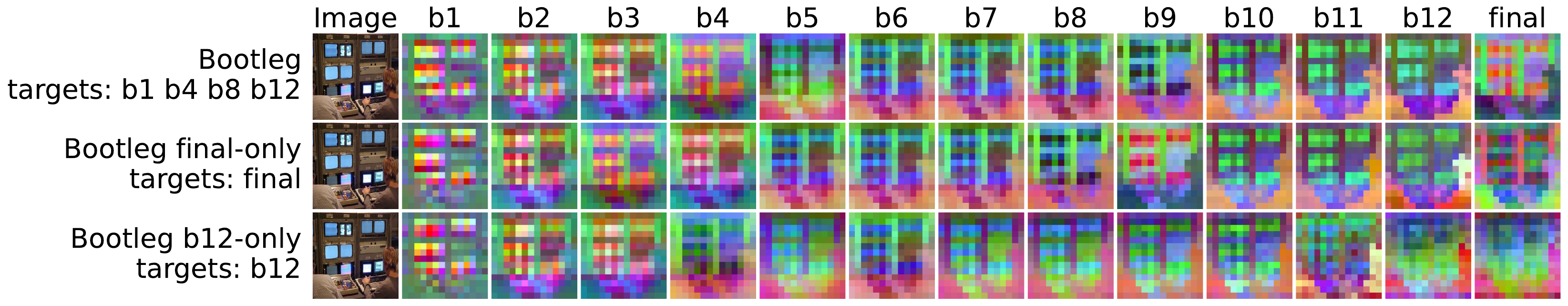}
\caption{PCA of the patch features at every block of ViT-S/16 Bootleg variants trained for 300 ep., on the image of \cref{fig:pca-vitb-layers}: four hidden targets (Bootleg), only the final output, and only the last block's output (b12, before the final norm) (per-image PCA, CAPI protocol).}
\label{fig:pca-vits-ablations}
\end{figure}

\Cref{fig:pca-vitb-layers} compares the methods across depth on a single image (ViT-B/16), and \cref{fig:pca-vitb-final} compares the output of their last block (b12, before the final norm) across images.
Bootleg retains the layout of the scene (here, the grid of monitors) in every block.
Our I-JEPA reproduction develops isolated outlying patches from block~8 onward, mostly along the image border.
In the last block across images (\cref{fig:pca-vitb-final}), Bootleg and DINOv2 give the most coherent object regions (\eg{} the dogs, the boat), whereas I-JEPA and data2vec\,2.0 are the noisiest.
data2vec\,2.0 loses spatial detail in blocks~7--11, MAE and CrossMAE give image-like maps (CrossMAE's being the smoothest), and DINOv2's blocks~9--11 collapse to a near-uniform map with a few outlying patches.

To test whether such outlying patches are the high-norm artifact tokens described by \citet{darcet2024registers}, we measured the norm of every patch token across the 18 images, at every layer.
Tokens with more than three times the median norm of their image occur in DINOv2 ViT-B/14 (blocks~9--11, 1.5\% of patches, in every image, up to $62\times$ the median) and ViT-L/14 (blocks~18--23), and in our I-JEPA ViT-L/16 (blocks~20--24, about two patches per image, in every image, $12$--$22\times$ the median).
None occur in Bootleg, MAE, CrossMAE, or data2vec\,2.0, at any model size.
In these blocks the high-norm tokens dominate the principal components, which is why I-JEPA ViT-L and DINOv2 lose the spatial structure of the scene there (\cref{fig:pca-vitl-layers}), whereas Bootleg ViT-L retains it throughout.
The final norm removes the norm outliers (at most $1.5\times$ the median at the final output).
The outlying patches of I-JEPA at ViT-S and ViT-B are not high-norm (at most $1.9\times$ the median), but at ViT-S they are positional: the most outlying patch of the last block lies at the same top-row position in 16 of the 18 images (and at ViT-L, block~19, at the same corner position in 14 of 18), whereas Bootleg's most outlying patches recur at a given position no more often than chance (at most 3 of 18).
Bootleg's encoder has four register tokens and a CLS token, and its absence of artifact tokens is consistent with registers absorbing them \citep{darcet2024registers}, although Bootleg also differs from I-JEPA in its training objective.

In \cref{fig:pca-vits-ablations}, we compare Bootleg with variants distilling only the final output, or only the output of the last block (b12, before the final norm), at ViT-S.
Distilling only the last block's output loses the layout of the scene in its last blocks (block~11 onward), whereas Bootleg, with its four hidden targets, retains it up to the final output.

\FloatBarrier
\subsection{Inter-layer similarity}
\label{sec:inter-layer-sim}

\cref{fig:pearson-matrices} shows Pearson correlation matrices between all pairs of encoder layers (tokenizer output, blocks~1--12, and the final layer-normed output used as I-JEPA's target).
Several patterns emerge.
Adjacent layers are highly correlated (mean adjacent Pearson $r = 0.81$ for ViT-S, $0.84$ for ViT-B), but correlation decays rapidly with layer distance, reaching near-zero between early and late blocks.
Block~12's output is notably distinct from earlier layers, with Pearson $r < 0.10$ against blocks~1--4 in ViT-S.
The ``final'' representation (post-LayerNorm) shows moderate correlation with late blocks but low correlation with early blocks, confirming that it is dominated by the deepest layers.

\cref{fig:cka-matrices} shows the corresponding CKA matrices.
CKA values are systematically higher than Pearson correlations for distant layer pairs (\eg{} CKA between tokenizer and final is 0.58 for ViT-S vs.\ Pearson $r = 0.05$), reflecting CKA's sensitivity to shared geometric structure even when per-element correspondence is weak.
Despite this scale difference, CKA and Pearson rankings are strongly correlated ($\rho = 0.89$ for both architectures; \cref{fig:cka-vs-pearson}).

\begin{figure}[tbh]
\centering
\includegraphics[width=\linewidth]{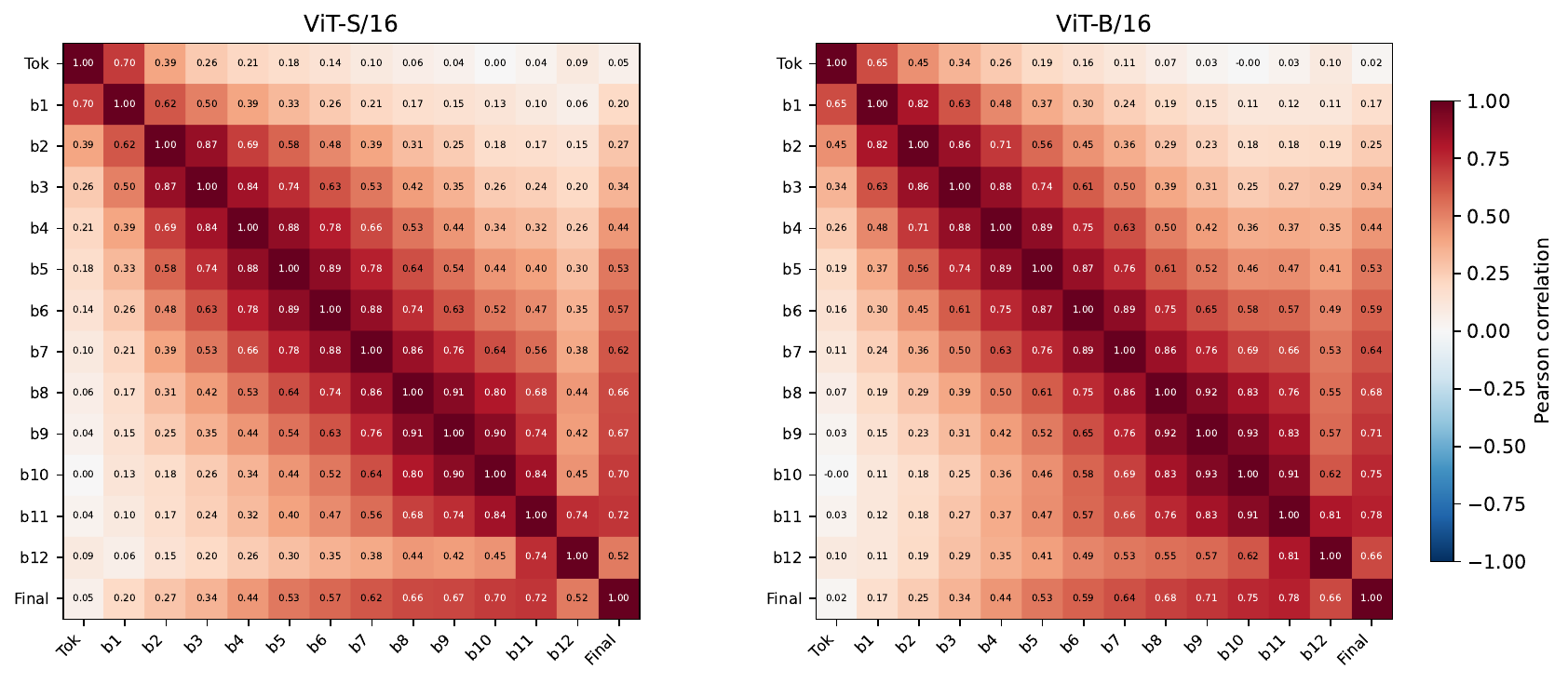}
\caption{%
Inter-layer Pearson correlation matrices for Bootleg ViT-S/16 and ViT-B/16 encoders.
Each cell shows the mean correlation between embedding vectors at corresponding spatial positions across 10{,}000 images.
Correlation decays with layer distance, and the final block is highly distinct from early blocks.
}
\label{fig:pearson-matrices}
\end{figure}

\begin{figure}[tbh]
\centering
\includegraphics[width=\linewidth]{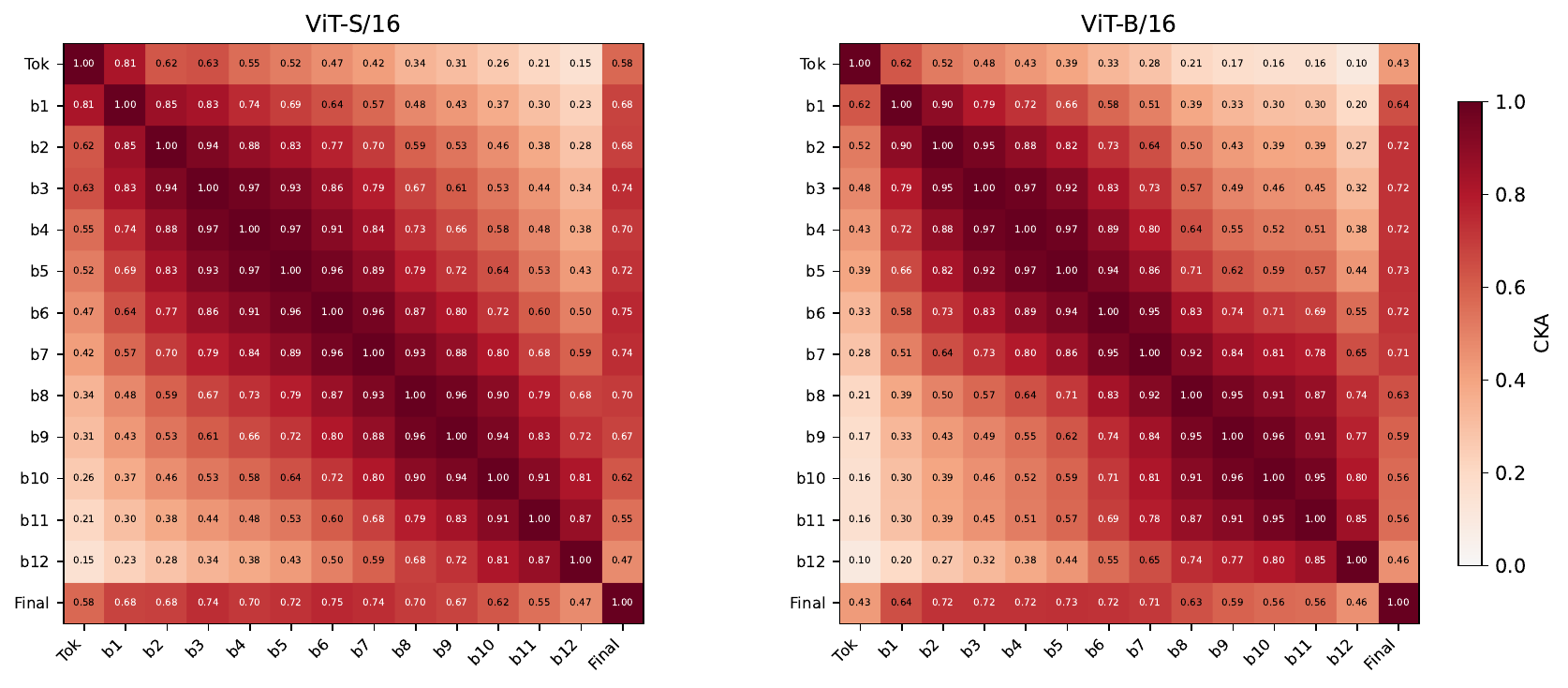}
\caption{%
Inter-layer CKA similarity matrices.
CKA captures shared representational geometry and yields higher values than Pearson correlation for distant layer pairs, but the two measures are strongly rank-correlated ($\rho \approx 0.89$).
}
\label{fig:cka-matrices}
\end{figure}

\begin{figure}[tbh]
\centering
\includegraphics[width=\linewidth]{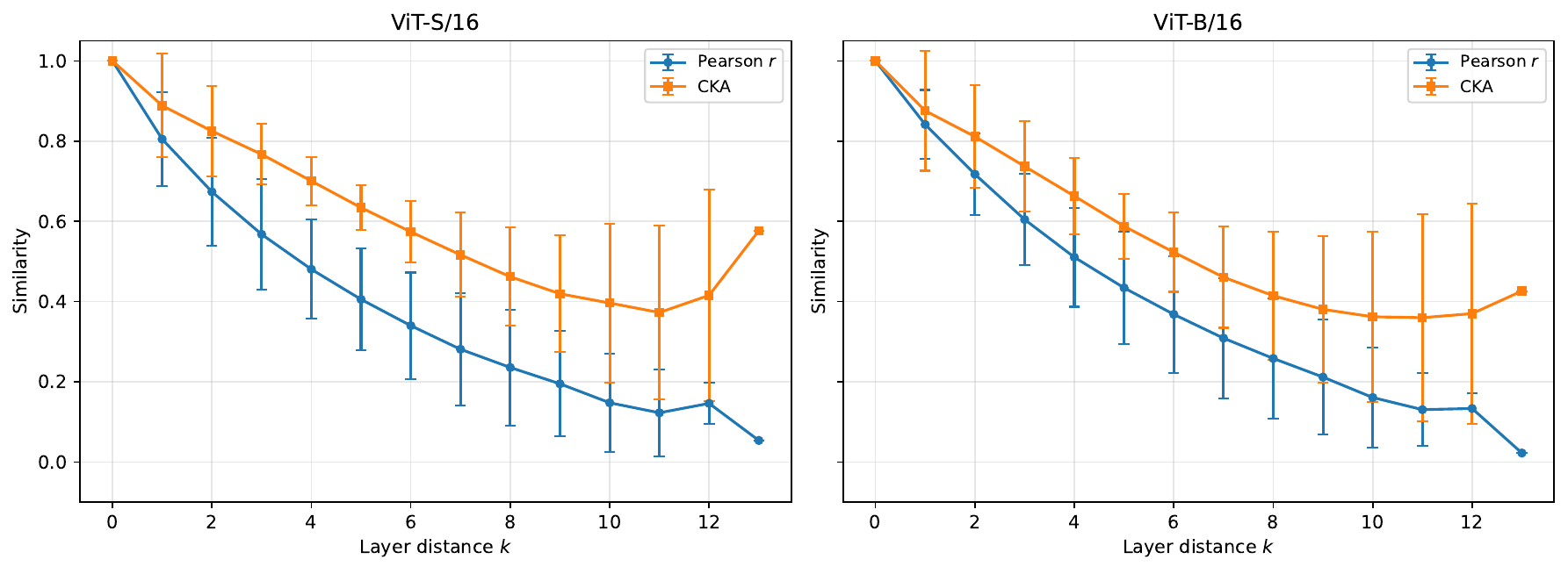}
\caption{%
Off-diagonal decay of Pearson correlation and CKA as a function of layer distance~$k$.
CKA decays more slowly, remaining above 0.15 even at the maximum distance, while Pearson correlation drops near zero.
Error bars show standard deviation across all layer pairs at each distance.
}
\label{fig:cka-vs-pearson}
\end{figure}

\FloatBarrier
\subsection{Distillation target layer analysis}
\label{sec:target-layer-analysis}

Bootleg uses blocks~1, 4, 8, and 12 as distillation targets.
\cref{fig:target-layer-corr} shows how each layer's representation correlates with these four targets.
Each target block correlates most strongly with its immediate neighbors and has a peaked profile, confirming that the targets sample distinct regions of the representation space.
The early target (block~1) and late target (block~12) are particularly dissimilar (Pearson $r < 0.06$ for ViT-S), supporting the hypothesis that multi-layer targets provide complementary training signal spanning low-level and semantic features.

\begin{figure}[tbh]
\centering
\includegraphics[width=\linewidth]{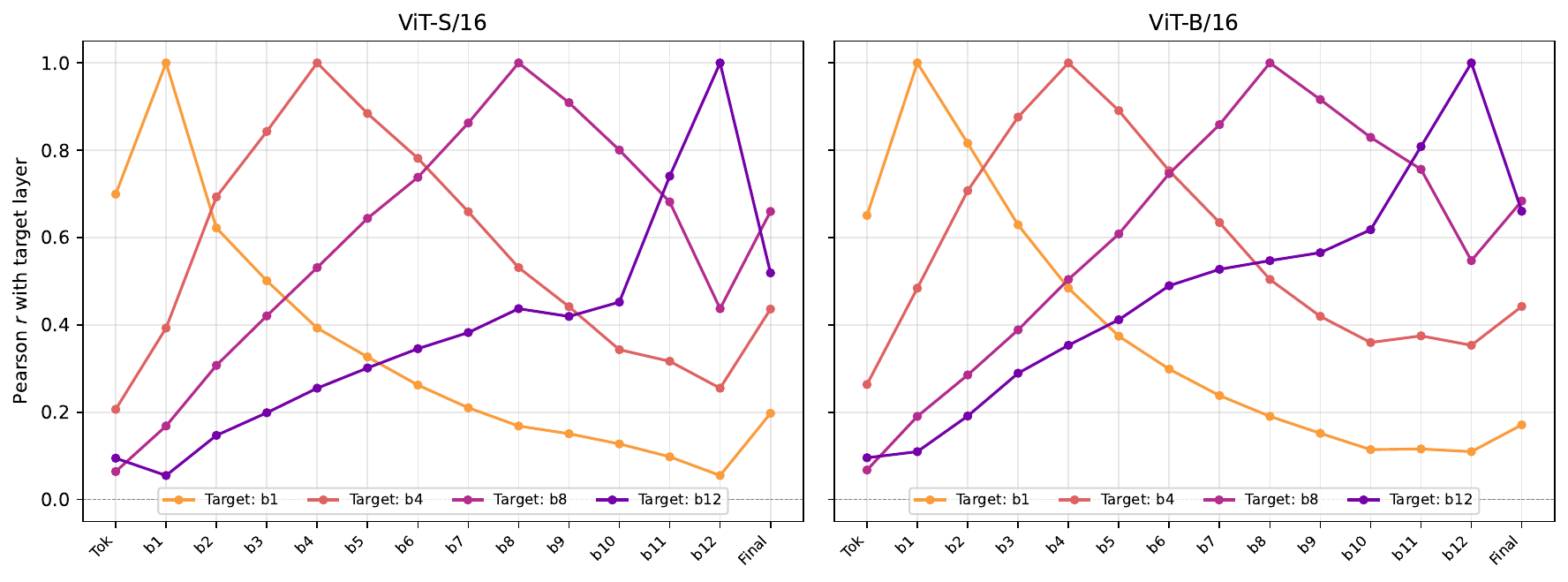}
\caption{%
Pearson correlation of each encoder layer with the four Bootleg distillation target layers (blocks 1, 4, 8, and 12).
Each target's profile peaks at its own block and decays with distance, confirming that the targets capture non-redundant representations.
}
\label{fig:target-layer-corr}
\end{figure}

\FloatBarrier
\subsection{Spatial correlation structure}
\label{sec:spatial-corr}

We also analyze how patch embeddings within a single layer correlate as a function of their spatial separation.
For each layer, we compute the Pearson correlation between all pairs of patch tokens at a given spatial offset $(\Delta r, \Delta c)$ and average over images.

\cref{fig:spatial-radial} shows the azimuthally averaged spatial correlation as a function of distance (in patch units).
In early layers, spatial correlation decays steeply---nearby patches are moderately correlated but distant patches are nearly independent.
In the deepest block (block~12), spatial correlations are uniformly high (global mean 0.76 for ViT-S, 0.61 for ViT-B), reflecting the emergence of a global, spatially uniform representation.
The final LayerNorm output partially reverses this effect, lowering the global mean correlation and restoring some spatial structure.
Across all layers, ViT-B shows slightly lower spatial correlations than ViT-S, consistent with the larger model retaining more spatially-specific information.

\begin{figure}[tbh]
\centering
\includegraphics[width=\linewidth]{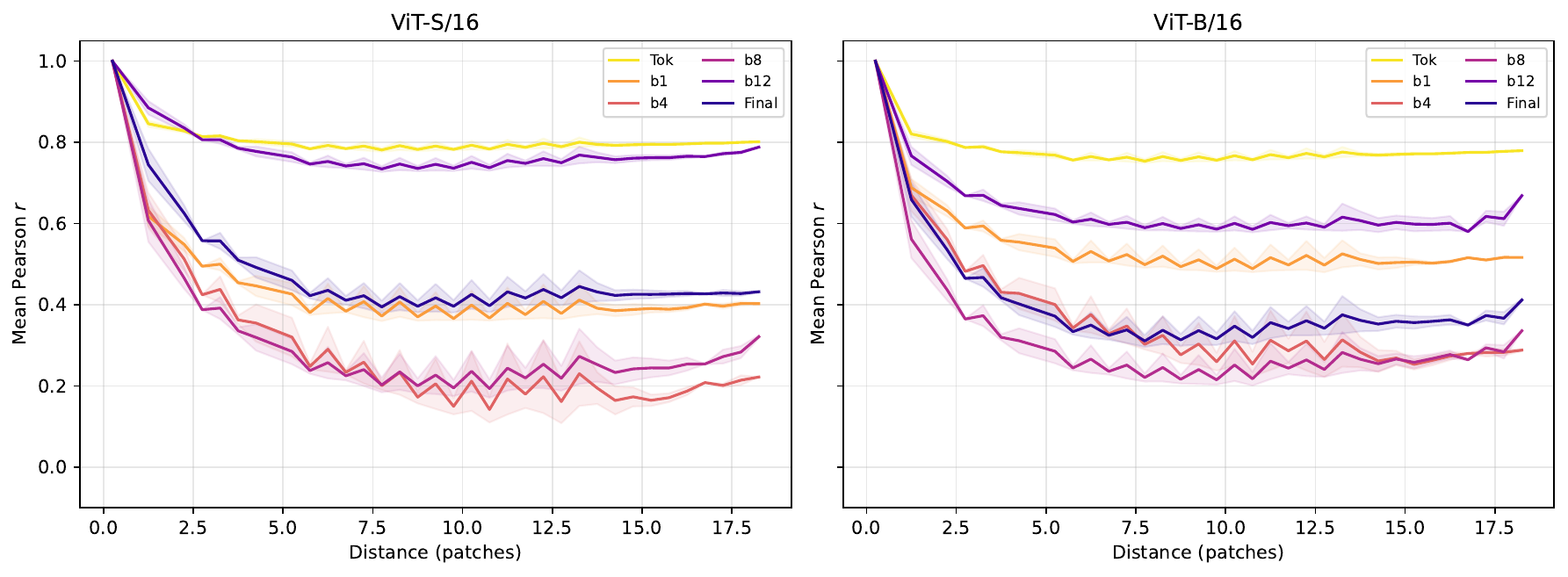}
\caption{%
Azimuthally averaged spatial correlation vs.\ distance for selected encoder layers.
Early layers exhibit localized correlations that decay with distance.
Block~12 shows near-uniform high correlation, indicating a globally coherent representation.
The post-LayerNorm final output partially restores spatial structure.
}
\label{fig:spatial-radial}
\end{figure}

\FloatBarrier
\subsection{Per-block decoding performance}
\label{sec:per-block-decoding}

The inter-layer correlation analyses above describe how representations differ across depth in geometric terms; here we ask the complementary functional question---how well does each block's output support an IN-1k classifier on its own?
\cref{fig:per-block-knn} plots IN-1k kNN top-1 (best $k$ per layer) at the tokenizer output, every transformer block, and the post-LayerNorm final output, using both patch-averaged tokens and the CLS token where available.
This complements \cref{sec:target-layer-analysis} by quantifying how much downstream-relevant signal each candidate target layer carries.

At all three sizes, the methods are similar through the early and middle blocks: the kNN curves rise together up to roughly block~9 at ViT-S/16 and ViT-B/16, and block~15 at ViT-L/16.
The main exceptions are data2vec\,2.0 at ViT-B/16 and ViT-L/16, which carries more information than the other models in the middle of the network and peaks early (block~7 and block~14, respectively) before falling off, and I-JEPA at ViT-L/16, which lags the others through the middle blocks.
When we get to the deepest blocks, we see a divergence.
At ViT-S/16 and ViT-B/16, I-JEPA and CrossMAE plateau or decline slightly after about block~9--11, while MAE keeps improving but remains far lower.
At ViT-L/16, MAE and CrossMAE peak around blocks~18--20 and then decline, whereas I-JEPA keeps improving up to its output.
Bootleg's CLS readout, in contrast, keeps improving through the last block at ViT-S/16 and ViT-B/16 (and up to block~23 of~24 at ViT-L/16), reaching 64.8\%, 72.2\%, and 75.5\% at ViT-S/16, ViT-B/16, and ViT-L/16, respectively.
This is $24$, $16$, and $14$ pp above the best I-JEPA layer, and $30$, $26$, and $13$ pp above the best MAE/CrossMAE layer.
The margin narrows with scale: with patch-averaged features at ViT-L/16, Bootleg's best layer (61.2\%) only matches I-JEPA's (61.2\%, at its output) and is below the mid-depth peak of data2vec\,2.0 (63.2\% at block~14), so its advantage at that size comes from the CLS readout.
In other words, the gain Bootleg achieves over its closest competitors is largely concentrated in the deepest few blocks: it does not learn substantially \emph{different} early or middle features, but it keeps the late features from saturating, plateauing, or collapsing into a less linearly separable space.
The post-LayerNorm ``final'' patch-avg readout falls well below the last block for Bootleg, MAE, and data2vec\,2.0 (ViT-B/16 and ViT-L/16), but changes little for CrossMAE and I-JEPA.
For Bootleg this mirrors the inter-layer Pearson decay in \cref{fig:pearson-matrices}: LayerNorm rescales the deepest features into the I-JEPA-style distillation target, and the rescaled features are less linearly separable than the pre-norm activations they are derived from.
Bootleg's CLS readout, however, remains close to its last-block value.

\paragraph{Relation to \citet{bolya2025perception}.}
Recent work on the Perception Encoder reports an analogous phenomenon for contrastive vision--language pretraining: ``the best visual embeddings are not at the output of the network''.
There, a CLIP-style ViT trained only with a global image--text contrastive loss develops its strongest dense representations several blocks before the output, and Bolya et~al.\ propose two \emph{post-hoc} alignment methods---language alignment for VLM and spatial alignment for dense prediction---to surface those hidden embeddings at the network's output.
\cref{fig:per-block-knn} shows that the same hidden-best-layer pattern holds for several other final-layer-only or pixel-reconstruction objectives: MAE, CrossMAE, and data2vec\,2.0 all peak strictly before the output and lose accuracy at the final layer, as does I-JEPA (more weakly) at ViT-S/16 and ViT-B/16, though not at ViT-L/16, where it improves up to its output.
Bootleg's hidden-layer self-distillation provides an alternative, \emph{built-in} resolution to this problem: by directly supervising features at multiple depths during pretraining, it pushes the information from the interior layers back into the final layer, boosting performance.
This removes the need for an external alignment stage searching for the part of the network which contains the most information at evaluation time.

\begin{figure}[tbh]
\centering
\includegraphics[width=\linewidth]{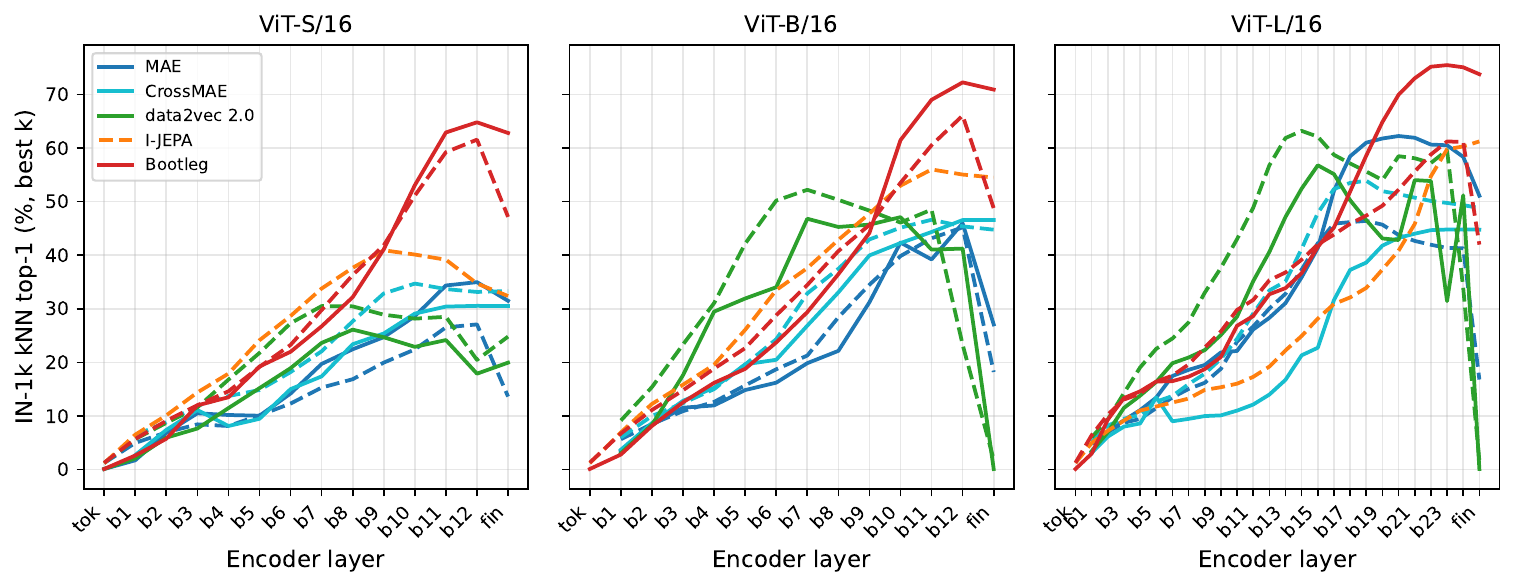}
\caption{%
Per-block IN-1k kNN top-1 accuracy for ViT-S, ViT-B, and ViT-L, evaluated at the tokenizer output (tok), every transformer block (b1--b12; b1--b24 for ViT-L/16), and the post-LayerNorm final output (fin).
We show CLS embeddings (solid) and patch-averaged embeddings (dashed); I-JEPA has no CLS token.
All methods track each other closely through the early and middle blocks and diverge in the deepest blocks: Bootleg's CLS accuracy continues to rise to (or nearly to) the last block, while the baselines plateau, or peak earlier and decline.
}
\label{fig:per-block-knn}
\end{figure}

\paragraph{Per-block decoding for the target ablation.}
\Cref{fig:knn-fig2-targets} repeats this analysis for the ViT-S/16 target-ablation runs of \cref{fig:stabilizing} (left; I-JEPA configuration, 300 ep.), with patch-averaged tokens.
Distilling only the final block, the class information peaks at block~10 (42.5\%) and decays to 36.1\% by block~12, the readout used by the downstream probes.
Adding an early or middle hidden target (b1--b7) to the final one (right panel) removes this late decay: accuracy keeps rising to a peak at block~11 or~12, reaching 43.4--46.9\% at block~12.
Deep auxiliary targets (b8--b10) instead lower the whole late-network curve, to 33.0--35.2\% at block~12, below the final-only run.
Without the final target (left panel), every hidden target yields a weaker block-12 representation than its paired counterpart; the mid-depth targets are the weakest on their own (b6: 24.7\%), and the same b6 target gives 45.8\% when paired with the final block.
This mirrors the attentive-probe results of \cref{fig:stabilizing}, and supports our interpretation that an early-to-mid target keeps the encoder building class information through its final blocks, while the final target supplies the abstraction that a mid-depth target cannot provide on its own.

\begin{figure}[tbh]
\centering
\includegraphics[width=\linewidth]{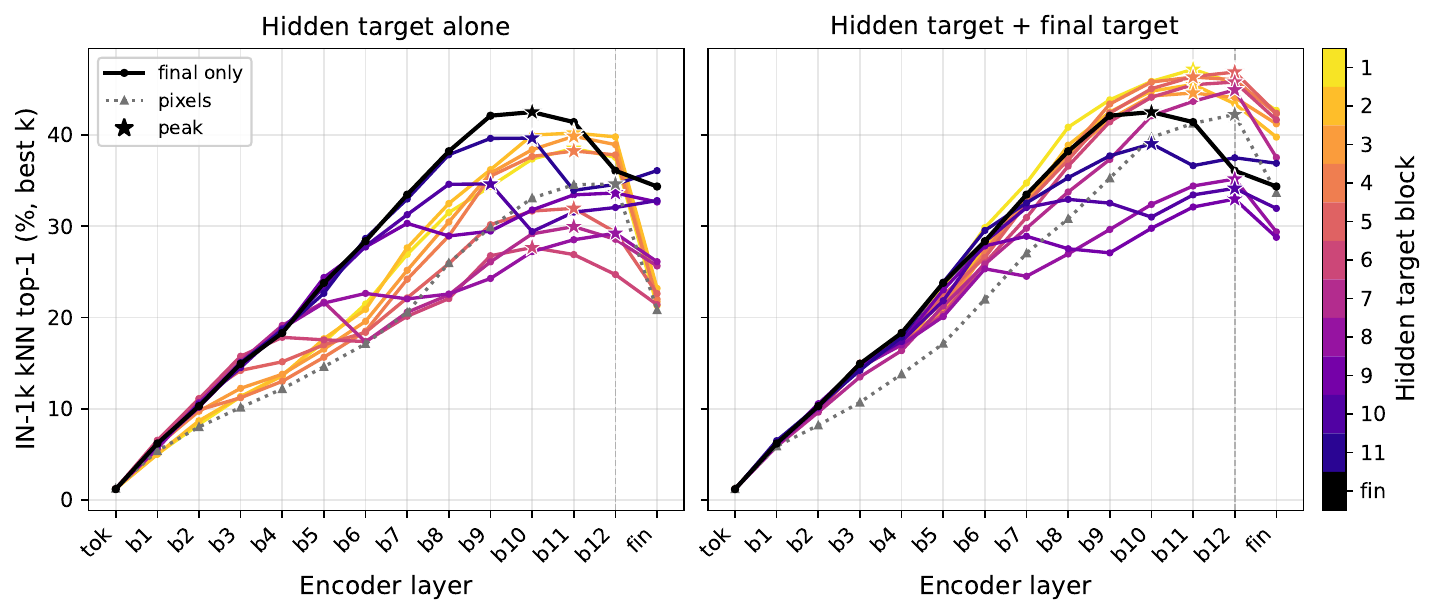}
\caption{%
Per-block IN-1k kNN top-1 accuracy (patch-averaged tokens, best $k$) for the ViT-S/16 target-ablation runs of \cref{fig:stabilizing} (left).
The x-axis is the depth within the \emph{trained} encoder at which features are read out for the kNN classifier: the tokenizer output (tok), every transformer block (b1--b12), and the post-LayerNorm final output (fin) passed to the predictor module.
Line colour indicates the depth of the hidden block used as the self-distillation target \emph{during pretraining}; the final-only (black) and pixel-only (grey) target runs are shown in both panels for reference.
Left: the hidden target alone. Right: the hidden target paired with the final-block target.
Stars mark each curve's peak, and the dashed vertical line marks block~12, the readout used by the downstream probes in the main results.
}
\label{fig:knn-fig2-targets}
\end{figure}

\FloatBarrier
\section{Target distribution, loss, and training stability}
\label{s:training-dynamics}

The training-stability analysis in \cref{s:ablation:epochs} (\cref{fig:train-curve}) shows that with prolonged training (beyond 200 epochs), distilling against only the final encoder block (``b12-only'' for ViT-S/16) causes downstream kNN performance to fall and embedding rank to collapse, while predicting multiple hidden layers remains stable.
In this section we examine the underlying training dynamics in more detail, decomposing the per-target loss for the multi-target run, characterizing the time-evolution of the target distributions, and contrasting these dynamics with the single-target collapse.
All analyses use the same three 1200-epoch ViT-S/16 runs (Bootleg with targets at blocks 1/4/8/12, single-target b12-only, and pixel-target) introduced in \cref{s:ablation:epochs}.

We use the Roy--Vetterli effective rank~\citep{roy2007effective} as our embedding-dimensionality diagnostic: $\mathrm{ER}(X) = \exp\bigl(H(\bm{p})\bigr)$, where $\bm{p}$ is the singular-value distribution $p_i = \sigma_i / \sum_j \sigma_j$ of the centred feature matrix and $H(\bm{p}) = -\sum_i p_i \log p_i$ is its Shannon entropy.
$\mathrm{ER}$ ranges from $1$ (rank-1 collapse) to the feature dimension $D$ (uniform spectrum); we report it both in raw units and as the ratio $\mathrm{ER}/D$.

\subsection{Per-target loss decomposition}
\label{s:training-dynamics:per-target-loss}

Bootleg's total loss is the unweighted average of one MSE term per distillation target.
For the ViT-S/16 multi-target run the total loss exhibits a non-monotonic trajectory: it falls from epoch 0 to a minimum around epoch~700 and then rises gently towards the end of training (from $\sim$0.32 at epoch~700 to $\sim$0.38 at epoch~1200; \cref{fig:train-curve}).
At first glance this resembles late-training degradation, but the kNN and effective-rank diagnostics in \cref{fig:train-curve} keep improving across the same window.

\cref{fig:per-target} resolves this apparent contradiction by plotting the four per-target MSE losses separately.
Block~12's loss falls monotonically (0.38\,$\to$\,0.16): the predictor genuinely improves at reconstructing deep semantic features.
The earlier blocks (1, 4, and especially 8) all rise (block~8 from 0.30 to 0.54).
Because the total objective is an unweighted mean, the rising early-block losses dominate the late-training behaviour, even while the deepest target keeps improving.
The single-target b12-only run, in contrast, shows monotonically decreasing loss for its sole target throughout the 1200-epoch run---a signature that, combined with degrading downstream metrics (\cref{s:training-dynamics:collapse}), indicates representational collapse rather than genuine improvement.

\begin{figure}[tbh]
\centering
\includegraphics[width=\linewidth]{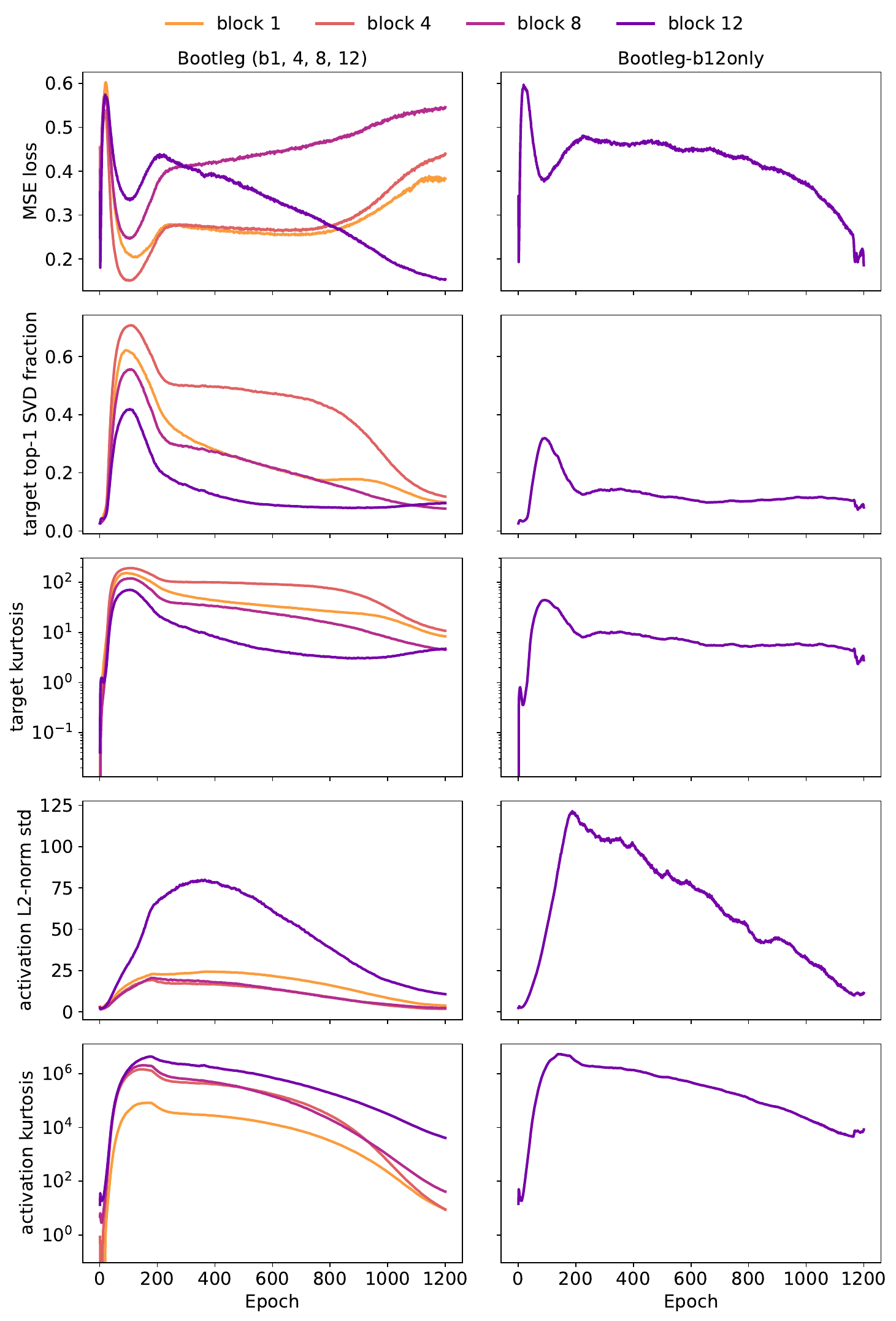}
\caption{%
Per-target MSE loss and target/activation statistics over 1200 epochs of ViT-S/16 pretraining.
\textit{Left:} multi-target Bootleg with targets at blocks 1, 4, 8, 12.
\textit{Right:} single-target b12-only.
Rows: per-target MSE loss; top-1 SVD variance fraction of the targets; target excess kurtosis (log y-axis); standard deviation of target token L2-norms; activation kurtosis at the target site (log y-axis).
Lines are coloured by target depth on a plasma sweep (yellow for early, purple for deep).
}
\label{fig:per-target}
\end{figure}

\subsection{Targets become more isotropic and Gaussian over training}
\label{s:training-dynamics:tailedness}

The per-target loss trajectories above are explained by how the target distributions themselves change over training.
We monitor two complementary measures of target structure logged at the EMA-teacher's hidden activations (after instance-norm standardization to match the loss):
(i)~the top-1 SVD variance fraction of the target embeddings (the share of total variance carried by the dominant SVD direction), and (ii)~the excess kurtosis of the target activations (a measure of heavy-tailedness).
\cref{tab:target-stats} summarises both quantities at two epoch windows for the multi-target run.

\begin{table}[tbh]
\centering
\caption{%
Target distribution statistics for multi-target Bootleg (ViT-S/16, 1200 epochs), averaged over epochs $[100, 200)$ (early training) and $[1000, 1100)$ (late training).
Higher top-1 SVD fraction means more variance concentrated along a single direction (more anisotropic).
Higher kurtosis means heavier-tailed activations (more outlier-driven).
}
\label{tab:target-stats}
\small
\begin{tabular}{lrrrr}
\toprule
 & \multicolumn{2}{c}{top-1 SVD frac.} & \multicolumn{2}{c}{kurtosis} \\
\cmidrule(lr){2-3} \cmidrule(lr){4-5}
Block & early & late & early & late \\
\midrule
 1 & 0.55 & 0.14 & 123 & 16 \\
 4 & 0.65 & 0.20 & 166 & 22 \\
 8 & 0.48 & 0.10 &  93 &  7 \\
12 & 0.33 & 0.09 &  49 &  4 \\
\bottomrule
\end{tabular}
\end{table}

Early in training, every block's targets are extremely heavy-tailed (excess kurtosis 50--170) and concentrate most of their variance along a single direction (top-1 SVD fraction 0.33--0.65).
A predictor that simply guesses the dominant direction with the appropriate magnitude will score well on the MSE objective.
By late training, the targets are much more isotropic (top-1 SVD fraction 0.09--0.20) and near-Gaussian (excess kurtosis 4--22).
Variance is now spread across many directions; predicting the targets is genuinely harder.

The exception is block~12.
Even though its tailedness shrinks at the same rate as the other blocks, its loss continues to fall throughout training, because semantic-level features are highly contextually predictable: from a small set of visible patches one can infer the global semantic content of the image, and the predictor exploits this regularity.
Low- and mid-level features at blocks 1/4/8 (textures, edges, mid-level patterns) are intrinsically less context-predictable, so as those features become richer the prediction task gets correspondingly harder.

The net interpretation is that Bootleg's late-training loss rise is a healthy signal, not a regression: rising early-block loss + falling deep-block loss + rising kNN + rising effective rank are all consistent with the encoder producing higher-rank, more class-separable features whose targets carry more information.

\subsection{Diagnostics of single-target collapse}
\label{s:training-dynamics:collapse}

The b12-only run shows the inverse pathology: total loss falls steadily, but downstream kNN accuracy and effective rank both degrade sharply from around epoch~1000 (\cref{fig:train-curve}).
With only block~12 as a distillation target, nothing anchors representations earlier in the encoder.
The EMA teacher follows the collapsing student into a lower-information subspace, and because the teacher \emph{is} the target generator, the matching task gets \emph{easier} the more both shrink---the classic single-target self-distillation collapse mode that multi-block targets specifically defend against.

We note that the block-12 activation statistics (norm-std, kurtosis) follow nearly identical arcs in Bootleg and b12-only because both runs put block~12 under EMA-distillation pressure (\cref{fig:per-target}, bottom rows).
These activation statistics reflect target-site dynamics rather than collapse-specific drift, so they are useful as within-run sanity checks but \emph{not} as cross-run discriminators.
The CLS-feature kNN accuracy and effective-rank diagnostics in \cref{fig:train-curve} are the right cross-run signals.

\FloatBarrier
\section{Effect of training-budget length}
\label{s:training-dynamics:epochs}

The 1200-epoch comparison in \cref{s:training-dynamics:collapse} establishes that single-target distillation collapses given enough training, while multi-target distillation does not.
A natural follow-up question is how this collapse interacts with the total training budget---do shorter schedules avoid the collapse simply because they finish before it sets in, or does the collapse onset move with the schedule?
We swept the total epoch budget across $\{50, 100, 200, 300, 600, 900, 1200\}$ for both run families using identical hyperparameters apart from the cosine-decayed learning-rate schedule, which is anchored to the run length.
\cref{fig:epochs-loss,fig:epochs-knn,fig:epochs-effrank} show three representative training-dynamics metrics across all 14 runs.

For multi-target Bootleg (left panels), every diagnostic improves monotonically with budget: longer schedules give higher kNN and higher effective rank.
The late-training rise in total loss (\cref{fig:epochs-loss}) is reproduced at every budget, confirming that it is an intrinsic feature of multi-target dynamics rather than a 1200-epoch artefact.

For single-target b12-only (right panels), the picture is qualitatively different.
Up to ${\sim}300$ epochs the run looks healthy on every diagnostic, but as the budget grows the late-training collapse re-appears at larger and larger absolute epoch counts: kNN peaks then falls and the effective rank drops sharply.
The collapse onset shifts later with longer schedules---roughly tracking the LR decay shape---confirming that it is the LR-schedule late tail that triggers the dynamic, not a fixed wall-clock epoch count.
Practically, at every budget where b12-only does not visibly collapse, multi-target Bootleg matches or exceeds it on every downstream-quality metric, so single-target distillation is dominated by multi-target distillation across the full compute regime.

\begin{figure}[tbh]
\centering
\includegraphics[width=\linewidth]{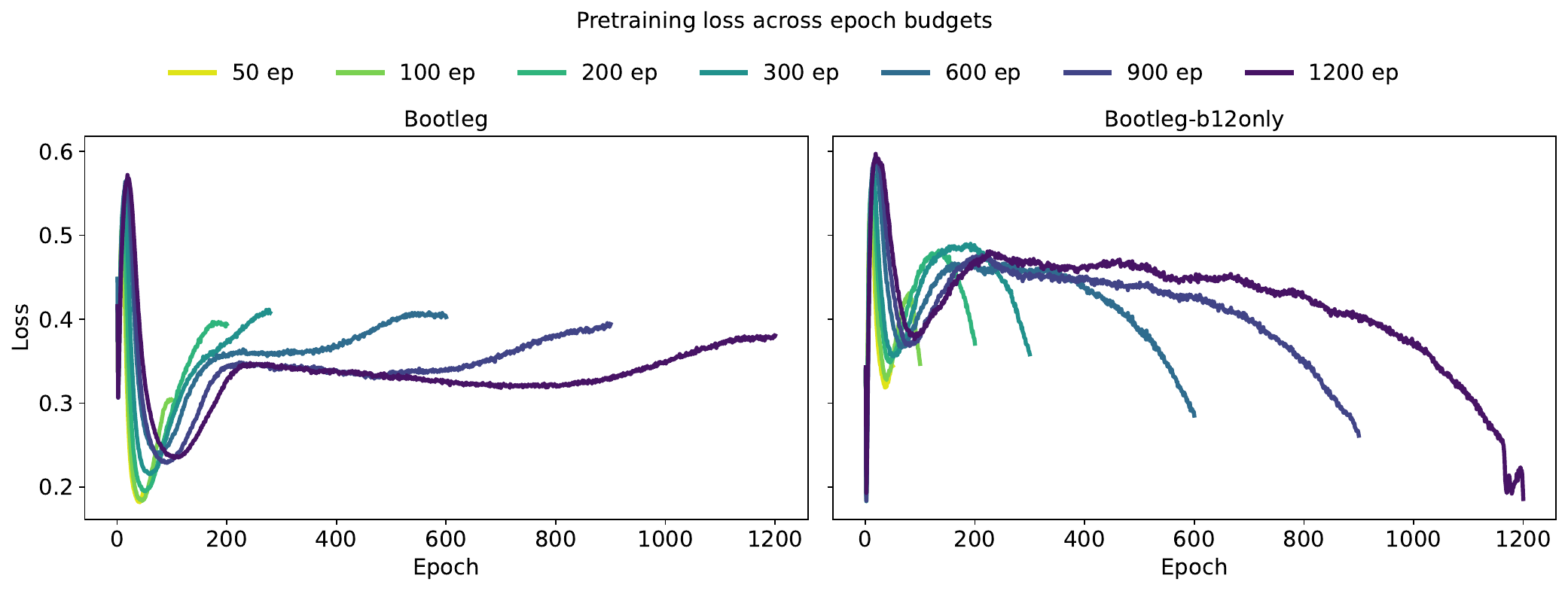}
\caption{Pretraining loss across epoch budgets, ViT-S/16. Lines are coloured by total epoch budget on a viridis sweep (bright yellow~=~50 ep, dark purple~=~1200 ep). \textit{Left:} multi-target Bootleg. \textit{Right:} single-target b12-only.}
\label{fig:epochs-loss}
\end{figure}

\begin{figure}[tbh]
\centering
\includegraphics[width=\linewidth]{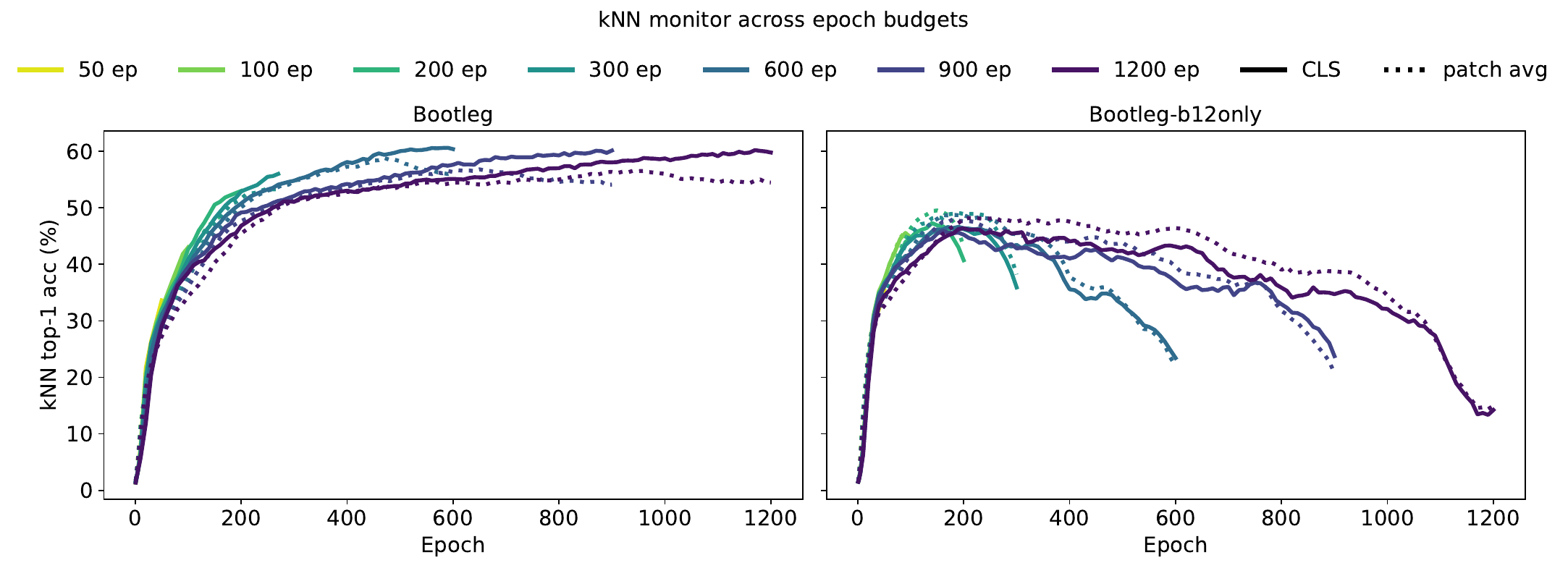}
\caption{kNN top-1 accuracy across epoch budgets. Solid: CLS feature; dotted: patch-average feature. Bootleg scales monotonically with budget; b12-only collapses on long schedules.}
\label{fig:epochs-knn}
\end{figure}

\begin{figure}[tbh]
\centering
\includegraphics[width=\linewidth]{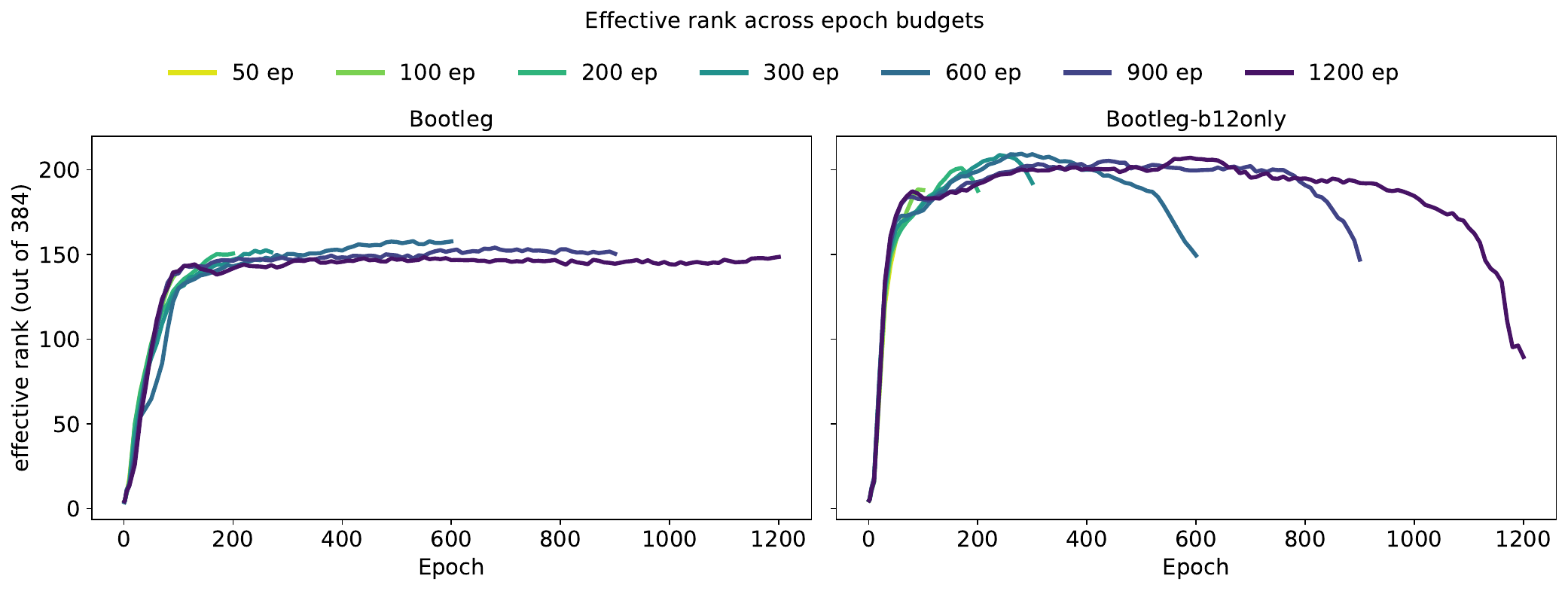}
\caption{CLS-feature effective rank~\citep{roy2007effective} across epoch budgets, out of $D = 384$.}
\label{fig:epochs-effrank}
\end{figure}

}{}

\end{document}